\documentclass[sigconf]{acmart}

\usepackage{caption}
\usepackage{subcaption}
\AtBeginDocument{%
  }

\usepackage{enumitem}
\usepackage{multirow} 

\copyrightyear{2025}
\acmYear{2025}
\setcopyright{cc}
\setcctype{by-nc-sa}
\acmConference[WWW Companion '25]{Companion Proceedings of the ACM Web Conference 2025}{April 28-May 2, 2025}{Sydney, NSW, Australia}
\acmBooktitle{Companion Proceedings of the ACM Web Conference 2025 (WWW Companion '25), April 28-May 2, 2025, Sydney, NSW, Australia}
\acmDOI{10.1145/3701716.3717744}
\acmISBN{979-8-4007-1331-6/2025/04}

\makeatletter
\DeclareRobustCommand\onedot{\futurelet\@let@token\bmv@onedotaux}
\def\bmv@onedotaux{\ifx\@let@token.\else.\null\fi\xspace}
\def\eg{\emph{e.g}.}

\def\etc{\emph{etc}.}

\makeatother

\begin{document}

\title{Do Language Models Understand Time?}


\author{Xi Ding}
\affiliation{%
  \institution{Australian National University}
    \city{Canberra}
  \state{Australian Capital Territory}
  \country{Australia}}
\email{Xi.Ding1@anu.edu.au}



\author{Lei Wang}\authornote{Corresponding author.}
\affiliation{%
  \institution{Griffith University}
  \city{Brisbane}
  \state{Queensland}
  \country{Australia}}
  \affiliation{%
  \institution{Australian National University}
  \city{Canberra}
  \state{Australian Capital Territory}
  \country{Australia}}
\email{l.wang4@griffith.edu.au}


\begin{abstract}
  Large language models (LLMs) have revolutionized video-based computer vision applications, including action recognition, anomaly detection, and video summarization. Videos inherently pose unique challenges, combining spatial complexity with temporal dynamics that are absent in static images or textual data. Current approaches to video understanding with LLMs often rely on pretrained video encoders to extract spatiotemporal features and text encoders to capture semantic meaning. These representations are integrated within LLM frameworks, enabling multimodal reasoning across diverse video tasks.
  However, the critical question persists: Can LLMs truly understand the concept of time, and how effectively can they reason about temporal relationships in videos? This work critically examines the role of LLMs in video processing, with a specific focus on their temporal reasoning capabilities. We identify key limitations in the interaction between LLMs and pretrained encoders, revealing gaps in their ability to model long-term dependencies and abstract temporal concepts such as causality and event progression. Furthermore, we analyze challenges posed by existing video datasets, including biases, lack of temporal annotations, and domain-specific limitations that constrain the temporal understanding of LLMs.
  To address these gaps, we explore promising future directions, including the co-evolution of LLMs and encoders, the development of enriched datasets with explicit temporal labels, and innovative architectures for integrating spatial, temporal, and semantic reasoning. By addressing these challenges, we aim to advance the temporal comprehension of LLMs, unlocking their full potential in video analysis and beyond.
  Our paper's GitHub repository can be found \href{https://github.com/Darcyddx/Video-LLM}{\color{blue}here}. 
\end{abstract}





\begin{teaserfigure}
\includegraphics[page=1, trim=22 170 18 158, clip, width=\textwidth]{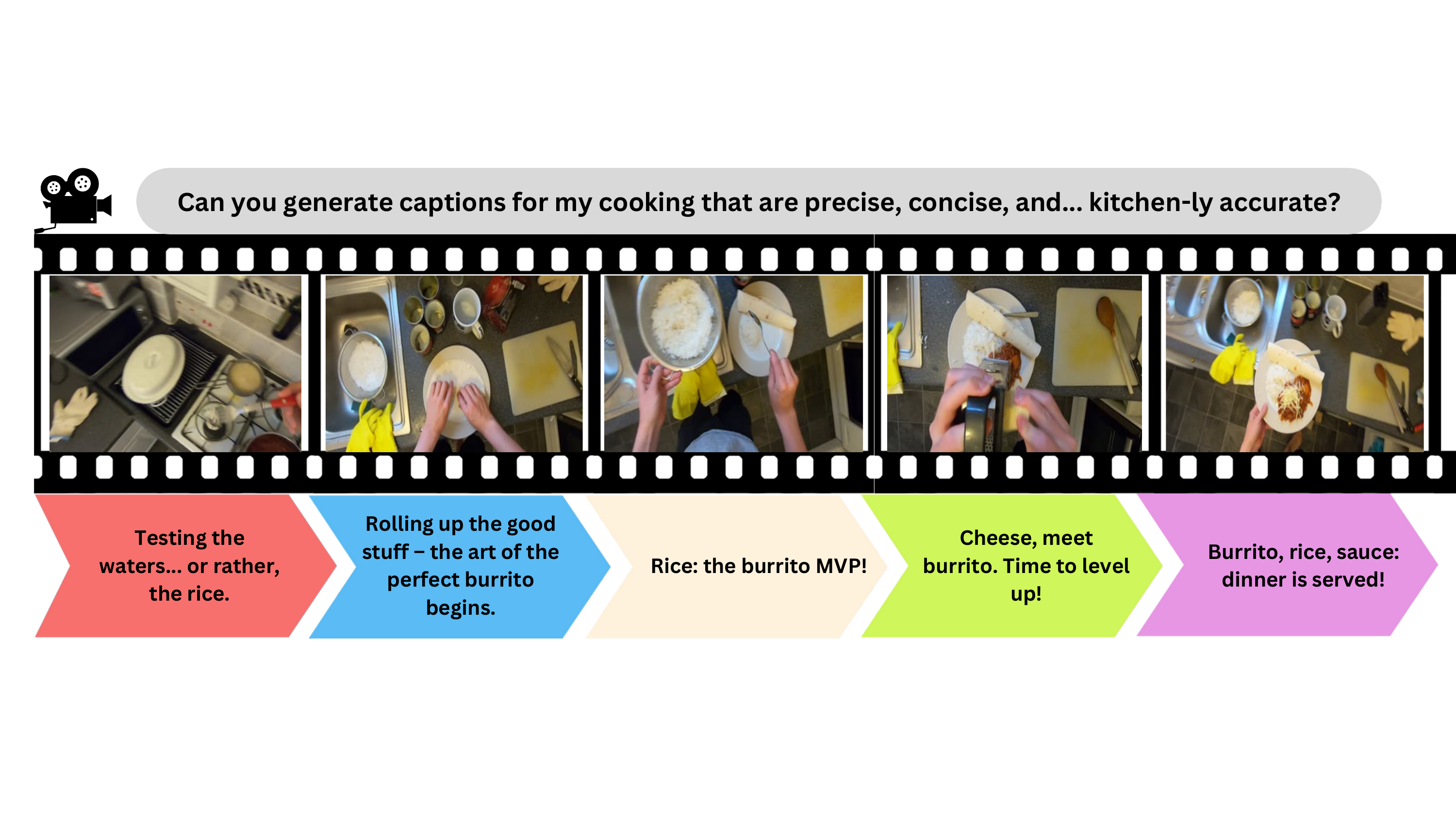}
\caption{Do language models understand time? In the kitchen arena, where burritos are rolled, rice waits patiently, and sauce steals the spotlight, LLMs try their best to keep up. Captions flow like a recipe—precise and tempting—but can they truly tell the difference between prepping, cooking, and eating? After all, in cooking, timing isn't just everything—it's the secret sauce! 
}
\label{fig:cook}
\end{teaserfigure}

\maketitle

\section{Introduction}

Large language models (LLMs) have brought transformative advancements to artificial intelligence (AI), excelling across a wide array of tasks in natural language processing and computer vision \cite{bugliarello2021multimodal, ghosh2024exploring, xu2023large}. Their ability to understand and generate human-like language has enabled groundbreaking applications, from machine translation to image and video captioning \cite{kim2024show} (see Figure \ref{fig:cook}, frames from EPIC-KITCHENS-100 \cite{damen2022rescaling}). More recently, the integration of LLMs into video processing has sparked significant interest, leading to advances in tasks such as action recognition \cite{wang2022internvideo, wang2024internvideo2}, anomaly detection \cite{wang2019loss, zhang2024holmes, msad2024}, and video summarization \cite{islam2024video, yang2024vript, liu2024kangaroo, zohar2024apolloexplorationvideounderstanding}. However, videos pose unique challenges compared to other modalities due to their dual reliance on both spatial and temporal information~\cite{chen2024spatial}. Unlike static images, videos capture the dimension of time, embedding sequential dynamics that demand sophisticated reasoning \cite{vondrick2016generating, 8328914}. Similarly, unlike textual data, videos involve rich, complex visual elements that require intricate modeling~\cite{wangtaylor, chen2024motion}.

Despite these advancements, a fundamental question remains unresolved: Do language models truly understand the concept of time? Temporal reasoning, the ability to comprehend and infer relationships between events over time, is essential for video-based tasks such as causal inference \cite{chen2021spatial, lin2022causal}, event prediction \cite{huang2024vtimellm, guo2024trace}, and understanding action progression~\cite{wang2023robust}. While pretrained video encoders provide LLMs with spatiotemporal embeddings and text encoders contribute semantic insights, the fusion of these components often lacks the nuanced understanding of time required for more advanced applications \cite{ding2024lego}. Current methods rely heavily on pretrained encoders and dataset-specific tuning \cite{qin2022fusing, wangtaylor, chen2024motion,chen2024spatial}, raising questions about the generalizability and scalability of these approaches. Table \ref{tab:llms} provides a summary of recent video-LLMs, detailing the visual encoders they use and their mechanisms for interacting with these encoders.

This work aims to address these gaps and critically examines the role of LLMs in understanding temporal dynamics in video data. We focus on the interplay between visual (image/video) encoders and LLMs, exploring how effectively they bridge the gap between raw spatiotemporal features and high-level temporal reasoning. By tackling these issues, we seek to shed light on the limitations of existing approaches and inspire innovations in LLM-based video understanding. Our \textbf{contributions} are threefold:
\renewcommand{\labelenumi}{\roman{enumi}.}
\begin{enumerate}[leftmargin=0.6cm]
\item We provide a detailed review of LLM applications in video processing, with a particular focus on their ability to comprehend temporal concepts, highlighting the state of the art and identifying key limitations.
\item We analyze the shortcomings of existing LLM-based approaches, particularly their reliance on pretrained encoders and the challenges posed by video datasets, such as lack of temporal annotations and biases towards short-term dependencies.
\item We propose actionable pathways for advancing LLMs’ temporal understanding, emphasizing joint training strategies, better dataset design, and improved alignment between spatiotemporal and semantic features.
\end{enumerate}

\begin{table*}[tbp]
    \centering
    \setlength{\tabcolsep}{0.05em}
\renewcommand{\arraystretch}{0.70}
    \resizebox{\linewidth}{!}{
    \begin{tabular}{lcllll}
        \toprule
        \textbf{Model} & \textbf{Venue} & \textbf{Visual encoder} & \textbf{Other modality encoders} & \textbf{Interaction / Fusion mechanism} & \textbf{Description} \\
        \midrule
        Flamingo \cite{alayrac2022flamingo} & NeurIPS 2022 & Normalizer-Free ResNet\cite{brock2021high} & Text: Chinchilla\cite{hoffmann2022training} & Perceiver Resampler \& Gated XATTN-DENSE & Visual-language model.
        \\
        LaViLa \cite{zhao2023learning} & CVPR 2022 & TimeSformer\cite{bertasius2021space} & Text: 12-layer Transformer & Cross-attention modules & 
        Large-scale language model. \\
        mPLUG-2 \cite{xu2023mplug} & ICML 2023 & CLIP ViT-L/14\cite{radford2021learning} & Text: BERT\cite{devlin2018bert} & Universal layers \& cross-attention modules & Modularized multi-modal foundation model. \\
        Vid2Seq \cite{yang2023vid2seq} & CVPR 2023 & CLIP ViT-L/14\cite{radford2021learning} & Text: T5-Base\cite{raffel2020exploring} & 
        Cross-modal attention & Sequence-to-sequence video-language model. 
        \\
        Video-LLaMA \cite{Zhang2023VideoLLaMAAI} & EMNLP 2023 & CLIP ViT-G\cite{radford2021learning} & Text: Vicuna\cite{chiang2023vicuna}, Audio: ImageBind\cite{girdhar2023imagebind} & Aligned via Q-Formers for video and audio & Instruction-tuned multimodal model. 
        \\
        ChatVideo \cite{wang2023chatvideo} & arXiv 2023 & \eg, OmniVL\cite{wang2022omnivl}, InternVideo\cite{wang2022internvideo} &  Text: ChatGPT\cite{wu2023visualchatgpttalkingdrawing}, Audio: \eg, Whisper\cite{radford2021learning}
        & Tracklet-centric with ChatGPT reasoning & Chat-based video understanding system. \\
        VideoChat \cite{li2023videochat} & arXiv 2023 & EVA-CLIP ViT-G/14 \cite{sun2023eva} & Text: StableVicuna\cite{stablelm2023}, Audio: Whisper\cite{radford2021learning} & Q-Former bridges visual features to LLMs for reasoning & Chat-centric model. 
        \\
        VideoLLM \cite{chen2023videollm} & arXiv 2023 & \eg, I3D\cite{carreira2017quo}, SlowFast \cite{feichtenhofer2019slowfast} & Text: \eg, BERT\cite{devlin2018bert}, T5\cite{raffel2020exploring} & Semantic translator aligns visual and text encodings & Video sequence modeling using LLMs. \\
        VAST \cite{chen2023vast} & NeurIPS 2023 & EVA-CLIP ViT-G/14\cite{sun2023eva} & Text: BERT\cite{devlin2018bert}, Audio: BEATs\cite{chen2022beats} & Cross-attention layers & Omni-modality foundational model. 
        \\
        Video-ChatGPT \cite{maaz2023video} & ACL 2023 & CLIP ViT-L/14 \cite{radford2021learning} &  Text: Vicuna-v1.1\cite{liu2023visualinstructiontuning} & Spatiotemporal features projected via linear layer & Integration of vision and language for video understanding. \\
        Valley \cite{luo2023valley} & arXiv 2023 & CLIP ViT-L/14 \cite{radford2021learning} & Text: StableVicuna\cite{stablelm2023} & Projection layer
        & LLM for video assistant tasks. \\
        Macaw-LLM \cite{lyu2023macaw} & arXiv 2023 & CLIP ViT-B/16 \cite{radford2021learning} & Text: LLAMA-7B\cite{touvron2023llama}, Audio: Whisper\cite{radford2021learning} & Alignment module unifies multi-modal representations & Multimodal integration using image, audio, and video inputs. \\
        Auto-AD II \cite{han2023autoad} & CVPR 2023 & CLIP ViT-B/32 \cite{radford2021learning} & Text: BERT\cite{devlin2018bert} & Cross-attention layers
        & Movie description using vision and language. \\
        Video-LLaVA \cite{lin2023video} & arXiv 2023 & LanguageBind-Video \cite{zhu2023languagebind} & Text: Vicuna v1.5\cite{chiang2023vicuna} & MLP projection layer
        & Unified visual representation learning for video. \\
        GPT4Video \cite{wang2024gpt4video} & ACMMM 2023 & CLIP ViT-L/14 \cite{radford2021learning} & Text: LLaMA 2\cite{touvron2023llama} & Transformer-based cross-attention layer & Video understanding with LLM-based reasoning. \\
        LLaMA-VID \cite{li2025llama} & ECCV 2023 & CLIP ViT-L/14 \cite{radford2021learning} & Text: Vicuna\cite{chiang2023vicuna} & Context attention and linear projector & LLaMA-VID for visual-textual alignment in video. \\
        InternVideo2 \cite{wang2024internvideo2} & ECCV 2023 & InternVL-6B\cite{chen2024internvl}, VideoMAE V2 \cite{wang2023videomae} & Text: BERT-Large\cite{devlin2018bert}, Audio: BEATs\cite{chen2022beats} & Q-Former aligns multi-modal embeddings & Foundation model for multimodal video understanding. \\
        COSMO \cite{wang2024cosmo} & arXiv 2024 & CLIP ViT-L/14\cite{radford2021learning} & Text: OPT-IML\cite{iyer2022opt}/RedPajama\cite{together2023redpajama}Mistral\cite{jiang2023mistral} & Gated cross-attention & Contrastive-streamlined multimodal model. 
        \\
        VTimeLLM \cite{huang2024vtimellm} & CVPR 2024 & CLIP ViT-L/14 \cite{radford2021learning} & Text: Vicuna\cite{chiang2023vicuna} & Linear layer
        & Temporal video understanding enhanced with LLMs. \\
        VILA \cite{lin2024vila} & CVPR 2024 & CLIP ViT-L/336px\cite{radford2021learning} & Text: LLaMA-2-7B/13B\cite{touvron2023llama} & Linear layer & Vision-language model. 
        \\
        Video ReCap \cite{islam2024video} & CVPR 2024 & TimeSformer \cite{bertasius2021space} & Text: GPT-2\cite{radford2019language} & Cross-attention layers
        & Recursive hierarchical captioning model
        \\
        OmniViD \cite{wang2024omnivid} & CVPR 2024 & VideoSwin \cite{liu2022video} & Text: BART\cite{lewis2020bart}& MQ-Former 
        & Generative model for universal video understanding. \\
        VTG-LLM \cite{guo2024vtg} & arXiv 2024 & EVA-CLIP ViT-G/14\cite{sun2023eva} & Text: LLaMA-2-7B\cite{touvron2023llama} & Projection layer & Enhanced video temporal grounding. 
        \\
        AutoAD III \cite{han2024autoad} & CVPR 2024 & EVA-CLIP ViT\cite{sun2023eva} & Text: GPT-3.5-turbo & Shared Q-Former 
        & Video description enhancement with LLMs. \\
        LAVAD \cite{zanella2024harnessing} & CVPR 2024 & BLIP-2 ViT-L/14, ImageBind \cite{girdhar2023imagebind} & Text: Llama-2-13b-chat\cite{touvron2023llama} & Converts video features into textual prompts for LLMs & Training-free video anomaly detection using LLMs. \\
        MA-LMM \cite{he2024ma} & CVPR 2024 & EVA-CLIP ViT-G/14 & Text: Vicuna\cite{chiang2023vicuna} & A trainable Q-Former & Memory-augmented large multimodal model. 
        \\
        MiniGPT4-Video \cite{ataallah2024minigpt4} & arXiv 2024 & EVA-CLIP ViT\cite{sun2023eva} & Text: LLaMA 2\cite{touvron2023llama} & Concatenates visual tokens and projects into LLM space & Video understanding with visual-textual token interleaving. \\
        PLLaVA \cite{xu2024pllava} & arXiv 2024 & CLIP ViT-L/14 \cite{radford2021learning} & Text: LLAMA-7B\cite{touvron2023llama} & MM projector with adaptive pooling & Parameter-free extension for video captioning tasks. \\
        V2Xum-LLaMA \cite{hua2024v2xum} & arXiv 2024 & CLIP ViT-L/14 \cite{radford2021learning} & Text: LLaMA 2\cite{touvron2023llama} & Vision adapter & Video summarization using temporal prompt tuning. \\
        VideoChat2 \cite{li2024mvbench} & CVPR 2024 & UMT-L\cite{liu2022umt} & Text: Vicuna\cite{chiang2023vicuna} & Linear projection & A comprehensive multi-modal video understanding benchmark. \\
        MotionLLM \cite{chen2024motionllm} & arXiv 2024 & LanguageBind\cite{zhu2023languagebind}, VQ-VAE\cite{zhang2023t2mgptgeneratinghumanmotion} & Text: Vicuna\cite{chiang2023vicuna} &  Modality translator:
        Motion / Video translator  & Understanding human behaviors from human motions and videos. \\
        VideoGPT+ \cite{maaz2024videogpt+} & arXiv 2024 & CLIP ViT-L/14\cite{radford2021learning}, InternVideo2\cite{wang2024internvideo2} & Text: Phi-3-Mini-3.8B\cite{abdin2024phi3technicalreporthighly} & MLP & 
        Enhanced video understanding. \\
        EmoLLM \cite{yang2024emollm} & arXiv 2024 & CLIP ViT-L/14\cite{radford2021learning} & Text: Vicuna-v1.5\cite{chiang2023vicuna}, Audio: Whisper\cite{radford2021learning} & Multi-perspective visual projection
        & Multimodal emotional understanding with improved reasoning. 
        \\
        Holmes-VAD \cite{zhang2024holmes} & arXiv 2024 & LanguageBind ViT-L/14\cite{zhu2023languagebind} & Text: LLaMA3-Instruct-70B\cite{meta2024llama3} & Temporal sampler & Multimodal LLM for video anomaly detection. \\
        ShareGPT4Video \cite{chen2024sharegpt4video} & arXiv 2024 & CLIP ViT-L/14\cite{radford2021learning} & Text: Mistral-7B-Instruct-v0.2\cite{jiang2023mistral} & MLP & Precise and detailed video captions with hierarchical prompts.\\
        Vriptor \cite{yang2024vript} & arXiv 2024 & EVA CLIP ViT-L/14\cite{sun2023eva} & Text: ST-LLM\cite{liu2025st}, Audio: Whisper\cite{radford2021learning} & Scene-level sequential alignment
        & Vriptor for dense video captioning. 
        \\
        VideoLLaMA 2 \cite{cheng2024videollama} & arXiv 2024 & CLIP ViT-L/14\cite{radford2021learning} & Text: LLAMA 1.5\cite{liu2024improvedbaselinesvisualinstruction}, Audio: BEATs\cite{chen2022beats} & Spatial-Temporal Convolution (STC) connector  & Advancing spatial-temporal modeling and audio understanding. \\
        VideoLLM-online \cite{chen2024videollm} & CVPR 2024 & CLIP ViT-L/14\cite{radford2021learning} & Text: Llama-2-Chat\cite{touvron2023llama}/Llama-3-Instruct\cite{meta2024llama3} & MLP projector & Online video large language model for streaming video. \\
        Video-CCAM \cite{fei2024video} & arXiv 2024 & SigLIP-SO400M\cite{zhai2023sigmoid} & Text: Phi-3-4k-instruct\cite{abdin2024phi3technicalreporthighly}/ Yi-1.5-9B-Chat\cite{ai2024yiopenfoundationmodels} & Cross-attention-based projector
        & Causal cross-attention masks for short and long videos. \\
        LongVA \cite{zhang2024analyzing} & arXiv 2024 & CLIP ViT-L/336px \cite{radford2021learning} & Text: Qwen2-Extended\cite{bai2023qwen, yang2024qwen2} & MLP & Long context video understanding. 
        \\
        InternLM-XComposer-2.5\cite{zhang2024task} & arXiv 2024 & CLIP ViT-L/14 \cite{radford2021learning} & Text: InternLM2-7B\cite{cai2024internlm2}, Audio: Whisper\cite{radford2021learning} & MLP & Long-context LVLM supporting ultra-high-resolution video tasks. \\
        VITA \cite{fu2024vita} & arXiv 2024 & InternViT-300M \cite{chen2024expanding, gao2024mini, chen2024far, chen2024internvl} & Text: Mixtral-8x7B \cite{jiang2024mixtral}, Audio: Mel Filter Bank 
        & MLP & Open-source interactive multimodal LLM. 
        \\
        Kangaroo \cite{liu2024kangaroo} & arXiv 2024 & EVA-CLIP-L \cite{sun2023eva} &  Text: Llama-3-8B-Instruct\cite{meta2024llama3} & Multi-modal projector & Video-language model supporting long-context video input. \\
        Qwen2-VL \cite{wang2024qwen2} & arXiv 2024 & CLIP ViT-L/14\cite{radford2021learning} & Text: Qwen2-7B\cite{bai2023qwen, yang2024qwen2} & Cross-attention modules & Vision-language model for multimodal tasks. \\
        Oryx \cite{liu2024oryx} & arXiv 2024 & OryxViT\cite{liu2024oryx, zhai2023sigmoid} & Text: Qwen2-7B/32B\cite{bai2023qwen, yang2024qwen2} & Cross-attention & Spatial-temporal model for high-resolution understanding. \\
        Video-XL \cite{shu2024video} & arXiv 2024 & CLIP ViT-L\cite{radford2021learning} & Text: Qwen2-7B\cite{bai2023qwen, yang2024qwen2} & Visual-language projector 
        & Long-context video understanding model.\\
        SlowFocus \cite{nie2024slowfocus} & NeurIPS 2024 & CLIP ViT-L/14\cite{radford2021learning} & Text: Vicuna-7B v1.5\cite{zheng2023judgingllmasajudgemtbenchchatbot} & Visual adapter (projector layer) & Fine-grained temporal understanding in video LLM. \\
        VideoStudio \cite{long2024videodrafter} & ECCV 2024 & CLIP ViT-H/14 \cite{radford2021learning} & Text: CLIP ViT-H/14\cite{radford2021learning} & Cross-attention modules & Multi-scene video generation.
        \\
        VideoINSTA \cite{liao2024videoinsta} & arXiv 2024 & CLIP ViT-L/14 \cite{radford2021learning} & Text: Llama-3-8B-Instruct\cite{meta2024llama3} &  Self-reflective spatial-temporal fusion & Zero-shot long video understanding model. \\
        Loong \cite{wang2024loong} & arXiv 2024 & Causal 3D CNN\cite{yu2024languagemodelbeatsdiffusion} & Text: Standard text tokenizer & Decoder-only autoregressive LLM with causal attention & Autoregressive language models. \\
        TRACE \cite{guo2024trace} & arXiv 2024 & CLIP ViT-L\cite{radford2021learning} & Text: Mistral-7B\cite{jiang2023mistral} & Task-interleaved sequence modeling \& Adaptive head-switching & Video temporal grounding via causal event modeling. 
        \\
        Apollo\cite{zohar2024apolloexplorationvideounderstanding} & arXiv 2024 & SigLIP-SO400M\cite{zhai2023sigmoid}, InternVideo2\cite{wang2024internvideo2} &  Text: Qwen2.5-7B\cite{bai2023qwen, yang2024qwen2} & Perceiver Resampler \& Token Integration with Timestamps & Video understanding model. 
        \\
        \bottomrule
    \end{tabular}}
    \caption{Summary of latest multimodal video-LLMs and their interaction / fusion mechanisms.}
    \label{tab:llms}
\end{table*}

By addressing these aspects, this paper seeks to advance our understanding of temporal reasoning in LLMs and pave the way for more robust, generalizable, and scalable solutions in video analysis. The insights offered here aim to engage researchers and practitioners alike, highlighting the importance of bridging the gap between static representations and dynamic reasoning in AI systems.

\section{Related Work}

The application of LLMs to video processing has attracted significant attention, owing to their capacity to bridge visual and textual modalities \cite{wang2024visionllm}. In this section, we review related work across four key areas: LLMs for video understanding, pretrained visual encoders, datasets for video understanding, and temporal reasoning in AI systems. We also highlight the distinct contributions of our work compared to prior studies.

\begin{table}[tbp]
\centering
\setlength{\tabcolsep}{0.05em}
\renewcommand{\arraystretch}{0.70}
\resizebox{\columnwidth}{!}{%
\begin{tabular}{lll}
\toprule
\textbf{Type} & \textbf{Visual encoder} & \textbf{Pretrained dataset} \\
\midrule
\multirow{6}{*}{Image} & Normalizer-Free ResNet\cite{brock2021high} & ImageNet-1K\cite{deng2009imagenet} \\
& CLIP ViT\cite{radford2021learning} & WebImageText\cite{10.1145/3404835.3463257} \\
& EVA-CLIP ViT\cite{sun2023eva} & LAION-2B\cite{schuhmann2022laion}, COYO-700M\cite{kakaobrain2022coyo-700m} \\
& BLIP-2 ViT\cite{li2023blip} & WebImageText\cite{10.1145/3404835.3463257} \\
& SigLIP (\eg, ViT)\cite{zhai2023sigmoid} & WebLI\cite{chen2022pali} \\
& OryxViT\cite{liu2024oryx, zhai2023sigmoid} & WebLI\cite{chen2022pali} \\
\midrule
\multirow{9}{*}{Video} & TimeSformer\cite{bertasius2021space} & Kinetics-400\cite{kay2017kinetics}, Kinetics-600\cite{carreira2018short} \\
& I3D\cite{carreira2017quo} & HMDB51\cite{Kuehne11}, UCF101\cite{soomro2012ucf101}, Kinetics-400\cite{kay2017kinetics} \\
& SlowFast\cite{feichtenhofer2019slowfast} & Kinetics-400\cite{kay2017kinetics}, Kinetics-600\cite{carreira2018short}, Charades\cite{sigurdsson2016hollywood} \\
& VideoSwin\cite{liu2022video} & ImageNet-21K\cite{ridnik2021imagenet21k} \\
& UMT\cite{liu2022umt} & Kinetics-400\cite{kay2017kinetics}, AudioSet\cite{gemmeke2017audio} \\
& LanguageBind\cite{zhu2023languagebind} & VIDAL-10M\cite{zhu2023languagebind} \\
& VideoMAE V2\cite{wang2023videomae} & Kinetics-400\cite{kay2017kinetics}, -600\cite{carreira2018short}, -700\cite{carreira2019short}, Something-Something V2\cite{goyal2017something}, AVA\cite{gu2018ava} \\
& InternVL (\eg, InternViT-6B) \cite{chen2024far} & Hybrid image-text datasets (\eg, LAION-en)\cite{cai2022reversible, changpinyo2021conceptual, gu2022wukong, schuhmann2022laion, sharma2018conceptual} \\
& InternVideo2\cite{wang2024internvideo2} & Hybrid video datasets (\eg, Kinetics-400\cite{kay2017kinetics}, InternVid\cite{wang2024internvidlargescalevideotextdataset}) \\
\bottomrule
\end{tabular}%
}
\caption{Visual encoders with their pretrained datasets.}
\label{tab:encoders}
\end{table}

\textbf{LLMs for video understanding.} LLMs have shown remarkable versatility in video-related tasks by incorporating multimodal learning frameworks \cite{wu2023next}. Notable works, such as Flamingo \cite{alayrac2022flamingo} by DeepMind, integrate visual and textual modalities for tasks like video captioning and video question answering (QA). Flamingo uses a cross-modal attention mechanism to align spatiotemporal video embeddings with text representations, showcasing the potential of LLMs in multimodal fusion. Other models \cite{yuan2021florence,girdhar2022omnivore, wang2022omnivl,girdhar2023omnimae}, including OmniVL \cite{wang2022omnivl} and Florence \cite{yuan2021florence}, explore unified architectures that handle images, videos, and text simultaneously, reducing reliance on domain-specific encoders \cite{wang2019hallucinating, wang2021self, wang20213d, koniusz2021tensor, wang2022uncertainty, wang2022temporal, wang20233mformer, wang2024flow,wang2024high,wang2024meet}. However, these works primarily focus on improving task performance without a deep analysis of how LLMs handle temporal dynamics, leaving their capacity for explicit time reasoning largely unexamined.

While prior studies \cite{10.1162/tacl_a_00459,jain2023do,bagad2023test,tan2024are,wang2024towards,gurnee2024language} primarily emphasize task performance, we specifically investigate whether LLMs truly understand the concept of time. Our work explores the interplay between spatiotemporal embeddings and LLM frameworks, providing a deeper analysis of their temporal reasoning capabilities.

\begin{figure*}[tbp]%
\centering
\begin{subfigure}[b]{0.292\linewidth}
\centering\includegraphics[width=\columnwidth]{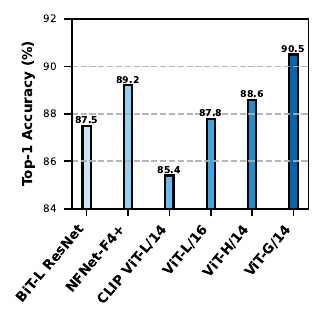}
\caption{\label{fig:image-encoder}Image encoders.}
\end{subfigure}\hfill
\begin{subfigure}[b]{0.705\linewidth}
\centering\includegraphics[width=\textwidth]{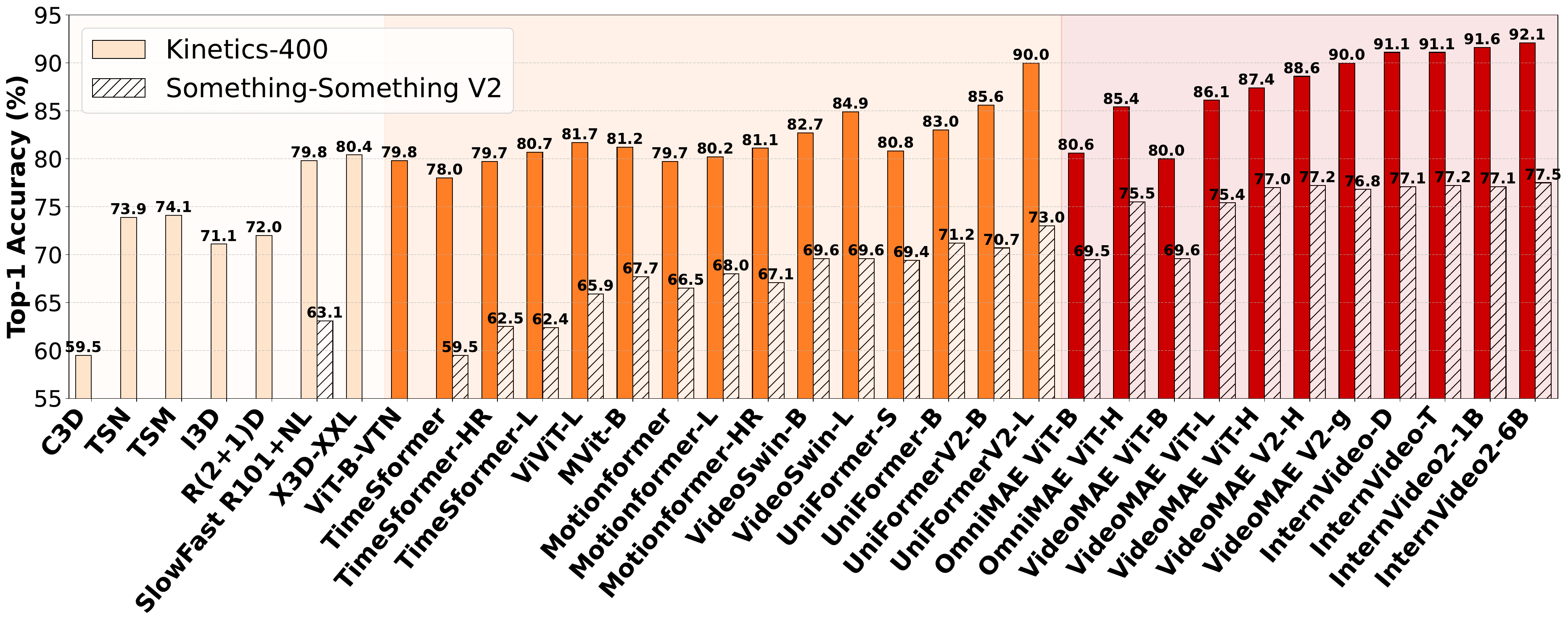}
\caption{\label{fig:video-encoder}Video encoders.}
\end{subfigure}
\caption{Performance comparison of visual encoders. (\textit{Left}): Image classification accuracy for various image encoders pretrained and fine-tuned on the ImageNet-1K dataset. (\textit{Right}): Action recognition accuracy for different video encoders pretrained and fine-tuned on the Kinetics-400 and Something-Something V2 datasets.}
\label{fig:visual-encoder}
\end{figure*}

\textbf{Pretrained visual encoders in multimodal learning.} The success of LLMs in video understanding often hinges on the use of pretrained visual encoders \cite{lin2024vila, chen2024expanding, wang2024mpo, gao2024mini}. Models like CLIP \cite{radford2021learning}, ResNet \cite{he2016deep}, and Vision Transformers (ViT) \cite{dosovitskiy2021an} are frequently used for spatial feature extraction, while video-specific encoders such as I3D \cite{carreira2017quo}, SlowFast \cite{feichtenhofer2019slowfast}, TimeSformer \cite{bertasius2021space}, and Video Swin Transformer \cite{liu2022video} extract spatiotemporal features. These encoders are trained on large-scale datasets like ImageNet \cite{deng2009imagenet,ridnik2021imagenet21k} and Kinetics \cite{kay2017kinetics,carreira2018short, carreira2019short}, enabling them to capture fine-grained features for downstream tasks. Table \ref{tab:encoders} presents an overview of widely used pretrained visual encoders along with their corresponding training datasets. Figure \ref{fig:visual-encoder} compares the performance of visual encoders, showcasing image encoders evaluated on ImageNet-1K\cite{deng2009imagenet} and video encoders assessed on Kinetics-400 and Something-Something V2\cite{goyal2017something}. While the modularity of these encoders facilitates efficient system design, their reliance on general-purpose pretrained features poses limitations, particularly in domain-specific tasks and long-term temporal reasoning \cite{wang2019hallucinating, wang2021self, koniusz2021tensor, qin2022fusing, wang2024flow,wang2024high}. Existing works often treat encoders as static components, overlooking the potential benefits of jointly optimizing encoders and LLMs for temporal understanding \cite{shi2024eagle, wang2023chatvideo}.

Unlike works that use pretrained encoders as black-box components \cite{shu2023audio}, we examine their limitations, including their bias toward short-term dependencies and their challenges in generalizing to abstract temporal concepts. We propose pathways for jointly optimizing encoders and LLMs to address these issues.

\textbf{Datasets for video understanding.} Datasets are fundamental to advancing video understanding, particularly for assessing the temporal reasoning abilities of LLMs \cite{girdhar2019cater, zhou2018temporal, jang2019video}. Current datasets often focus on action recognition, video captioning, or question-answering, capturing spatiotemporal patterns and semantic connections \cite{goyal2017something, jang2017tgif, rohrbach2017movie}. However, many datasets emphasize short-term motions or provide only surface-level annotations, lacking temporal details such as event order, causality, or duration \cite{kuehne2011hmdb, soomro2012ucf101, kay2017kinetics, wang2019vatex}.

While multimodal datasets, combining video with text or audio, offer opportunities for LLMs to align spatiotemporal and semantic reasoning, challenges remain \cite{wang2024comprehensive, fu2024video}. These include limited diversity in scenarios, imprecise annotations, and the difficulty of representing long-term dependencies~\cite{wang2023robust,msad2024}. To truly evaluate and enhance LLMs’ temporal understanding, future datasets must provide richer, more diverse temporal annotations and robust benchmarks across varied domains \cite{nie2024slowfocus}. Table \ref{tab:dataset} provides a comprehensive summary of video datasets used for tasks such as action recognition, video question answering (QA), video captioning, video retrieval, and anomaly detection.

In this work, we emphasize the role of datasets in shaping temporal understanding. We analyze the shortcomings of current video datasets, including their lack of temporal annotations, short-term bias, and limited diversity, and propose directions for improving dataset design.

\begin{table*}[tbp]
\setlength{\tabcolsep}{0.12em}
\renewcommand{\arraystretch}{0.70}
\centering
\resizebox{\linewidth}{!}{%
\begin{tabular}{llcccccll}
\toprule
\textbf{Task} & \textbf{Dataset} & \textbf{Year} & \textbf{Source} & \textbf{\# Videos} & \textbf{Modality} & \textbf{Avg. length (s)} & \textbf{Temporal annotation} & \textbf{Description} \\ 
\midrule
\multirow{13}{*}{Action Recognition} 
& HMDB51\cite{kuehne2011hmdb} & 2011 & YouTube & 6,766 & Video & 3\textasciitilde4 & No & Daily human actions \\
& UCF101\cite{soomro2012ucf101} & 2012 & YouTube & 13,320 & Video+Audio & 7.21 & No & Human actions (\eg, sports, daily activities) \\
& ActivityNet\cite{caba2015activitynet} & 2015 & YouTube & 27,801 & Video+Text & 300\textasciitilde1200 & Temporal extent provided & Human-centric activities \\
& Charades\cite{sigurdsson2016hollywood} & 2016 & Crowdsourced & 9,848 & Video+Text & 30.1 & Start and end timestamps provided & Household activities \\
& Kinetics-400\cite{kay2017kinetics} & 2017 & YouTube & 306,245 & Video & 10 & No & Human actions (\eg, sports, tasks) \\
& AVA\cite{gu2018ava} & 2018 & Movies & 430 & Video & Variable & Start and end timestamps provided & Action localization in movie scenes \\
& Kinetics-600\cite{carreira2018short} & 2018 & YouTube & 392,622 & Video & 10 & No & Human actions (\eg, sports, tasks) \\
& Something-Something V2\cite{goyal2017something} & 2018 & Crowdsourced & 220,847 & Video & 2\textasciitilde6 & Weak & Human-object interactions \\
& EPIC-KITCHENS\cite{Damen2018ScalingEV} & 2018 & Participant kitchens & 432 & Video+Text+Audio & \textasciitilde458 & Start and end timestamps provided & Large-scale egocentric cooking dataset
\\
& COIN\cite{tang2019coin} & 2019 & YouTube & 11,827 & Video+Text & 141.6 & Start and end timestamps provided & Comprehensive instructional tasks (\eg, cooking, repair) \\
& Kinetics-700\cite{carreira2019short} & 2019 & YouTube & 650,317 & Video & 10 & No & Expanded version of Kinetics-400 and Kinetics-600 \\
& EPIC-KITCHENS-100\cite{damen2022rescaling} & 2020 & Participant kitchens & 700 & Video+Text+Audio & \textasciitilde514 & Start and end timestamps provided & Large-scale egocentric cooking dataset
\\
& Ego4D\cite{grauman2022ego4d} & 2021 & Wearable Cameras & 3,850 hours & Video+Text+Audio & Variable & Start and end timestamps provided & First-person activities and interactions \\
& VidSitu\cite{sadhu2021visual} & 2021 & YouTube & 29,000 & Video+Text & \textasciitilde10 & Temporal extent for events provided & Event-centric and causal activity annotations \\
\midrule
\multirow{11}{*}{Video QA} 
& MovieQA\cite{tapaswi2016movieqa} & 2016 & Multiple platforms
& 408 & Video+Text & 202.7 & Start and end timestamps provided & QA for movie scenes \\
& TGIF-QA\cite{jang2017tgif} & 2016 & Tumblr GIFs & 56,720 & Video+Text & 3\textasciitilde5 & Action timestamps provided & QA over social media GIFs \\
& MSVD-QA\cite{xu2017video} & 2017 & YouTube & 1,970 & Video+Text & 27.5 & Start and end timestamps provided & QA for actions description \\
& MSRVTT-QA\cite{xu2017video} & 2017 & YouTube & 10,000 & Video+Text & 15\textasciitilde30 
& Weak & QA across diverse scenes \\
& TVQA\cite{lei2018tvqa} & 2019 & TV Shows & 21,793 & Video+Text & 60\textasciitilde90 & Start and end timestamps provided & QA over medical dramas, sitcoms, crime shows \\
& ActivityNet-QA\cite{yu2019activitynet} & 2019 & YouTube & 5,800 & Video+Text & 180 & Implicit (derived from ActivityNet) & QA for human-annotated videos \\
& How2QA\cite{li2020hero} & 2020 & HowTo100M (YouTube) & 22,000 & Video+Text & 60 & Temporal extent provided & QA over instructional videos \\
& YouCookQA\cite{wang2021make} & 2021 & YouCook2 (YouTube) & 2,000 & Video+Text & 316.2 & Temporal boundaries provided & Cooking-related instructional QA \\
& STAR\cite{wu2024star} & 2021 & Human activity datasets & 22,000 & Video+Text & Variable & Action-level boundaries provided & QA over human-object interactions \\
& MVBench\cite{li2024mvbench} & 2023 & Public datasets & 3,641 & Video+Text & 5\textasciitilde35 & Start and end timestamps provided & Multi-domain QA (\eg, sports, indoor scenes) \\
& EgoSchema\cite{mangalam2023egoschema} & 2023 & Ego4D (Wearable Cameras) & 5,063 & Video+Text & 180 & Timestamped narrations provided & Long-form egocentric activities \\
\midrule
\multirow{6}{*}{Video Captioning} 
& YouCook\cite{das2013thousand} & 2013 & YouTube & 88 & Video+Text & 180\textasciitilde300 & Weak & Cooking instructional videos \\
& MSR-VTT\cite{xu2016msr} & 2016 & YouTube & 7,180 & Video+Text+Audio & 10\textasciitilde30 & Weak & General scenarios (\eg, sports, transport) \\
& ActivityNet Captions\cite{krishna2017dense} & 2017 & YouTube & 20,000 & Video+Text & 180 & Start and end timestamps provided & Dense captions for human-centered activities \\
& VATEX\cite{wang2019vatex} & 2019 & YouTube & 41,250 & Video+Text & \textasciitilde10 & Weak & Multilingual descriptions with English-Chinese parallel captions \\
& HowTo100M\cite{miech2019howto100m} & 2019 & YouTube & 1.22M & Video+Text+Audio & 390 & Subtitle timestamps provided & Instructional video captions \\
& TVC\cite{lei2020tvr} & 2020 & TV Shows & 108,965 & Video+Text & 76.2 & Start and end timestamps provided & Multimodal video captioning dataset \\
\midrule
\multirow{6}{*}{Video Retrieval} 
& LSMDC\cite{rohrbach2017movie} & 2015 & Movies & 118,114 & Video+Text & 4.8 & Start and end timestamps provided & Large-scale dataset for movie description tasks \\
& DiDeMo\cite{hendricks2017localizingmomentsvideonatural} & 2017 & Flickr (YFCC100M) & 10,464 & Video+Text & 27.5 & Start and end timestamps provided & Moment localization in diverse, unedited personal videos \\
& FIVR-200K\cite{kordopatis2019fivr} & 2019 & YouTube & 225,960 & Video & \textasciitilde120 & Start and end timestamps provided & Large-scale incident video retrieval dataset with diverse news events \\
& TVR\cite{lei2020tvr} & 2020 & TV Shows & 21,793 & Video+Text & 76.2 & Start and end timestamps provided & Video-subtitle multimodal moment retrieval dataset \\
& TextVR\cite{wu2023largecrossmodalvideoretrieval} & 2023 & YouTube & 10,500 & Video+Text & 15 & Weak & Cross-modal video retrieval with text reading comprehension \\
& EgoCVR\cite{hummel2024egocvr} & 2024 & Ego4D & 2,295 & Video+Text & 3.9\textasciitilde8.1 & Weak & Egocentric dataset for fine-grained composed video retrieval 
\\
\midrule
\multirow{7}{*}{Anomaly Detection} 
& Subway Entrance\cite{4407716} & 2008 & Surveillance cameras & 1 & Video & 4,800 & No & Crowd monitoring for unusual event detection at subway entrances \\
& Subway Exit\cite{4407716} & 2008 & Surveillance cameras & 1 & Video & 5,400 & No & Crowd monitoring for unusual event detection at subway exits \\
& CUHK Avenue\cite{6751449} & 2013 & Surveillance cameras & 15 & Video & 120 & No & Urban avenue scenes with anomalies like running, loitering, \etc \\
& Street Scene\cite{ramachandra2020streetscenenewdataset} & 2020 & Urban street surveillance & 81 & Video & 582 & Spatial and temporal bounding boxes & Urban street anomalies, \eg, jaywalking, loitering, illegal parking, \etc \\
& XD-Violence\cite{wu2020looklistenlearningmultimodal} & 2020 & Movies and in-the-wild scenes & 4,754 & Video+Audio & \textasciitilde180 & Start and end timestamps provided
& Multimodal violence detection covering six violence types \\
& CUVA\cite{du2024uncoveringwhathowcomprehensive} & 2024 & YouTube, Bilibili & 1,000 & Video+Text & \textasciitilde117 & Start and end timestamps provided & Causation-focused anomaly understanding across 42 anomaly types \\
& MSAD\cite{msad2024} & 2024 & Online Surveillance & 720 & Video & \textasciitilde20 & Frame-level annotations in test set & Multi-scenario dataset with 14 scenarios \\
\midrule
\multirow{2}{*}{Multimodal video tasks} 
& VIDAL-10M\cite{zhu2023languagebind} & 2023 & Multiple platforms 
& 10M & Video+Infrared+Depth+Audio+Text & \textasciitilde20 & Weak & Multi-domain retrieval dataset \\
& Video-MME\cite{fu2024video} & 2024 & YouTube & 900 & Video+Text+Audio & 1017.9 & Temporal ranges via certificate length & Comprehensive evaluation benchmark across many domains \\
\bottomrule
\end{tabular}%
}
\caption{Comprehensive overview of video datasets across tasks.}
\label{tab:dataset}
\end{table*}

\textbf{Temporal reasoning in video AI.} Temporal reasoning is critical for tasks such as action recognition, video summarization, and temporal event ordering \cite{wang2021analysis,wang2019loss,wang2019hallucinating, wang2021self, wang20213d, koniusz2021tensor, wang2022uncertainty, wang2022temporal, qin2022fusing, wang20233mformer, wang2024flow,wang2024high,wang2024meet, raj2024tracknetv4, chen2024motion,chen2024spatial,chen2024sato,msad2024,ding2024lego}. Classical approaches rely on recurrent neural networks (RNNs) \cite{sherstinsky2020fundamentals}, 3D convolutional networks (3D-CNNs) \cite{tran2015learning,carreira2017quo}, and attention mechanisms \cite{NIPS2017_3f5ee243,chen2024motion,wang20233mformer} to model temporal dependencies. More recently, transformers and temporal tokenization techniques have been explored for long-term video understanding \cite{bertasius2021space,patrick2021keeping}. Despite these advances, the explicit modeling of abstract temporal concepts such as causality, sequence progression, and event duration remains underexplored. LLM-based approaches typically use spatiotemporal embeddings from pretrained encoders but fall short in demonstrating robust temporal reasoning capabilities, especially for complex, real-world video scenarios \cite{wang2023robust,msad2024}.

While many studies report empirical results \cite{10.1162/tacl_a_00459,jain2023do,bagad2023test,tan2024are,wang2024towards,gurnee2024language,liu2024tempcompass}, we go further by offering actionable insights for advancing LLM-based temporal reasoning. These include joint training approaches, better dataset curation, and innovative multimodal fusion techniques. 
By addressing these gaps, our work not only advances the understanding of temporal reasoning in LLMs but also provides a roadmap for future research, distinguishing it from prior efforts\cite{tang2023video, nguyen2024video, zhou2024survey} in this rapidly evolving field.

\section{Analysis and Discussion}

\textbf{Can LLMs understand the concept of time?} LLMs exhibit several strengths in processing temporal information. Trained on textual data containing narrations or instructions, they can infer temporal relationships through contextual cues such as ``first'', ``then'', and ``after''. When paired with video encoders, LLMs can process spatiotemporal embeddings, enabling tasks like action recognition and temporal event ordering.

However, LLMs lack direct temporal awareness. Standard models do not inherently model the flow of time unless explicitly trained on sequential video data. Instead, they rely on external encoders to provide temporal structure. Capturing long-term dependencies over extended video sequences is another challenge, as LLMs often operate on tokenized inputs within limited context windows. Additionally, video encoders like 3D CNNs \cite{tran2015learning, carreira2017quo} or video transformers \cite{patrick2021keeping,wang2023videomae}, which act as the ``eyes'' of the system, may excel in capturing motion patterns but struggle to generalize abstract temporal concepts like causality or duration.

A significant limitation lies in the representation of visual time. Unlike textual representations, visual cues require explicit modeling of motion and event transitions. This ambiguity underscores the need for improved temporal modeling in both encoders and LLMs.

\textbf{LLMs applied to videos using pretrained visual encoders.} Most existing approaches use pretrained image or video encoders to extract visual or spatiotemporal information, rather than designing entirely new encoders (Table \ref{tab:llms} and \ref{tab:encoders}). Pretrained image encoders such as CLIP \cite{radford2021learning}, ResNet \cite{he2016deep}, and ViT \cite{NIPS2017_3f5ee243} excel in capturing spatial information, while video encoders like I3D \cite{carreira2017quo}, SlowFast \cite{feichtenhofer2019slowfast}, TimeSformer \cite{bertasius2021space}, and Video Swin Transformer \cite{liu2022video} are widely used for spatiotemporal feature extraction. These encoders, trained on large-scale datasets such as ImageNet \cite{deng2009imagenet,ridnik2021imagenet21k} or Kinetics \cite{kay2017kinetics,carreira2017quo,carreira2019short}, are adept at learning rich feature representations, which can then be fine-tuned for specific tasks \cite{wang2019hallucinating, wang2021self, koniusz2021tensor, wang2023robust, wang2024flow, wang2024high,  chen2024motion, ding2024lego, msad2024, wangtaylor, chen2024spatial}. Notably, video encoders incorporate mechanisms to model temporal dependencies, such as optical flow \cite{wang2024flow}, Taylor videos \cite{wangtaylor}, and motion tracking \cite{chen2024motion, raj2024tracknetv4}, addressing a challenge that LLMs alone cannot handle effectively. Figure \ref{fig:visual-encoder} illustrates the performance of popular visual encoders on two key tasks: image classification (ImageNet-1K) and video action recognition (Kinetics-400 and Something-Something V2).

The use of pretrained visual encoders offers practical advantages. These encoders are optimized for handling visual and spatiotemporal features, reducing computational overhead and enabling faster convergence. Their modular design also allows them to function as ``plug-and-play'' components, seamlessly integrating with LLMs for multimodal learning. This modularity ensures that LLM-based frameworks remain adaptable and scalable across diverse applications.
However, pretrained encoders are not without limitations. First, encoders trained on general datasets may underperform on domain-specific video tasks \cite{msad2024, chen2024spatial}. Second, many pretrained encoders prioritize spatial over temporal information, necessitating additional modules, such as temporal transformers, to capture complex temporal dynamics \cite{chen2024motionllm, zhang2024holmes}. Lastly, large-scale video datasets required for training such encoders are costly to annotate, limiting their ability to encompass diverse or abstract video content.

Some recent efforts, such as DeepMind’s Flamingo \cite{alayrac2022flamingo} and unified architectures like Florence \cite{yuan2021florence} and OmniVL \cite{wang2022omnivl}, explore custom encoders optimized for multimodal learning. These models aim to balance performance across multiple modalities (image, video, and text) without relying heavily on separate pretrained components \cite{wang2019hallucinating,wang2021self,wang2023robust}.

\textbf{How encoders and LLMs interact?} Encoders play a crucial role in preprocessing video frames or sequences to extract visual or spatiotemporal features, which are then transformed into a format compatible with LLMs, often as token embeddings. For example, video transformers process sequences of video frames, while text encoders like CLIP encode textual inputs \cite{yang2023vid2seq, Zhang2023VideoLLaMAAI, ataallah2024minigpt4}. These features are projected into a shared representation space to align different modalities, such as visual and textual data.

\begin{figure}[tbp]%
\centering
\begin{subfigure}[b]{0.56\linewidth}
\centering\includegraphics[width=\linewidth]{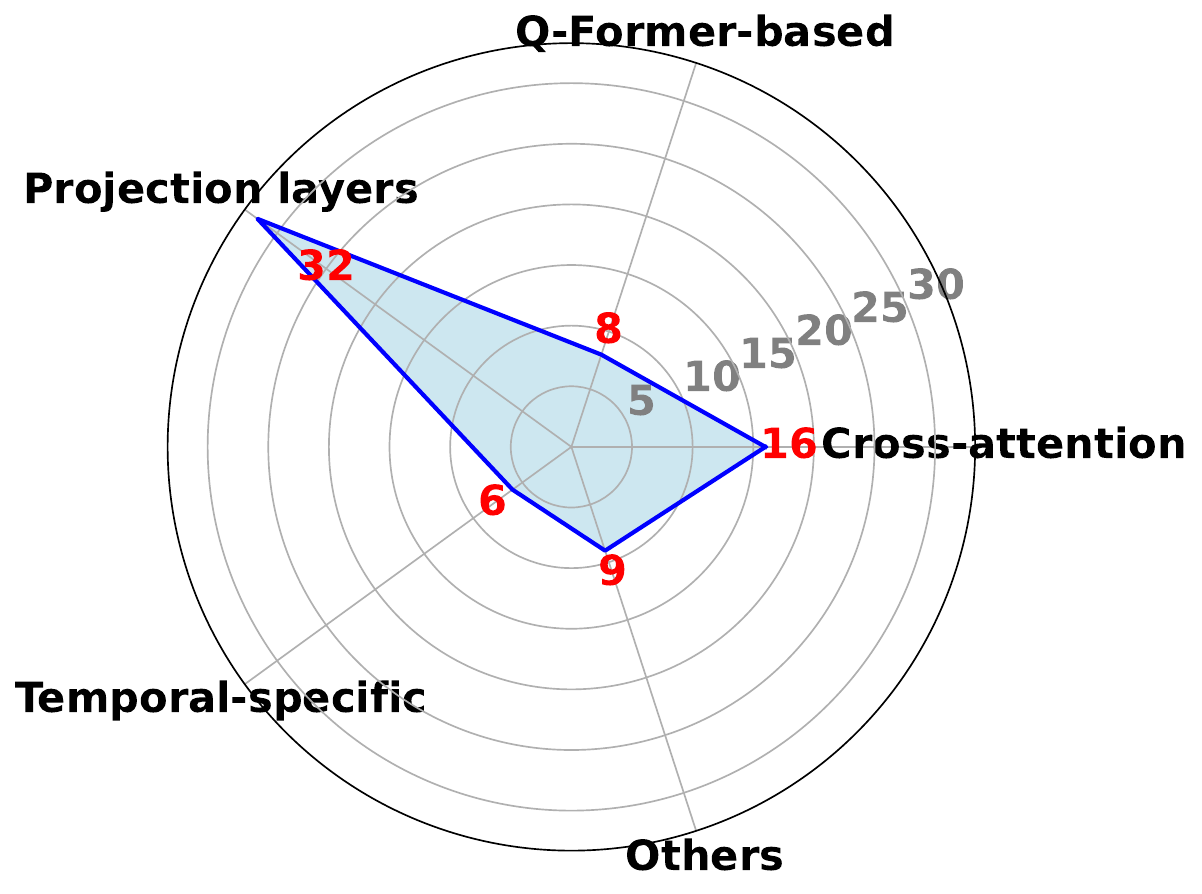}
\caption{\label{fig:interact}Interaction/fusion.}
\end{subfigure}\hfill
\begin{subfigure}[b]{0.43\linewidth}
\centering\includegraphics[width=\linewidth]{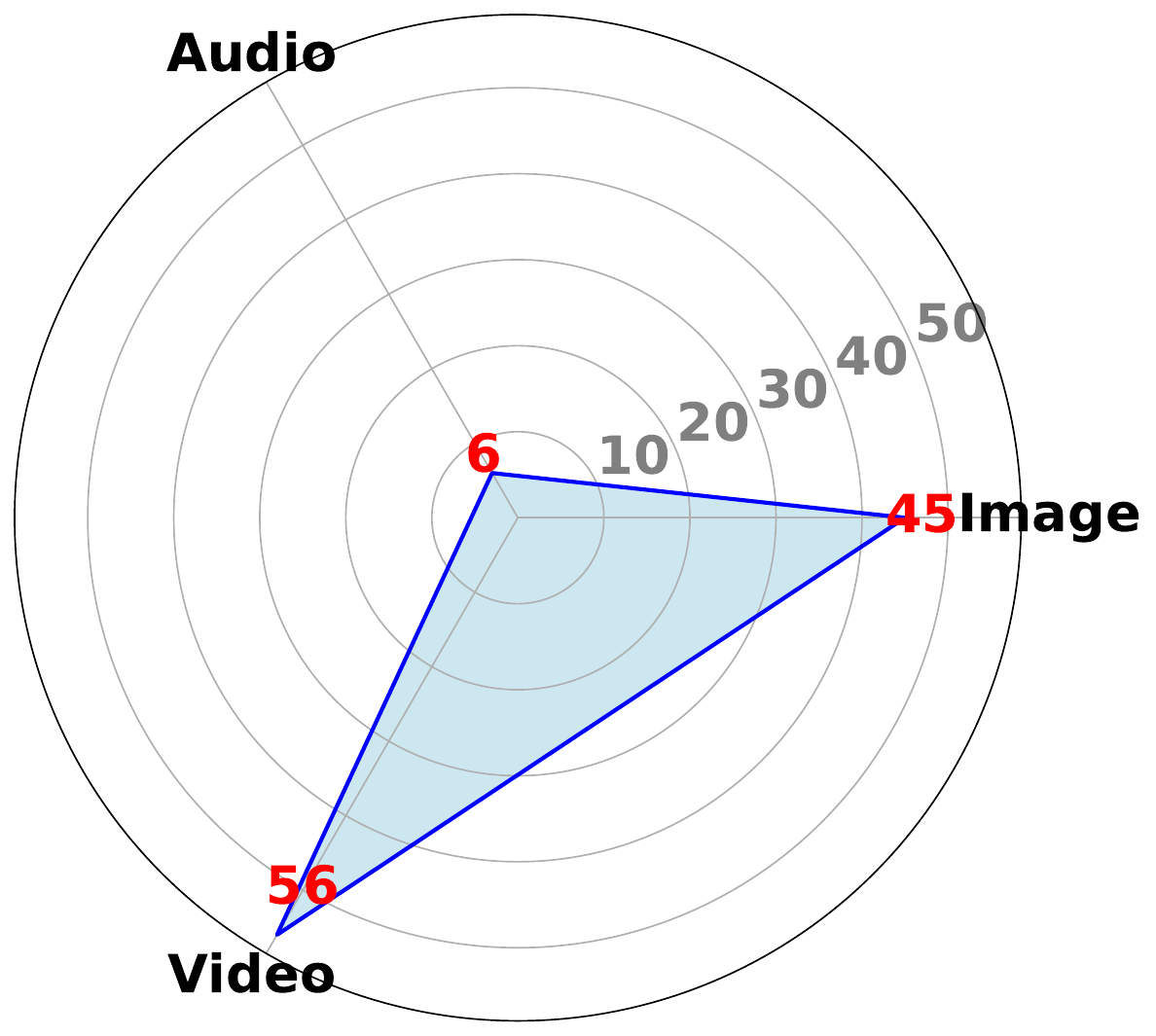}
\caption{\label{fig:modality}Modalities.}
\end{subfigure}
\caption{The distributions of interaction/fusion mechanisms and data modalities in 66 closely related video-LLMs from January 2024 to December 2024. (\textit{Left}): Fusion mechanisms are classified into five categories: Cross-attention (\eg, cross-attention modules, gated cross-attention), Projection layers (\eg, linear projection, MLP projection), Q-Former-based methods (\eg, Q-Former aligns multi-modal embeddings, Trainable Q-Former), Motion/Temporal-Specific mechanisms (\eg, temporal samplers, scene-level sequential alignment), and Other Methods (\eg, Tracklet-centric, Perceiver Resampler, MQ-Former). (\textit{Right}): The distribution of data modalities used in these video-LLMs, with text modalities appearing across all models. Note that a model may use multiple fusion methods and/or data modalities.}
\label{fig:interact-distr}
\end{figure}

Fusion mechanisms are pivotal in enabling LLMs to process multimodal inputs. Encoded features are tokenized and treated as input tokens for the LLM. Attention mechanisms, such as cross-modal attention \cite{zhao2023learning, chen2023vast}, are then used to integrate and relate features from different encoders. Positional embeddings encode spatial and temporal positions of video features, helping the LLM model sequence dependencies. For instance, cross-modal transformers align visual and textual representations, while LLMs refine these embeddings to generate outputs like action labels, video descriptions, or anomaly detection results \cite{fei2024video}. 

Table \ref{tab:llms} highlights the interaction/fusion mechanisms used in the latest video-LLMs under the `Interaction / Fusion mechanism' column. Figure \ref{fig:interact} visualizes the distribution of various interaction and fusion techniques, while Figure \ref{fig:modality} showcases the modalities used in closely related works from 2024 (notably, text modalities are consistently used across these studies). As shown in these plots, the majority of works use projection layers (\eg, linear projection \cite{li2024mvbench}, MLP projection \cite{zhang2024analyzing,zhang2024task,fu2024vita}, semantic translators \cite{chen2023videollm, chen2024motionllm}) and cross-attention mechanisms (\eg, cross-modal attention \cite{yang2023vid2seq}, gated cross-attention \cite{wang2024cosmo}) to facilitate interaction between encoders and LLMs. Temporal-specific mechanisms, such as temporal sampler \cite{zhang2024holmes} and scene-level sequential alignment \cite{yang2024vript}, are also used. Emerging video-LLMs are beginning to explore novel fusion mechanisms, including tracklet-centric approach \cite{wang2023chatvideo}, Perceiver resamplers \cite{alayrac2022flamingo, zohar2024apolloexplorationvideounderstanding}, and MQ-Former \cite{wang2024omnivid}.

However, several challenges arise in this interaction. Encoders must output features in a format compatible with LLMs, requiring careful dimension alignment and embedding space mapping. Additionally, processing video data generates a substantial volume of information, which can strain the LLM’s capacity. Capturing long-term dependencies across extended video sequences also remains a challenge, even with support from pretrained encoders.

\textbf{Bridging the gap between raw video data and temporal reasoning.} The interaction between LLMs and encoders helps bridge the gap between raw video data and higher-level temporal reasoning. Pretrained encoders extract spatiotemporal embeddings that encapsulate low-level motion cues, such as velocity \cite{wangtaylor} and trajectory \cite{patrick2021keeping}, as well as higher-level temporal patterns like scene transitions and sequence progression. These embeddings provide the foundation for LLMs to interpret complex temporal concepts such as causality, event progression, and anticipation.

LLMs use attention mechanisms to prioritize spatiotemporal features, aligning them with semantic or task-specific contexts. Encoders often supply frame-level features as temporal tokens, enabling the LLM to model dependencies and transitions across frames. However, significant challenges remain in achieving comprehensive temporal understanding. Encoders frequently focus on short-term motion patterns, neglecting long-term dependencies \cite{wang20213d,wang2022temporal, wang2024meet}. Similarly, datasets used for encoder pretraining often lack diversity and fail to represent abstract temporal relationships effectively \cite{wang2023robust}. Table \ref{tab:dataset} provides a comparison of datasets based on average video length, data source, modalities, and the number of videos.

LLMs also face inherent limitations. Temporal embeddings require careful tokenization to preserve sequence information when input into LLMs (see Table \ref{tab:llms}). Additionally, LLMs pretrained on static datasets such as text or images may lack the dynamic reasoning capabilities required for video tasks \cite{han2023autoad}. Combining spatial, temporal, and semantic information without losing critical cues remains a complex challenge \cite{chen2024spatial, chen2024motion, raj2024tracknetv4}, further compounded by the computational expense of processing long video sequences while retaining both global and local temporal details.

\begin{figure}[tbp]
    \centering
    \includegraphics[width=\linewidth]{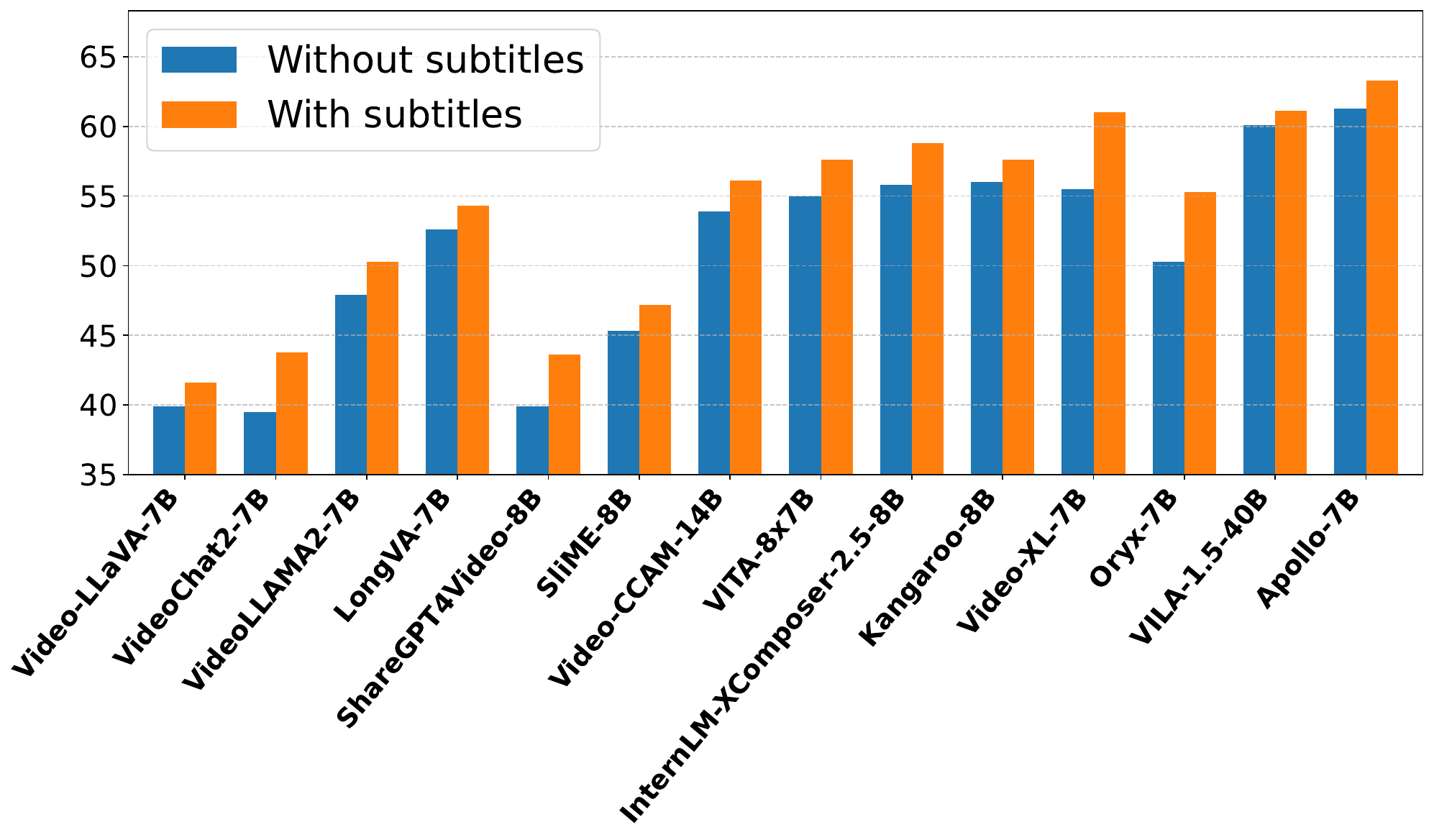}
    \vspace{-0.8cm}
    \caption{Performance (accuracy) comparison of recent video-LLMs on the Video-MME benchmark.} 
    \label{fig:multimodal}
\end{figure}

\textbf{Video datasets: an enabler or bottleneck?} Video datasets are foundational to LLMs on video tasks, yet they often act as a bottleneck \cite{wang2023robust}. Action recognition datasets, such as Kinetics \cite{kay2017kinetics, carreira2018short} and Something-Something V2 \cite{goyal2017something}, are effective for analyzing short-term motion patterns but lack the temporal annotations necessary for reasoning about action sequences or causal relationships. Compared to the Something-Something V2 dataset, the Kinetics datasets exhibit a stronger spatial bias, as demonstrated in \cite{patrick2021keeping,girdhar2022omnivore,girdhar2023omnimae}. Similarly, video QA datasets like TVQA \cite{lei2018tvqa} and How2QA \cite{li2020hero} align well with LLM architectures but often rely on scripted scenarios that limit generalizability to real-world tasks.

Video captioning datasets like MSR-VTT \cite{xu2016msr} and ActivityNet Captions \cite{krishna2017dense} enable multimodal learning by fusing video and text embeddings. However, captions are often superficial and fail to probe deeper temporal reasoning. Long-term video understanding datasets, such as Ego4D \cite{grauman2022ego4d} and Charades \cite{sigurdsson2016hollywood}, focus on extended activities and interactions, offering a richer testing ground for temporal reasoning. Nonetheless, LLMs often struggle with the scale and complexity of such datasets, given their limited context windows.

Multimodal datasets like HowTo100M \cite{miech2019howto100m} and COIN \cite{tang2019coin} align video content with auxiliary modalities, providing opportunities for pretraining in a multimodal setup. However, the inherent noise and lack of temporal annotations in these datasets can hinder performance. To advance temporal understanding in LLMs, datasets must evolve to include richer temporal annotations, long-term dependencies, and diverse real-world scenarios. The latest multimodal datasets include VIDAL-10M \cite{zhu2023languagebind} and Video-MME \cite{fu2024video}.

Table \ref{tab:dataset} summarizes popular datasets used across various video tasks, including action recognition, anomaly detection, video question answering, captioning, and retrieval, as well as some recent multimodal video understanding datasets.

\begin{figure}[tbp]
    \centering
    \includegraphics[width=\linewidth]{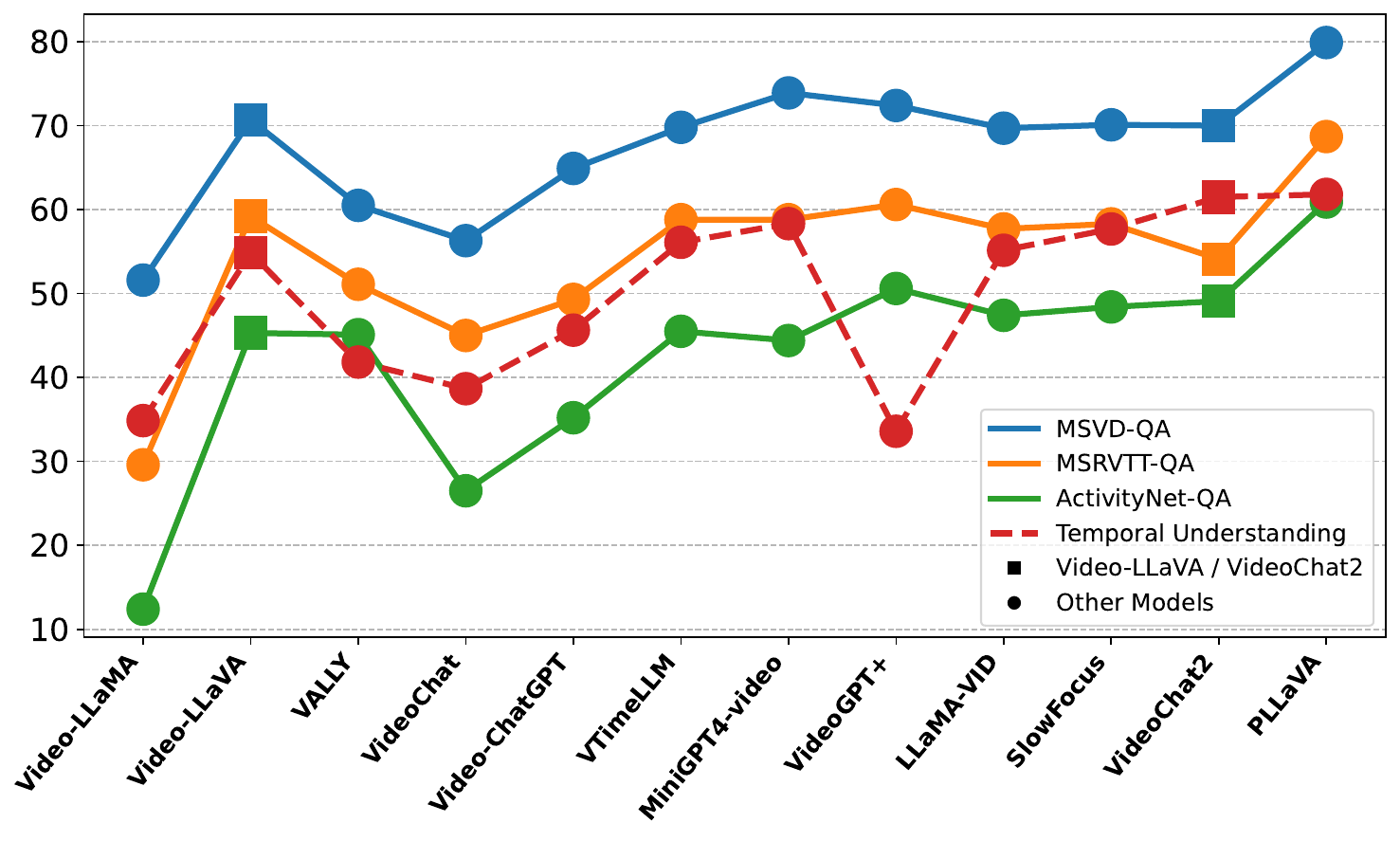}
    \vspace{-0.8cm}
    \caption{Performance comparison of recent video-LLMs on video QA benchmarks. Models using pretrained video encoders (\eg, Video-LLaVA and VideoChat2) are marked with squares, while models using pretrained image encoders are represented by circles.} 
    \label{fig:QAcompare}
\end{figure}

\textbf{State-of-the-Art video LLMs.} Recent advancements in video LLMs have significantly enhanced the processing of spatiotemporal information \cite{chen2024videollm,nie2024slowfocus,li2024mvbench}. Traditional LLMs, focused on textual data, often struggle with temporal dynamics in video \cite{liu2024tempcompass}. Models like VideoChat2 \cite{li2024mvbench} and SlowFocus \cite{nie2024slowfocus} are pushing the boundaries by integrating temporal reasoning with video analysis. VideoChat2 enables real-time multimodal dialogue, processing video sequences to answer questions about actions, events, and causal relationships, making it highly effective for interactive applications. SlowFocus improves fine-grained temporal understanding, capturing long-term dependencies and transitions within video, essential for tasks like video summarization and anomaly detection. TimeSformer \cite{bertasius2021space} further refines this by using attention mechanisms to simultaneously model spatial and temporal features, enhancing video understanding in complex scenarios like action recognition.

\begin{figure}[tbp]%
\centering
\begin{subfigure}[b]{0.495\linewidth}
\centering\includegraphics[width=\linewidth]{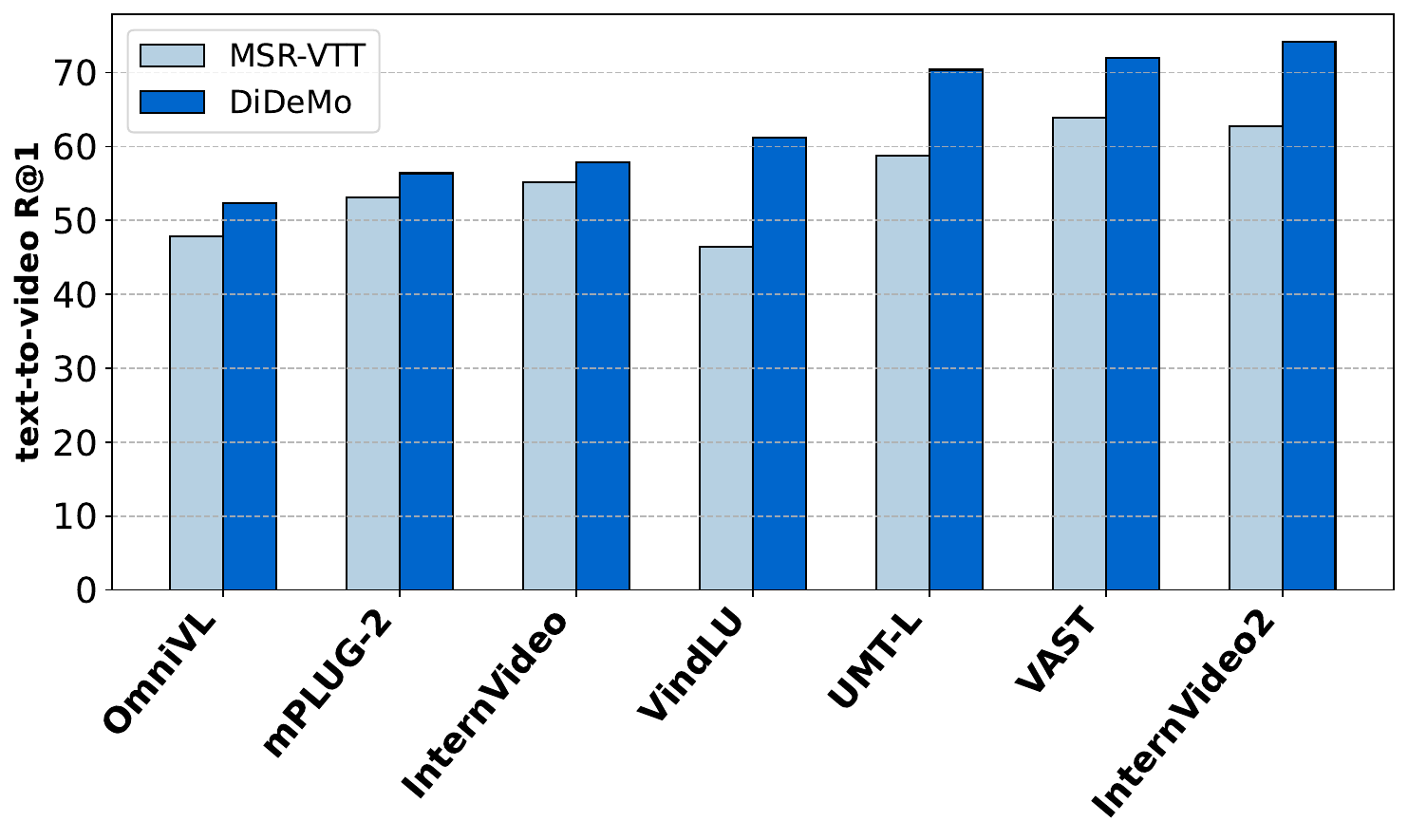}
\caption{\label{fig:caption}Video retrieval.}
\end{subfigure}\hfill
\begin{subfigure}[b]{0.495\linewidth}
\centering\includegraphics[width=\linewidth]{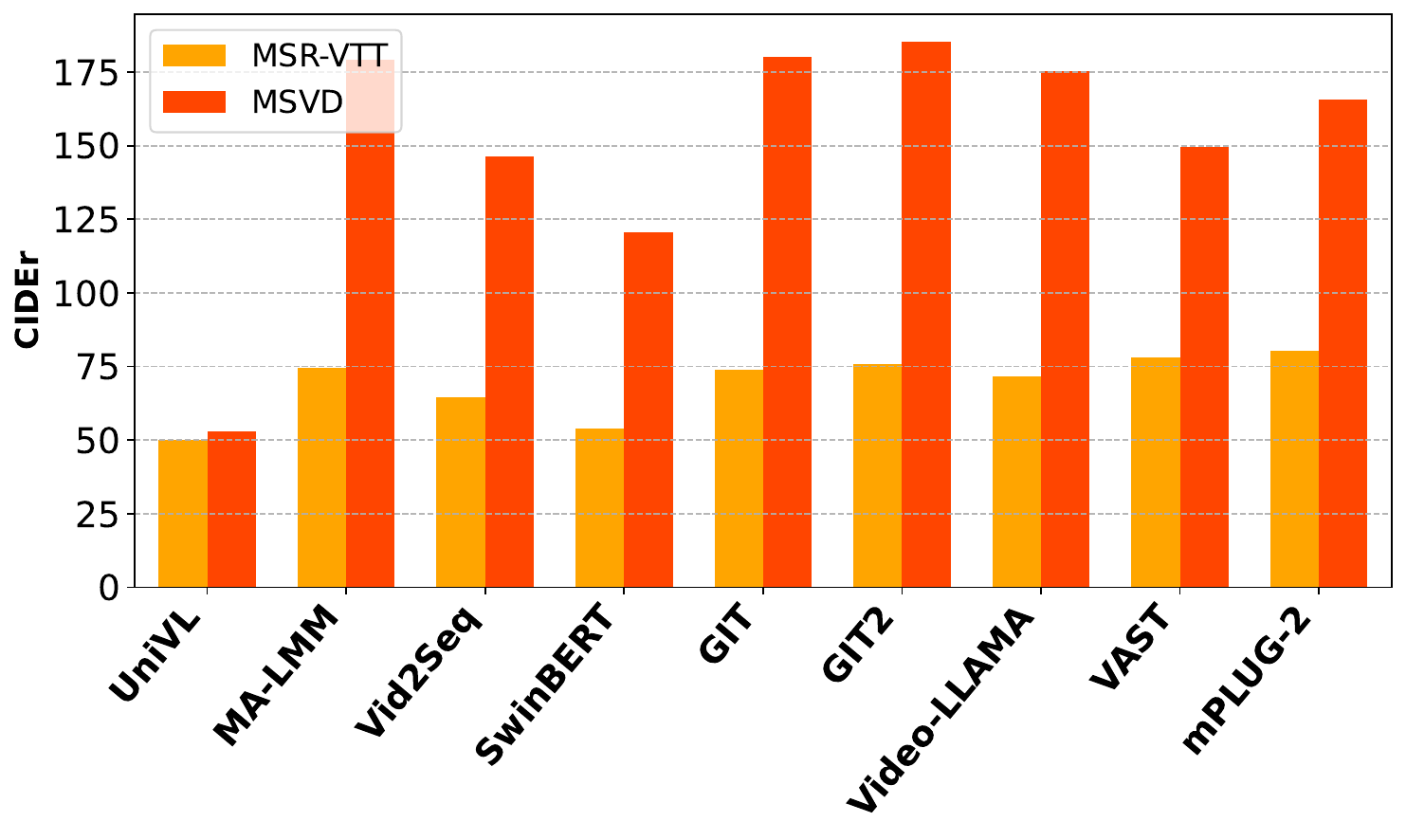}
\caption{\label{fig:retrieval}Video captioning.}
\end{subfigure}
\vspace{-0.3cm}
\caption{Performance comparison of recent video-LLMs on (a) video retrieval and (b) video captioning benchmarks.}
\label{fig:caption-retrieval}
\end{figure}

Additionally, models, \eg, VideoLLM-Online \cite{chen2024videollm} and VideoBERT \cite{sun2019videobert}, focus on continuous learning from video streams, allowing for real-time updates and better adaptability to dynamic content. These models are crucial for applications requiring ongoing video analysis, such as surveillance, event detection, and interactive media. Flamingo \cite{alayrac2022flamingo} takes multimodal learning a step further by combining visual, textual, and temporal data, offering a more holistic approach to video processing. ActionFormer \cite{zhang2022actionformer}, on the other hand, specializes in action recognition through long-range temporal dependencies, making it effective for tasks like sports video analysis and human-computer interaction. These advancements reflect a significant leap in video LLM capabilities, making them better equipped for real-world video understanding, interaction, and analysis.
Figures \ref{fig:multimodal}, \ref{fig:QAcompare}, and \ref{fig:caption-retrieval} present comparisons of recent popular video-LLMs across multiple video tasks, including multimodal video understanding, video QA, video retrieval, and video captioning. As shown in these figures, no single video-LLM excels across all tasks: (i) a comprehensive evaluation system covering all video tasks is lacking, and (ii) most video-LLMs are designed to address only a subset of these challenges.

\textbf{Fair evaluation is needed.} Evaluations and comparisons of video-LLMs are often conducted inconsistently (see Figures \ref{fig:multimodal}, \ref{fig:QAcompare}, and \ref{fig:caption-retrieval}), which can result in unfair assessments and misleading conclusions \cite{tang2023video, nguyen2024video}. A frequent issue is the comparison of video-LLMs, designed for multimodal reasoning, with traditional video models such as I3D \cite{carreira2017quo}, SlowFast \cite{feichtenhofer2019slowfast}, or Video Swin Transformer \cite{liu2022video}, which are tailored for video-specific tasks like action recognition. While traditional models excel at spatiotemporal feature extraction due to their focused design, video-LLMs must simultaneously handle visual and linguistic alignment, which adds inherent complexity to their objectives. Directly comparing video-LLMs against such specialized models is therefore not entirely fair. Instead, video-LLMs should be systematically benchmarked against other video-LLMs or multimodal frameworks to better reflect their relative strengths and limitations in tasks like video action recognition, video captioning, or video QA.

Furthermore, inconsistencies in evaluation practices exacerbate the problem. Different models are often trained and tested on varying datasets, such as Kinetics-400 \cite{kay2017kinetics}, Ego4D \cite{grauman2022ego4d}, or HowTo100M \cite{miech2019howto100m}, without standardized protocols for pretraining, finetuning, or testing. This creates biases in results and hampers fair comparisons. For instance, a video-LLM pretrained on massive multimodal datasets might show superior results simply due to larger data availability rather than architectural improvements. To address this, evaluations must adopt consistent and standardized benchmarks, training splits, and metrics. Frameworks that systematically assess multimodal alignment, temporal reasoning, and downstream task performance would help ensure transparency and comparability.

Finally, establishing fair evaluation practices requires prioritizing within-paradigm comparisons. For video action recognition, for example, models like VideoChat2 \cite{li2024mvbench} and VideoLLM-online \cite{chen2024videollm} should be compared against each other rather than with traditional video-only transformers. This approach highlights progress within the multimodal video understanding space and reveals areas for improvement, such as better temporal consistency or more efficient multimodal alignment. By addressing these challenges, fair and systematic evaluation will provide deeper insights into video-LLM capabilities and foster future advancements in the field.

\textbf{Factors driving superior performance in video LLMs.} The superior performance of certain video LLMs can be attributed to their ability to effectively integrate spatial, temporal, and semantic information, often through advanced architectures and training strategies. Models like SlowFocus \cite{nie2024slowfocus} and VideoChat2 \cite{li2024mvbench} excel by incorporating fine-grained temporal reasoning and long-range dependencies, which are crucial for understanding complex video dynamics such as event progression and causal relationships. 
The use of hierarchical or multi-level attention mechanisms, seen in models like TimeSformer \cite{bertasius2021space} and VideoLLM-Online \cite{chen2024videollm}, enables them to capture both short-term motions and long-term narrative structures, addressing the temporal gaps that limit earlier models. 

Additionally, models that use large-scale, multimodal pretraining on diverse video datasets, such as Flamingo, benefit from the cross-domain knowledge transfer between visual, textual, and temporal modalities, leading to a more holistic understanding of video content. These innovations enable the models to generalize better across different video tasks, including action recognition, video captioning, and dynamic scene interpretation.

\section{Future Directions}

Building on the preceding analysis and discussion, we outline below several promising future research directions for those interested in advancing video LLMs.

\textbf{Overcoming dataset challenges for LLMs.}
Datasets remain a critical bottleneck in advancing LLM-based video systems. Addressing their limitations requires both creative solutions and resource investments:

\renewcommand{\labelenumi}{\roman{enumi}.}
\begin{enumerate}[leftmargin=0.4cm]
\item Temporal annotations and structure: Enriching datasets with temporal annotations, such as event order, duration, and causal relationships, is essential. Techniques like crowd-sourced annotations, AI-assisted labeling, or synthetic data generation could address the scarcity of such datasets.
\item Balancing scale and quality: While large-scale datasets like HowTo100M offer broad coverage, they often include noise and irrelevant data. Future efforts should focus on curating high-quality, balanced datasets that prioritize diversity and accuracy without sacrificing scale. Semi-supervised and unsupervised learning approaches could also mitigate the need for large annotated datasets.
\item Addressing short-term bias: The dominance of short video clips in existing datasets limits models’ ability to reason over extended sequences. Introducing datasets with long-term dependencies, such as episodic or procedural videos, would help train models to understand overarching narratives and transitions.
\item Expanding dataset diversity: Current datasets often focus on narrow domains, such as sports or cooking, which restricts model generalizability \cite{wang2023robust}. Diverse datasets encompassing a wider range of cultural contexts, real-world scenarios, and tasks are essential to improve performance across applications. Collaborative initiatives between academic and industry stakeholders could accelerate the development of such datasets.
\item Improving multimodal alignment: Misalignment between video frames and associated text or audio annotations introduces inconsistencies that can degrade learning quality. Future work could explore more precise alignment methods \cite{wang20213d, wang2022temporal, wang2022uncertainty, wang2024meet}, such as neural alignment models or reinforcement learning frameworks, to ensure that temporal and semantic signals are accurately correlated during training.
\end{enumerate}

\textbf{Enhancing temporal understanding.} To improve temporal reasoning, joint training of encoders and LLMs is a promising direction. Models co-trained on datasets with temporal reasoning tasks can develop a deeper understanding of complex time-related concepts such as causality, event sequencing, and duration. Architectures like temporal transformers, recurrent neural networks, or hybrid systems that combine hierarchical and sequential processing should be further explored to handle both short-term dynamics and long-term dependencies in video data.

Explicit supervision for abstract temporal concepts through enriched annotations is another critical step \cite{ding2024lego}. Annotated datasets with detailed temporal labels, covering relationships, transitions, and event causality, can significantly boost the temporal reasoning capacity of these systems. Furthermore, multimodal training on datasets that integrate video, text, audio, and temporal metadata could align spatiotemporal and semantic representations more effectively, enhancing the model’s ability to reason about time in real-world scenarios \cite{ding2024lego, raj2024tracknetv4, msad2024}.

\textbf{Multimodal LLMs for holistic temporal reasoning.} The development of truly multimodal LLMs capable of holistic temporal reasoning requires a synergistic interplay between spatiotemporal and semantic understanding. Future research should explore adaptive attention mechanisms that dynamically weigh temporal, spatial, and textual information based on context. Advances in memory-efficient architectures and progressive learning strategies could enable LLMs to process longer sequences without losing critical details.

Additionally, fine-tuning LLMs on domain-specific datasets or tasks, such as medical video analysis or video-based social interaction studies \cite{christophe2024med42, jeong2024fine}, could expand their applicability and temporal reasoning depth. Transfer learning approaches \cite{zhao2024expel}, where models pretrained on general datasets are fine-tuned for specific temporal reasoning tasks, can also be effective in reducing resource demands.

Other modalities, such as depth videos \cite{wang2021analysis,wang2019comparative} and motion-specific data like skeletons (useful for analyzing human-related movements) \cite{wang20213d, wang2022temporal, wang2022uncertainty, wang20233mformer, wang2024meet}, optical flow \cite{wang2024flow}, and Taylor videos \cite{wangtaylor}, can significantly enhance the performance of LLM frameworks in video processing tasks when their pretrained models are incorporated. Depth videos capture three-dimensional spatial information, offering a richer understanding of scene geometry. Skeleton data focuses on joint movements, making it particularly effective for applications like action recognition and gesture analysis. Optical flow and Taylor videos excel in capturing frame-to-frame motion changes, providing detailed temporal cues essential for understanding dynamic content.

By integrating these diverse modalities, LLMs can achieve a more comprehensive representation of motion dynamics and spatial structures, broadening their applicability to complex video-based challenges. Moreover, the inclusion of learned video motion prompts \cite{chen2024motion}, which highlight relevant movements within a scene, introduces a novel modality that further refines the system’s ability to process and interpret intricate video content.

\textbf{Advancing visual encoders for multimodal learning.} While most current LLM-based video systems use pretrained encoders for their efficiency and robust feature extraction, there is significant potential in designing novel encoders optimized specifically for multimodal learning. These encoders should aim to seamlessly integrate spatial, temporal, and semantic information into a unified framework, reducing the reliance on modular preprocessing steps. Future research could explore adaptive encoder architectures that dynamically adjust to varying video characteristics, such as scene complexity or temporal dynamics. Additionally, encoders tailored for specific domains, like healthcare, education, or autonomous driving, could enhance the accuracy and relevance of multimodal systems in specialized applications.

\textbf{Ethical and practical considerations.} As LLM-based video systems advance, addressing ethical concerns becomes increasingly important \cite{wangtaylor,msad2024}. Ensuring fairness and avoiding biases in datasets, particularly regarding cultural and contextual diversity, will be critical. Practical considerations like energy efficiency and the environmental impact of large-scale training should also guide future research. Lightweight models or distillation techniques could balance performance with computational sustainability.
Addressing these directions, researchers can unlock the full potential of LLMs in video-based applications, driving advancements in temporal reasoning, multimodal understanding, and real-world usability.

\section{Conclusion}

This work critically examines the temporal reasoning capabilities of large language models (LLMs) in video processing, identifying significant limitations in both models and datasets. While LLMs paired with pretrained visual encoders have achieved success in tasks such as action recognition, anomaly detection, and video summarization, they fall short in understanding long-term temporal dependencies. This stems from the encoders' focus on short-term patterns, fragmented temporal cues, and challenges in aligning spatial, temporal, and semantic information. Additionally, existing datasets lack explicit temporal annotations, often focus on short clips over long sequences, and struggle with diversity and multimodal alignment, further hindering progress.

To unlock the full potential of LLMs in video processing, future research must address these gaps. This includes designing integrated frameworks to jointly train encoders and LLMs on temporal reasoning, enriching datasets with detailed annotations for long-term dependencies, and creating innovative architectures that fuse spatiotemporal and semantic information. By addressing these challenges, we can pave the way for systems that not only excel in video analysis but also advance broader applications requiring robust temporal comprehension.

\begin{acks}
Xi Ding, a Research Assistant with the Temporal Intelligence and Motion Extraction (TIME) Lab at ANU, contributed to this work. 
This research was conducted under the supervision of Lei Wang.
\end{acks}

\bibliographystyle{ACM-Reference-Format}
\bibliography{citations}


\begin{thebibliography}{213}


\ifx \showCODEN    \undefined \def \showCODEN     #1{\unskip}     \fi
\ifx \showDOI      \undefined \def \showDOI       #1{#1}\fi
\ifx \showISBNx    \undefined \def \showISBNx     #1{\unskip}     \fi
\ifx \showISBNxiii \undefined \def \showISBNxiii  #1{\unskip}     \fi
\ifx \showISSN     \undefined \def \showISSN      #1{\unskip}     \fi
\ifx \showLCCN     \undefined \def \showLCCN      #1{\unskip}     \fi
\ifx \shownote     \undefined \def \shownote      #1{#1}          \fi
\ifx \showarticletitle \undefined \def \showarticletitle #1{#1}   \fi
\ifx \showURL      \undefined \def \showURL       {\relax}        \fi
\providecommand\bibfield[2]{#2}
\providecommand\bibinfo[2]{#2}
\providecommand\natexlab[1]{#1}
\providecommand\showeprint[2][]{arXiv:#2}

\bibitem[Abdin et~al\mbox{.}(2024)]%
        {abdin2024phi3technicalreporthighly}
\bibfield{author}{\bibinfo{person}{Marah Abdin}, \bibinfo{person}{Jyoti Aneja}, \bibinfo{person}{Hany Awadalla}, \bibinfo{person}{Ahmed Awadallah}, \bibinfo{person}{Ammar~Ahmad Awan}, \bibinfo{person}{Nguyen Bach}, \bibinfo{person}{Amit Bahree}, \bibinfo{person}{Arash Bakhtiari}, \bibinfo{person}{Jianmin Bao}, \bibinfo{person}{Harkirat Behl}, \bibinfo{person}{Alon Benhaim}, \bibinfo{person}{Misha Bilenko}, \bibinfo{person}{Johan Bjorck}, \bibinfo{person}{Sébastien Bubeck}, \bibinfo{person}{Martin Cai}, \bibinfo{person}{Qin Cai}, \bibinfo{person}{Vishrav Chaudhary}, \bibinfo{person}{Dong Chen}, \bibinfo{person}{Dongdong Chen}, \bibinfo{person}{Weizhu Chen}, \bibinfo{person}{Yen-Chun Chen}, \bibinfo{person}{Yi-Ling Chen}, \bibinfo{person}{Hao Cheng}, \bibinfo{person}{Parul Chopra}, \bibinfo{person}{Xiyang Dai}, \bibinfo{person}{Matthew Dixon}, \bibinfo{person}{Ronen Eldan}, \bibinfo{person}{Victor Fragoso}, \bibinfo{person}{Jianfeng Gao}, \bibinfo{person}{Mei Gao}, \bibinfo{person}{Min Gao},
  \bibinfo{person}{Amit Garg}, \bibinfo{person}{Allie~Del Giorno}, \bibinfo{person}{Abhishek Goswami}, \bibinfo{person}{Suriya Gunasekar}, \bibinfo{person}{Emman Haider}, \bibinfo{person}{Junheng Hao}, \bibinfo{person}{Russell~J. Hewett}, \bibinfo{person}{Wenxiang Hu}, \bibinfo{person}{Jamie Huynh}, \bibinfo{person}{Dan Iter}, \bibinfo{person}{Sam~Ade Jacobs}, \bibinfo{person}{Mojan Javaheripi}, \bibinfo{person}{Xin Jin}, \bibinfo{person}{Nikos Karampatziakis}, \bibinfo{person}{Piero Kauffmann}, \bibinfo{person}{Mahoud Khademi}, \bibinfo{person}{Dongwoo Kim}, \bibinfo{person}{Young~Jin Kim}, \bibinfo{person}{Lev Kurilenko}, \bibinfo{person}{James~R. Lee}, \bibinfo{person}{Yin~Tat Lee}, \bibinfo{person}{Yuanzhi Li}, \bibinfo{person}{Yunsheng Li}, \bibinfo{person}{Chen Liang}, \bibinfo{person}{Lars Liden}, \bibinfo{person}{Xihui Lin}, \bibinfo{person}{Zeqi Lin}, \bibinfo{person}{Ce Liu}, \bibinfo{person}{Liyuan Liu}, \bibinfo{person}{Mengchen Liu}, \bibinfo{person}{Weishung Liu}, \bibinfo{person}{Xiaodong Liu},
  \bibinfo{person}{Chong Luo}, \bibinfo{person}{Piyush Madan}, \bibinfo{person}{Ali Mahmoudzadeh}, \bibinfo{person}{David Majercak}, \bibinfo{person}{Matt Mazzola}, \bibinfo{person}{Caio César~Teodoro Mendes}, \bibinfo{person}{Arindam Mitra}, \bibinfo{person}{Hardik Modi}, \bibinfo{person}{Anh Nguyen}, \bibinfo{person}{Brandon Norick}, \bibinfo{person}{Barun Patra}, \bibinfo{person}{Daniel Perez-Becker}, \bibinfo{person}{Thomas Portet}, \bibinfo{person}{Reid Pryzant}, \bibinfo{person}{Heyang Qin}, \bibinfo{person}{Marko Radmilac}, \bibinfo{person}{Liliang Ren}, \bibinfo{person}{Gustavo de Rosa}, \bibinfo{person}{Corby Rosset}, \bibinfo{person}{Sambudha Roy}, \bibinfo{person}{Olatunji Ruwase}, \bibinfo{person}{Olli Saarikivi}, \bibinfo{person}{Amin Saied}, \bibinfo{person}{Adil Salim}, \bibinfo{person}{Michael Santacroce}, \bibinfo{person}{Shital Shah}, \bibinfo{person}{Ning Shang}, \bibinfo{person}{Hiteshi Sharma}, \bibinfo{person}{Yelong Shen}, \bibinfo{person}{Swadheen Shukla}, \bibinfo{person}{Xia Song},
  \bibinfo{person}{Masahiro Tanaka}, \bibinfo{person}{Andrea Tupini}, \bibinfo{person}{Praneetha Vaddamanu}, \bibinfo{person}{Chunyu Wang}, \bibinfo{person}{Guanhua Wang}, \bibinfo{person}{Lijuan Wang}, \bibinfo{person}{Shuohang Wang}, \bibinfo{person}{Xin Wang}, \bibinfo{person}{Yu Wang}, \bibinfo{person}{Rachel Ward}, \bibinfo{person}{Wen Wen}, \bibinfo{person}{Philipp Witte}, \bibinfo{person}{Haiping Wu}, \bibinfo{person}{Xiaoxia Wu}, \bibinfo{person}{Michael Wyatt}, \bibinfo{person}{Bin Xiao}, \bibinfo{person}{Can Xu}, \bibinfo{person}{Jiahang Xu}, \bibinfo{person}{Weijian Xu}, \bibinfo{person}{Jilong Xue}, \bibinfo{person}{Sonali Yadav}, \bibinfo{person}{Fan Yang}, \bibinfo{person}{Jianwei Yang}, \bibinfo{person}{Yifan Yang}, \bibinfo{person}{Ziyi Yang}, \bibinfo{person}{Donghan Yu}, \bibinfo{person}{Lu Yuan}, \bibinfo{person}{Chenruidong Zhang}, \bibinfo{person}{Cyril Zhang}, \bibinfo{person}{Jianwen Zhang}, \bibinfo{person}{Li~Lyna Zhang}, \bibinfo{person}{Yi Zhang}, \bibinfo{person}{Yue Zhang},
  \bibinfo{person}{Yunan Zhang}, {and} \bibinfo{person}{Xiren Zhou}.} \bibinfo{year}{2024}\natexlab{}.
\newblock \bibinfo{title}{Phi-3 Technical Report: A Highly Capable Language Model Locally on Your Phone}.
\newblock
\newblock
\showeprint[arxiv]{2404.14219}~[cs.CL]
\urldef\tempurl%
\url{https://arxiv.org/abs/2404.14219}
\showURL{%
\tempurl}


\bibitem[Adam et~al\mbox{.}(2008)]%
        {4407716}
\bibfield{author}{\bibinfo{person}{Amit Adam}, \bibinfo{person}{Ehud Rivlin}, \bibinfo{person}{Ilan Shimshoni}, {and} \bibinfo{person}{Daviv Reinitz}.} \bibinfo{year}{2008}\natexlab{}.
\newblock \showarticletitle{Robust Real-Time Unusual Event Detection using Multiple Fixed-Location Monitors}.
\newblock \bibinfo{journal}{\emph{IEEE Transactions on Pattern Analysis and Machine Intelligence}} \bibinfo{volume}{30}, \bibinfo{number}{3} (\bibinfo{year}{2008}), \bibinfo{pages}{555--560}.
\newblock
\urldef\tempurl%
\url{https://doi.org/10.1109/TPAMI.2007.70825}
\showDOI{\tempurl}


\bibitem[AI et~al\mbox{.}(2024)]%
        {ai2024yiopenfoundationmodels}
\bibfield{author}{\bibinfo{person}{01. AI}, \bibinfo{person}{:}, \bibinfo{person}{Alex Young}, \bibinfo{person}{Bei Chen}, \bibinfo{person}{Chao Li}, \bibinfo{person}{Chengen Huang}, \bibinfo{person}{Ge Zhang}, \bibinfo{person}{Guanwei Zhang}, \bibinfo{person}{Heng Li}, \bibinfo{person}{Jiangcheng Zhu}, \bibinfo{person}{Jianqun Chen}, \bibinfo{person}{Jing Chang}, \bibinfo{person}{Kaidong Yu}, \bibinfo{person}{Peng Liu}, \bibinfo{person}{Qiang Liu}, \bibinfo{person}{Shawn Yue}, \bibinfo{person}{Senbin Yang}, \bibinfo{person}{Shiming Yang}, \bibinfo{person}{Tao Yu}, \bibinfo{person}{Wen Xie}, \bibinfo{person}{Wenhao Huang}, \bibinfo{person}{Xiaohui Hu}, \bibinfo{person}{Xiaoyi Ren}, \bibinfo{person}{Xinyao Niu}, \bibinfo{person}{Pengcheng Nie}, \bibinfo{person}{Yuchi Xu}, \bibinfo{person}{Yudong Liu}, \bibinfo{person}{Yue Wang}, \bibinfo{person}{Yuxuan Cai}, \bibinfo{person}{Zhenyu Gu}, \bibinfo{person}{Zhiyuan Liu}, {and} \bibinfo{person}{Zonghong Dai}.} \bibinfo{year}{2024}\natexlab{}.
\newblock \bibinfo{title}{Yi: Open Foundation Models by 01.AI}.
\newblock
\newblock
\showeprint[arxiv]{2403.04652}~[cs.CL]
\urldef\tempurl%
\url{https://arxiv.org/abs/2403.04652}
\showURL{%
\tempurl}


\bibitem[AI(2024)]%
        {meta2024llama3}
\bibfield{author}{\bibinfo{person}{Meta AI}.} \bibinfo{year}{2024}\natexlab{}.
\newblock \bibinfo{title}{Introducing Meta Llama 3: The most capable openly available LLM to date}.
\newblock
\newblock
\urldef\tempurl%
\url{https://ai.meta.com/blog/meta-llama-3/}
\showURL{%
\tempurl}


\bibitem[Alayrac et~al\mbox{.}(2022)]%
        {alayrac2022flamingo}
\bibfield{author}{\bibinfo{person}{Jean-Baptiste Alayrac}, \bibinfo{person}{Jeff Donahue}, \bibinfo{person}{Pauline Luc}, \bibinfo{person}{Antoine Miech}, \bibinfo{person}{Iain Barr}, \bibinfo{person}{Yana Hasson}, \bibinfo{person}{Karel Lenc}, \bibinfo{person}{Arthur Mensch}, \bibinfo{person}{Katherine Millican}, \bibinfo{person}{Malcolm Reynolds}, {et~al\mbox{.}}} \bibinfo{year}{2022}\natexlab{}.
\newblock \showarticletitle{Flamingo: a visual language model for few-shot learning}.
\newblock \bibinfo{journal}{\emph{Advances in neural information processing systems}}  \bibinfo{volume}{35} (\bibinfo{year}{2022}), \bibinfo{pages}{23716--23736}.
\newblock


\bibitem[Ataallah et~al\mbox{.}(2024)]%
        {ataallah2024minigpt4}
\bibfield{author}{\bibinfo{person}{Kirolos Ataallah}, \bibinfo{person}{Xiaoqian Shen}, \bibinfo{person}{Eslam Abdelrahman}, \bibinfo{person}{Essam Sleiman}, \bibinfo{person}{Deyao Zhu}, \bibinfo{person}{Jian Ding}, {and} \bibinfo{person}{Mohamed Elhoseiny}.} \bibinfo{year}{2024}\natexlab{}.
\newblock \showarticletitle{Minigpt4-video: Advancing multimodal llms for video understanding with interleaved visual-textual tokens}.
\newblock \bibinfo{journal}{\emph{arXiv preprint arXiv:2404.03413}} (\bibinfo{year}{2024}).
\newblock


\bibitem[Bagad et~al\mbox{.}(2023)]%
        {bagad2023test}
\bibfield{author}{\bibinfo{person}{Piyush Bagad}, \bibinfo{person}{Makarand Tapaswi}, {and} \bibinfo{person}{Cees~GM Snoek}.} \bibinfo{year}{2023}\natexlab{}.
\newblock \showarticletitle{Test of time: Instilling video-language models with a sense of time}. In \bibinfo{booktitle}{\emph{Proceedings of the IEEE/CVF Conference on Computer Vision and Pattern Recognition}}. \bibinfo{pages}{2503--2516}.
\newblock


\bibitem[Bai et~al\mbox{.}(2023)]%
        {bai2023qwen}
\bibfield{author}{\bibinfo{person}{Jinze Bai}, \bibinfo{person}{Shuai Bai}, \bibinfo{person}{Yunfei Chu}, \bibinfo{person}{Zeyu Cui}, \bibinfo{person}{Kai Dang}, \bibinfo{person}{Xiaodong Deng}, \bibinfo{person}{Yang Fan}, \bibinfo{person}{Wenbin Ge}, \bibinfo{person}{Yu Han}, \bibinfo{person}{Fei Huang}, {et~al\mbox{.}}} \bibinfo{year}{2023}\natexlab{}.
\newblock \showarticletitle{Qwen technical report}.
\newblock \bibinfo{journal}{\emph{arXiv preprint arXiv:2309.16609}} (\bibinfo{year}{2023}).
\newblock


\bibitem[Bertasius et~al\mbox{.}(2021)]%
        {bertasius2021space}
\bibfield{author}{\bibinfo{person}{Gedas Bertasius}, \bibinfo{person}{Heng Wang}, {and} \bibinfo{person}{Lorenzo Torresani}.} \bibinfo{year}{2021}\natexlab{}.
\newblock \showarticletitle{Is space-time attention all you need for video understanding?}. In \bibinfo{booktitle}{\emph{ICML}}, Vol.~\bibinfo{volume}{2}. \bibinfo{pages}{4}.
\newblock


\bibitem[Brock et~al\mbox{.}(2021)]%
        {brock2021high}
\bibfield{author}{\bibinfo{person}{Andy Brock}, \bibinfo{person}{Soham De}, \bibinfo{person}{Samuel~L Smith}, {and} \bibinfo{person}{Karen Simonyan}.} \bibinfo{year}{2021}\natexlab{}.
\newblock \showarticletitle{High-performance large-scale image recognition without normalization}. In \bibinfo{booktitle}{\emph{International conference on machine learning}}. PMLR, \bibinfo{pages}{1059--1071}.
\newblock


\bibitem[Bugliarello et~al\mbox{.}(2021)]%
        {bugliarello2021multimodal}
\bibfield{author}{\bibinfo{person}{Emanuele Bugliarello}, \bibinfo{person}{Ryan Cotterell}, \bibinfo{person}{Naoaki Okazaki}, {and} \bibinfo{person}{Desmond Elliott}.} \bibinfo{year}{2021}\natexlab{}.
\newblock \showarticletitle{Multimodal pretraining unmasked: A meta-analysis and a unified framework of vision-and-language BERTs}.
\newblock \bibinfo{journal}{\emph{Transactions of the Association for Computational Linguistics}}  \bibinfo{volume}{9} (\bibinfo{year}{2021}), \bibinfo{pages}{978--994}.
\newblock


\bibitem[Byeon et~al\mbox{.}(2022)]%
        {kakaobrain2022coyo-700m}
\bibfield{author}{\bibinfo{person}{Minwoo Byeon}, \bibinfo{person}{Beomhee Park}, \bibinfo{person}{Haecheon Kim}, \bibinfo{person}{Sungjun Lee}, \bibinfo{person}{Woonhyuk Baek}, {and} \bibinfo{person}{Saehoon Kim}.} \bibinfo{year}{2022}\natexlab{}.
\newblock \bibinfo{title}{COYO-700M: Image-Text Pair Dataset}.
\newblock \bibinfo{howpublished}{\url{https://github.com/kakaobrain/coyo-dataset}}.
\newblock


\bibitem[Caba~Heilbron et~al\mbox{.}(2015)]%
        {caba2015activitynet}
\bibfield{author}{\bibinfo{person}{Fabian Caba~Heilbron}, \bibinfo{person}{Victor Escorcia}, \bibinfo{person}{Bernard Ghanem}, {and} \bibinfo{person}{Juan Carlos~Niebles}.} \bibinfo{year}{2015}\natexlab{}.
\newblock \showarticletitle{Activitynet: A large-scale video benchmark for human activity understanding}. In \bibinfo{booktitle}{\emph{Proceedings of the ieee conference on computer vision and pattern recognition}}. \bibinfo{pages}{961--970}.
\newblock


\bibitem[Cai et~al\mbox{.}(2022)]%
        {cai2022reversible}
\bibfield{author}{\bibinfo{person}{Yuxuan Cai}, \bibinfo{person}{Yizhuang Zhou}, \bibinfo{person}{Qi Han}, \bibinfo{person}{Jianjian Sun}, \bibinfo{person}{Xiangwen Kong}, \bibinfo{person}{Jun Li}, {and} \bibinfo{person}{Xiangyu Zhang}.} \bibinfo{year}{2022}\natexlab{}.
\newblock \showarticletitle{Reversible column networks}.
\newblock \bibinfo{journal}{\emph{arXiv preprint arXiv:2212.11696}} (\bibinfo{year}{2022}).
\newblock


\bibitem[Cai et~al\mbox{.}(2024)]%
        {cai2024internlm2}
\bibfield{author}{\bibinfo{person}{Zheng Cai}, \bibinfo{person}{Maosong Cao}, \bibinfo{person}{Haojiong Chen}, \bibinfo{person}{Kai Chen}, \bibinfo{person}{Keyu Chen}, \bibinfo{person}{Xin Chen}, \bibinfo{person}{Xun Chen}, \bibinfo{person}{Zehui Chen}, \bibinfo{person}{Zhi Chen}, \bibinfo{person}{Pei Chu}, \bibinfo{person}{Xiaoyi Dong}, \bibinfo{person}{Haodong Duan}, \bibinfo{person}{Qi Fan}, \bibinfo{person}{Zhaoye Fei}, \bibinfo{person}{Yang Gao}, \bibinfo{person}{Jiaye Ge}, \bibinfo{person}{Chenya Gu}, \bibinfo{person}{Yuzhe Gu}, \bibinfo{person}{Tao Gui}, \bibinfo{person}{Aijia Guo}, \bibinfo{person}{Qipeng Guo}, \bibinfo{person}{Conghui He}, \bibinfo{person}{Yingfan Hu}, \bibinfo{person}{Ting Huang}, \bibinfo{person}{Tao Jiang}, \bibinfo{person}{Penglong Jiao}, \bibinfo{person}{Zhenjiang Jin}, \bibinfo{person}{Zhikai Lei}, \bibinfo{person}{Jiaxing Li}, \bibinfo{person}{Jingwen Li}, \bibinfo{person}{Linyang Li}, \bibinfo{person}{Shuaibin Li}, \bibinfo{person}{Wei Li}, \bibinfo{person}{Yining Li},
  \bibinfo{person}{Hongwei Liu}, \bibinfo{person}{Jiangning Liu}, \bibinfo{person}{Jiawei Hong}, \bibinfo{person}{Kaiwen Liu}, \bibinfo{person}{Kuikun Liu}, \bibinfo{person}{Xiaoran Liu}, \bibinfo{person}{Chengqi Lv}, \bibinfo{person}{Haijun Lv}, \bibinfo{person}{Kai Lv}, \bibinfo{person}{Li Ma}, \bibinfo{person}{Runyuan Ma}, \bibinfo{person}{Zerun Ma}, \bibinfo{person}{Wenchang Ning}, \bibinfo{person}{Linke Ouyang}, \bibinfo{person}{Jiantao Qiu}, \bibinfo{person}{Yuan Qu}, \bibinfo{person}{Fukai Shang}, \bibinfo{person}{Yunfan Shao}, \bibinfo{person}{Demin Song}, \bibinfo{person}{Zifan Song}, \bibinfo{person}{Zhihao Sui}, \bibinfo{person}{Peng Sun}, \bibinfo{person}{Yu Sun}, \bibinfo{person}{Huanze Tang}, \bibinfo{person}{Bin Wang}, \bibinfo{person}{Guoteng Wang}, \bibinfo{person}{Jiaqi Wang}, \bibinfo{person}{Jiayu Wang}, \bibinfo{person}{Rui Wang}, \bibinfo{person}{Yudong Wang}, \bibinfo{person}{Ziyi Wang}, \bibinfo{person}{Xingjian Wei}, \bibinfo{person}{Qizhen Weng}, \bibinfo{person}{Fan Wu},
  \bibinfo{person}{Yingtong Xiong}, \bibinfo{person}{Chao Xu}, \bibinfo{person}{Ruiliang Xu}, \bibinfo{person}{Hang Yan}, \bibinfo{person}{Yirong Yan}, \bibinfo{person}{Xiaogui Yang}, \bibinfo{person}{Haochen Ye}, \bibinfo{person}{Huaiyuan Ying}, \bibinfo{person}{Jia Yu}, \bibinfo{person}{Jing Yu}, \bibinfo{person}{Yuhang Zang}, \bibinfo{person}{Chuyu Zhang}, \bibinfo{person}{Li Zhang}, \bibinfo{person}{Pan Zhang}, \bibinfo{person}{Peng Zhang}, \bibinfo{person}{Ruijie Zhang}, \bibinfo{person}{Shuo Zhang}, \bibinfo{person}{Songyang Zhang}, \bibinfo{person}{Wenjian Zhang}, \bibinfo{person}{Wenwei Zhang}, \bibinfo{person}{Xingcheng Zhang}, \bibinfo{person}{Xinyue Zhang}, \bibinfo{person}{Hui Zhao}, \bibinfo{person}{Qian Zhao}, \bibinfo{person}{Xiaomeng Zhao}, \bibinfo{person}{Fengzhe Zhou}, \bibinfo{person}{Zaida Zhou}, \bibinfo{person}{Jingming Zhuo}, \bibinfo{person}{Yicheng Zou}, \bibinfo{person}{Xipeng Qiu}, \bibinfo{person}{Yu Qiao}, {and} \bibinfo{person}{Dahua Lin}.} \bibinfo{year}{2024}\natexlab{}.
\newblock \bibinfo{title}{InternLM2 Technical Report}.
\newblock
\newblock
\showeprint[arxiv]{2403.17297}~[cs.CL]


\bibitem[Carreira et~al\mbox{.}(2018)]%
        {carreira2018short}
\bibfield{author}{\bibinfo{person}{Joao Carreira}, \bibinfo{person}{Eric Noland}, \bibinfo{person}{Andras Banki-Horvath}, \bibinfo{person}{Chloe Hillier}, {and} \bibinfo{person}{Andrew Zisserman}.} \bibinfo{year}{2018}\natexlab{}.
\newblock \showarticletitle{A short note about kinetics-600}.
\newblock \bibinfo{journal}{\emph{arXiv preprint arXiv:1808.01340}} (\bibinfo{year}{2018}).
\newblock


\bibitem[Carreira et~al\mbox{.}(2019)]%
        {carreira2019short}
\bibfield{author}{\bibinfo{person}{Joao Carreira}, \bibinfo{person}{Eric Noland}, \bibinfo{person}{Chloe Hillier}, {and} \bibinfo{person}{Andrew Zisserman}.} \bibinfo{year}{2019}\natexlab{}.
\newblock \showarticletitle{A short note on the kinetics-700 human action dataset}.
\newblock \bibinfo{journal}{\emph{arXiv preprint arXiv:1907.06987}} (\bibinfo{year}{2019}).
\newblock


\bibitem[Carreira and Zisserman(2017)]%
        {carreira2017quo}
\bibfield{author}{\bibinfo{person}{Joao Carreira} {and} \bibinfo{person}{Andrew Zisserman}.} \bibinfo{year}{2017}\natexlab{}.
\newblock \showarticletitle{Quo vadis, action recognition? a new model and the kinetics dataset}. In \bibinfo{booktitle}{\emph{proceedings of the IEEE Conference on Computer Vision and Pattern Recognition}}. \bibinfo{pages}{6299--6308}.
\newblock


\bibitem[Changpinyo et~al\mbox{.}(2021)]%
        {changpinyo2021conceptual}
\bibfield{author}{\bibinfo{person}{Soravit Changpinyo}, \bibinfo{person}{Piyush Sharma}, \bibinfo{person}{Nan Ding}, {and} \bibinfo{person}{Radu Soricut}.} \bibinfo{year}{2021}\natexlab{}.
\newblock \showarticletitle{Conceptual 12m: Pushing web-scale image-text pre-training to recognize long-tail visual concepts}. In \bibinfo{booktitle}{\emph{Proceedings of the IEEE/CVF conference on computer vision and pattern recognition}}. \bibinfo{pages}{3558--3568}.
\newblock


\bibitem[Chen et~al\mbox{.}(2023b)]%
        {chen2023videollm}
\bibfield{author}{\bibinfo{person}{Guo Chen}, \bibinfo{person}{Yin-Dong Zheng}, \bibinfo{person}{Jiahao Wang}, \bibinfo{person}{Jilan Xu}, \bibinfo{person}{Yifei Huang}, \bibinfo{person}{Junting Pan}, \bibinfo{person}{Yi Wang}, \bibinfo{person}{Yali Wang}, \bibinfo{person}{Yu Qiao}, \bibinfo{person}{Tong Lu}, {et~al\mbox{.}}} \bibinfo{year}{2023}\natexlab{b}.
\newblock \showarticletitle{Videollm: Modeling video sequence with large language models}.
\newblock \bibinfo{journal}{\emph{arXiv preprint arXiv:2305.13292}} (\bibinfo{year}{2023}).
\newblock


\bibitem[Chen et~al\mbox{.}(2024d)]%
        {chen2024spatial}
\bibfield{author}{\bibinfo{person}{Huilin Chen}, \bibinfo{person}{Lei Wang}, \bibinfo{person}{Yifan Chen}, \bibinfo{person}{Tom Gedeon}, {and} \bibinfo{person}{Piotr Koniusz}.} \bibinfo{year}{2024}\natexlab{d}.
\newblock \showarticletitle{When Spatial meets Temporal in Action Recognition}.
\newblock \bibinfo{journal}{\emph{arXiv preprint arXiv:2411.15284}} (\bibinfo{year}{2024}).
\newblock


\bibitem[Chen et~al\mbox{.}(2024b)]%
        {chen2024videollm}
\bibfield{author}{\bibinfo{person}{Joya Chen}, \bibinfo{person}{Zhaoyang Lv}, \bibinfo{person}{Shiwei Wu}, \bibinfo{person}{Kevin~Qinghong Lin}, \bibinfo{person}{Chenan Song}, \bibinfo{person}{Difei Gao}, \bibinfo{person}{Jia-Wei Liu}, \bibinfo{person}{Ziteng Gao}, \bibinfo{person}{Dongxing Mao}, {and} \bibinfo{person}{Mike~Zheng Shou}.} \bibinfo{year}{2024}\natexlab{b}.
\newblock \showarticletitle{VideoLLM-online: Online Video Large Language Model for Streaming Video}. In \bibinfo{booktitle}{\emph{Proceedings of the IEEE/CVF Conference on Computer Vision and Pattern Recognition}}. \bibinfo{pages}{18407--18418}.
\newblock


\bibitem[Chen et~al\mbox{.}(2021)]%
        {chen2021spatial}
\bibfield{author}{\bibinfo{person}{Jin Chen}, \bibinfo{person}{Xinxiao Wu}, \bibinfo{person}{Yao Hu}, {and} \bibinfo{person}{Jiebo Luo}.} \bibinfo{year}{2021}\natexlab{}.
\newblock \showarticletitle{Spatial-temporal causal inference for partial image-to-video adaptation}. In \bibinfo{booktitle}{\emph{Proceedings of the AAAI Conference on Artificial Intelligence}}, Vol.~\bibinfo{volume}{35}. \bibinfo{pages}{1027--1035}.
\newblock


\bibitem[Chen et~al\mbox{.}(2024f)]%
        {chen2024sharegpt4video}
\bibfield{author}{\bibinfo{person}{Lin Chen}, \bibinfo{person}{Xilin Wei}, \bibinfo{person}{Jinsong Li}, \bibinfo{person}{Xiaoyi Dong}, \bibinfo{person}{Pan Zhang}, \bibinfo{person}{Yuhang Zang}, \bibinfo{person}{Zehui Chen}, \bibinfo{person}{Haodong Duan}, \bibinfo{person}{Bin Lin}, \bibinfo{person}{Zhenyu Tang}, {et~al\mbox{.}}} \bibinfo{year}{2024}\natexlab{f}.
\newblock \showarticletitle{Sharegpt4video: Improving video understanding and generation with better captions}.
\newblock \bibinfo{journal}{\emph{arXiv preprint arXiv:2406.04325}} (\bibinfo{year}{2024}).
\newblock


\bibitem[Chen et~al\mbox{.}(2024a)]%
        {chen2024motionllm}
\bibfield{author}{\bibinfo{person}{Ling-Hao Chen}, \bibinfo{person}{Shunlin Lu}, \bibinfo{person}{Ailing Zeng}, \bibinfo{person}{Hao Zhang}, \bibinfo{person}{Benyou Wang}, \bibinfo{person}{Ruimao Zhang}, {and} \bibinfo{person}{Lei Zhang}.} \bibinfo{year}{2024}\natexlab{a}.
\newblock \showarticletitle{MotionLLM: Understanding Human Behaviors from Human Motions and Videos}.
\newblock \bibinfo{journal}{\emph{arXiv preprint arXiv:2405.20340}} (\bibinfo{year}{2024}).
\newblock


\bibitem[Chen et~al\mbox{.}({[n.\,d.]})]%
        {chen2024motion}
\bibfield{author}{\bibinfo{person}{Qixiang Chen}, \bibinfo{person}{Lei Wang}, \bibinfo{person}{Piotr Koniusz}, {and} \bibinfo{person}{Tom Gedeon}.} \bibinfo{year}{[n.\,d.]}\natexlab{}.
\newblock \showarticletitle{Motion meets attention: Video motion prompts}. In \bibinfo{booktitle}{\emph{The 16th Asian Conference on Machine Learning (Conference Track)}}.
\newblock


\bibitem[Chen et~al\mbox{.}(2023a)]%
        {chen2023vast}
\bibfield{author}{\bibinfo{person}{Sihan Chen}, \bibinfo{person}{Handong Li}, \bibinfo{person}{Qunbo Wang}, \bibinfo{person}{Zijia Zhao}, \bibinfo{person}{Mingzhen Sun}, \bibinfo{person}{Xinxin Zhu}, {and} \bibinfo{person}{Jing Liu}.} \bibinfo{year}{2023}\natexlab{a}.
\newblock \showarticletitle{Vast: A vision-audio-subtitle-text omni-modality foundation model and dataset}.
\newblock \bibinfo{journal}{\emph{Advances in Neural Information Processing Systems}}  \bibinfo{volume}{36} (\bibinfo{year}{2023}), \bibinfo{pages}{72842--72866}.
\newblock


\bibitem[Chen et~al\mbox{.}(2022b)]%
        {chen2022beats}
\bibfield{author}{\bibinfo{person}{Sanyuan Chen}, \bibinfo{person}{Yu Wu}, \bibinfo{person}{Chengyi Wang}, \bibinfo{person}{Shujie Liu}, \bibinfo{person}{Daniel Tompkins}, \bibinfo{person}{Zhuo Chen}, {and} \bibinfo{person}{Furu Wei}.} \bibinfo{year}{2022}\natexlab{b}.
\newblock \showarticletitle{Beats: Audio pre-training with acoustic tokenizers}.
\newblock \bibinfo{journal}{\emph{arXiv preprint arXiv:2212.09058}} (\bibinfo{year}{2022}).
\newblock


\bibitem[Chen et~al\mbox{.}(2024h)]%
        {chen2024sato}
\bibfield{author}{\bibinfo{person}{Wenshuo Chen}, \bibinfo{person}{Hongru Xiao}, \bibinfo{person}{Erhang Zhang}, \bibinfo{person}{Lijie Hu}, \bibinfo{person}{Lei Wang}, \bibinfo{person}{Mengyuan Liu}, {and} \bibinfo{person}{Chen Chen}.} \bibinfo{year}{2024}\natexlab{h}.
\newblock \showarticletitle{SATO: Stable Text-to-Motion Framework}. In \bibinfo{booktitle}{\emph{Proceedings of the 32nd ACM International Conference on Multimedia}}. \bibinfo{pages}{6989--6997}.
\newblock


\bibitem[Chen et~al\mbox{.}(2022a)]%
        {chen2022pali}
\bibfield{author}{\bibinfo{person}{Xi Chen}, \bibinfo{person}{Xiao Wang}, \bibinfo{person}{Soravit Changpinyo}, \bibinfo{person}{AJ Piergiovanni}, \bibinfo{person}{Piotr Padlewski}, \bibinfo{person}{Daniel Salz}, \bibinfo{person}{Sebastian Goodman}, \bibinfo{person}{Adam Grycner}, \bibinfo{person}{Basil Mustafa}, \bibinfo{person}{Lucas Beyer}, {et~al\mbox{.}}} \bibinfo{year}{2022}\natexlab{a}.
\newblock \showarticletitle{Pali: A jointly-scaled multilingual language-image model}.
\newblock \bibinfo{journal}{\emph{arXiv preprint arXiv:2209.06794}} (\bibinfo{year}{2022}).
\newblock


\bibitem[Chen et~al\mbox{.}(2024c)]%
        {chen2024expanding}
\bibfield{author}{\bibinfo{person}{Zhe Chen}, \bibinfo{person}{Weiyun Wang}, \bibinfo{person}{Yue Cao}, \bibinfo{person}{Yangzhou Liu}, \bibinfo{person}{Zhangwei Gao}, \bibinfo{person}{Erfei Cui}, \bibinfo{person}{Jinguo Zhu}, \bibinfo{person}{Shenglong Ye}, \bibinfo{person}{Hao Tian}, \bibinfo{person}{Zhaoyang Liu}, {et~al\mbox{.}}} \bibinfo{year}{2024}\natexlab{c}.
\newblock \showarticletitle{Expanding Performance Boundaries of Open-Source Multimodal Models with Model, Data, and Test-Time Scaling}.
\newblock \bibinfo{journal}{\emph{arXiv preprint arXiv:2412.05271}} (\bibinfo{year}{2024}).
\newblock


\bibitem[Chen et~al\mbox{.}(2024e)]%
        {chen2024far}
\bibfield{author}{\bibinfo{person}{Zhe Chen}, \bibinfo{person}{Weiyun Wang}, \bibinfo{person}{Hao Tian}, \bibinfo{person}{Shenglong Ye}, \bibinfo{person}{Zhangwei Gao}, \bibinfo{person}{Erfei Cui}, \bibinfo{person}{Wenwen Tong}, \bibinfo{person}{Kongzhi Hu}, \bibinfo{person}{Jiapeng Luo}, \bibinfo{person}{Zheng Ma}, {et~al\mbox{.}}} \bibinfo{year}{2024}\natexlab{e}.
\newblock \showarticletitle{How Far Are We to GPT-4V? Closing the Gap to Commercial Multimodal Models with Open-Source Suites}.
\newblock \bibinfo{journal}{\emph{arXiv preprint arXiv:2404.16821}} (\bibinfo{year}{2024}).
\newblock


\bibitem[Chen et~al\mbox{.}(2024g)]%
        {chen2024internvl}
\bibfield{author}{\bibinfo{person}{Zhe Chen}, \bibinfo{person}{Jiannan Wu}, \bibinfo{person}{Wenhai Wang}, \bibinfo{person}{Weijie Su}, \bibinfo{person}{Guo Chen}, \bibinfo{person}{Sen Xing}, \bibinfo{person}{Muyan Zhong}, \bibinfo{person}{Qinglong Zhang}, \bibinfo{person}{Xizhou Zhu}, \bibinfo{person}{Lewei Lu}, {et~al\mbox{.}}} \bibinfo{year}{2024}\natexlab{g}.
\newblock \showarticletitle{Internvl: Scaling up vision foundation models and aligning for generic visual-linguistic tasks}. In \bibinfo{booktitle}{\emph{Proceedings of the IEEE/CVF Conference on Computer Vision and Pattern Recognition}}. \bibinfo{pages}{24185--24198}.
\newblock


\bibitem[Cheng et~al\mbox{.}(2024)]%
        {cheng2024videollama}
\bibfield{author}{\bibinfo{person}{Zesen Cheng}, \bibinfo{person}{Sicong Leng}, \bibinfo{person}{Hang Zhang}, \bibinfo{person}{Yifei Xin}, \bibinfo{person}{Xin Li}, \bibinfo{person}{Guanzheng Chen}, \bibinfo{person}{Yongxin Zhu}, \bibinfo{person}{Wenqi Zhang}, \bibinfo{person}{Ziyang Luo}, \bibinfo{person}{Deli Zhao}, {et~al\mbox{.}}} \bibinfo{year}{2024}\natexlab{}.
\newblock \showarticletitle{VideoLLaMA 2: Advancing Spatial-Temporal Modeling and Audio Understanding in Video-LLMs}.
\newblock \bibinfo{journal}{\emph{arXiv preprint arXiv:2406.07476}} (\bibinfo{year}{2024}).
\newblock


\bibitem[Chiang et~al\mbox{.}(2023)]%
        {chiang2023vicuna}
\bibfield{author}{\bibinfo{person}{Wei-Lin Chiang}, \bibinfo{person}{Zhuohan Li}, \bibinfo{person}{Zi Lin}, \bibinfo{person}{Ying Sheng}, \bibinfo{person}{Zhanghao Wu}, \bibinfo{person}{Hao Zhang}, \bibinfo{person}{Lianmin Zheng}, \bibinfo{person}{Siyuan Zhuang}, \bibinfo{person}{Yonghao Zhuang}, \bibinfo{person}{Joseph~E Gonzalez}, {et~al\mbox{.}}} \bibinfo{year}{2023}\natexlab{}.
\newblock \showarticletitle{Vicuna: An open-source chatbot impressing gpt-4 with 90\%* chatgpt quality}.
\newblock \bibinfo{journal}{\emph{See https://vicuna. lmsys. org (accessed 14 April 2023)}} \bibinfo{volume}{2}, \bibinfo{number}{3} (\bibinfo{year}{2023}), \bibinfo{pages}{6}.
\newblock


\bibitem[Christophe et~al\mbox{.}(2024)]%
        {christophe2024med42}
\bibfield{author}{\bibinfo{person}{Cl{\'e}ment Christophe}, \bibinfo{person}{Praveen~K Kanithi}, \bibinfo{person}{Prateek Munjal}, \bibinfo{person}{Tathagata Raha}, \bibinfo{person}{Nasir Hayat}, \bibinfo{person}{Ronnie Rajan}, \bibinfo{person}{Ahmed Al-Mahrooqi}, \bibinfo{person}{Avani Gupta}, \bibinfo{person}{Muhammad~Umar Salman}, \bibinfo{person}{Gurpreet Gosal}, {et~al\mbox{.}}} \bibinfo{year}{2024}\natexlab{}.
\newblock \showarticletitle{Med42--Evaluating Fine-Tuning Strategies for Medical LLMs: Full-Parameter vs. Parameter-Efficient Approaches}.
\newblock \bibinfo{journal}{\emph{arXiv preprint arXiv:2404.14779}} (\bibinfo{year}{2024}).
\newblock


\bibitem[contributors(2023)]%
        {stablelm2023}
\bibfield{author}{\bibinfo{person}{StableLM contributors}.} \bibinfo{year}{2023}\natexlab{}.
\newblock \bibinfo{title}{StableLM: Stability AI language models}.
\newblock
\newblock
\urldef\tempurl%
\url{https://github.com/stability-AI/stableLM}
\showURL{%
\tempurl}


\bibitem[Damen et~al\mbox{.}(2018)]%
        {Damen2018ScalingEV}
\bibfield{author}{\bibinfo{person}{Dima Damen}, \bibinfo{person}{Hazel Doughty}, \bibinfo{person}{Giovanni~Maria Farinella}, \bibinfo{person}{Sanja Fidler}, \bibinfo{person}{Antonino Furnari}, \bibinfo{person}{Evangelos Kazakos}, \bibinfo{person}{Davide Moltisanti}, \bibinfo{person}{Jonathan Munro}, \bibinfo{person}{Toby Perrett}, \bibinfo{person}{Will Price}, {and} \bibinfo{person}{Michael Wray}.} \bibinfo{year}{2018}\natexlab{}.
\newblock \showarticletitle{Scaling Egocentric Vision: The EPIC-KITCHENS Dataset}.
\newblock \bibinfo{journal}{\emph{ArXiv}}  \bibinfo{volume}{abs/1804.02748} (\bibinfo{year}{2018}).
\newblock
\urldef\tempurl%
\url{https://api.semanticscholar.org/CorpusID:4710439}
\showURL{%
\tempurl}


\bibitem[Damen et~al\mbox{.}(2022)]%
        {damen2022rescaling}
\bibfield{author}{\bibinfo{person}{Dima Damen}, \bibinfo{person}{Hazel Doughty}, \bibinfo{person}{Giovanni~Maria Farinella}, \bibinfo{person}{Antonino Furnari}, \bibinfo{person}{Evangelos Kazakos}, \bibinfo{person}{Jian Ma}, \bibinfo{person}{Davide Moltisanti}, \bibinfo{person}{Jonathan Munro}, \bibinfo{person}{Toby Perrett}, \bibinfo{person}{Will Price}, {et~al\mbox{.}}} \bibinfo{year}{2022}\natexlab{}.
\newblock \showarticletitle{Rescaling egocentric vision: Collection, pipeline and challenges for epic-kitchens-100}.
\newblock \bibinfo{journal}{\emph{International Journal of Computer Vision}} (\bibinfo{year}{2022}), \bibinfo{pages}{1--23}.
\newblock


\bibitem[Das et~al\mbox{.}(2013)]%
        {das2013thousand}
\bibfield{author}{\bibinfo{person}{Pradipto Das}, \bibinfo{person}{Chenliang Xu}, \bibinfo{person}{Richard~F Doell}, {and} \bibinfo{person}{Jason~J Corso}.} \bibinfo{year}{2013}\natexlab{}.
\newblock \showarticletitle{A thousand frames in just a few words: Lingual description of videos through latent topics and sparse object stitching}. In \bibinfo{booktitle}{\emph{Proceedings of the IEEE conference on computer vision and pattern recognition}}. \bibinfo{pages}{2634--2641}.
\newblock


\bibitem[Deng et~al\mbox{.}(2009)]%
        {deng2009imagenet}
\bibfield{author}{\bibinfo{person}{Jia Deng}, \bibinfo{person}{Wei Dong}, \bibinfo{person}{Richard Socher}, \bibinfo{person}{Li-Jia Li}, \bibinfo{person}{Kai Li}, {and} \bibinfo{person}{Li Fei-Fei}.} \bibinfo{year}{2009}\natexlab{}.
\newblock \showarticletitle{Imagenet: A large-scale hierarchical image database}. In \bibinfo{booktitle}{\emph{2009 IEEE conference on computer vision and pattern recognition}}. Ieee, \bibinfo{pages}{248--255}.
\newblock


\bibitem[Devlin et~al\mbox{.}(2019)]%
        {devlin2018bert}
\bibfield{author}{\bibinfo{person}{Jacob Devlin}, \bibinfo{person}{Ming{-}Wei Chang}, \bibinfo{person}{Kenton Lee}, {and} \bibinfo{person}{Kristina Toutanova}.} \bibinfo{year}{2019}\natexlab{}.
\newblock \showarticletitle{{BERT:} Pre-training of Deep Bidirectional Transformers for Language Understanding}. In \bibinfo{booktitle}{\emph{Proceedings of the 2019 Conference of the North American Chapter of the Association for Computational Linguistics: Human Language Technologies, {NAACL-HLT} 2019, Minneapolis, MN, USA, June 2-7, 2019, Volume 1 (Long and Short Papers)}}, \bibfield{editor}{\bibinfo{person}{Jill Burstein}, \bibinfo{person}{Christy Doran}, {and} \bibinfo{person}{Thamar Solorio}} (Eds.). \bibinfo{publisher}{Association for Computational Linguistics}, \bibinfo{pages}{4171--4186}.
\newblock
\urldef\tempurl%
\url{https://doi.org/10.18653/V1/N19-1423}
\showDOI{\tempurl}


\bibitem[Dhingra et~al\mbox{.}(2022)]%
        {10.1162/tacl_a_00459}
\bibfield{author}{\bibinfo{person}{Bhuwan Dhingra}, \bibinfo{person}{Jeremy~R. Cole}, \bibinfo{person}{Julian~Martin Eisenschlos}, \bibinfo{person}{Daniel Gillick}, \bibinfo{person}{Jacob Eisenstein}, {and} \bibinfo{person}{William~W. Cohen}.} \bibinfo{year}{2022}\natexlab{}.
\newblock \showarticletitle{Time-Aware Language Models as Temporal Knowledge Bases}.
\newblock \bibinfo{journal}{\emph{Transactions of the Association for Computational Linguistics}}  \bibinfo{volume}{10} (\bibinfo{date}{03} \bibinfo{year}{2022}), \bibinfo{pages}{257--273}.
\newblock
\showISSN{2307-387X}
\urldef\tempurl%
\url{https://doi.org/10.1162/tacl_a_00459}
\showDOI{\tempurl}
\showeprint{https://direct.mit.edu/tacl/article-pdf/doi/10.1162/tacl\_a\_00459/2004543/tacl\_a\_00459.pdf}


\bibitem[Ding et~al\mbox{.}(2024)]%
        {ding2024lego}
\bibfield{author}{\bibinfo{person}{Dexuan Ding}, \bibinfo{person}{Lei Wang}, \bibinfo{person}{Liyun Zhu}, \bibinfo{person}{Tom Gedeon}, {and} \bibinfo{person}{Piotr Koniusz}.} \bibinfo{year}{2024}\natexlab{}.
\newblock \showarticletitle{Lego: Learnable expansion of graph operators for multi-modal feature fusion}.
\newblock \bibinfo{journal}{\emph{arXiv preprint arXiv:2410.01506}} (\bibinfo{year}{2024}).
\newblock


\bibitem[Dosovitskiy et~al\mbox{.}(2021)]%
        {dosovitskiy2021an}
\bibfield{author}{\bibinfo{person}{Alexey Dosovitskiy}, \bibinfo{person}{Lucas Beyer}, \bibinfo{person}{Alexander Kolesnikov}, \bibinfo{person}{Dirk Weissenborn}, \bibinfo{person}{Xiaohua Zhai}, \bibinfo{person}{Thomas Unterthiner}, \bibinfo{person}{Mostafa Dehghani}, \bibinfo{person}{Matthias Minderer}, \bibinfo{person}{Georg Heigold}, \bibinfo{person}{Sylvain Gelly}, \bibinfo{person}{Jakob Uszkoreit}, {and} \bibinfo{person}{Neil Houlsby}.} \bibinfo{year}{2021}\natexlab{}.
\newblock \showarticletitle{An Image is Worth 16x16 Words: Transformers for Image Recognition at Scale}. In \bibinfo{booktitle}{\emph{International Conference on Learning Representations}}.
\newblock
\urldef\tempurl%
\url{https://openreview.net/forum?id=YicbFdNTTy}
\showURL{%
\tempurl}


\bibitem[Du et~al\mbox{.}(2024)]%
        {du2024uncoveringwhathowcomprehensive}
\bibfield{author}{\bibinfo{person}{Hang Du}, \bibinfo{person}{Sicheng Zhang}, \bibinfo{person}{Binzhu Xie}, \bibinfo{person}{Guoshun Nan}, \bibinfo{person}{Jiayang Zhang}, \bibinfo{person}{Junrui Xu}, \bibinfo{person}{Hangyu Liu}, \bibinfo{person}{Sicong Leng}, \bibinfo{person}{Jiangming Liu}, \bibinfo{person}{Hehe Fan}, \bibinfo{person}{Dajiu Huang}, \bibinfo{person}{Jing Feng}, \bibinfo{person}{Linli Chen}, \bibinfo{person}{Can Zhang}, \bibinfo{person}{Xuhuan Li}, \bibinfo{person}{Hao Zhang}, \bibinfo{person}{Jianhang Chen}, \bibinfo{person}{Qimei Cui}, {and} \bibinfo{person}{Xiaofeng Tao}.} \bibinfo{year}{2024}\natexlab{}.
\newblock \bibinfo{title}{Uncovering What, Why and How: A Comprehensive Benchmark for Causation Understanding of Video Anomaly}.
\newblock
\newblock
\showeprint[arxiv]{2405.00181}~[cs.CV]
\urldef\tempurl%
\url{https://arxiv.org/abs/2405.00181}
\showURL{%
\tempurl}


\bibitem[Fei et~al\mbox{.}(2024)]%
        {fei2024video}
\bibfield{author}{\bibinfo{person}{Jiajun Fei}, \bibinfo{person}{Dian Li}, \bibinfo{person}{Zhidong Deng}, \bibinfo{person}{Zekun Wang}, \bibinfo{person}{Gang Liu}, {and} \bibinfo{person}{Hui Wang}.} \bibinfo{year}{2024}\natexlab{}.
\newblock \showarticletitle{Video-ccam: Enhancing video-language understanding with causal cross-attention masks for short and long videos}.
\newblock \bibinfo{journal}{\emph{arXiv preprint arXiv:2408.14023}} (\bibinfo{year}{2024}).
\newblock


\bibitem[Feichtenhofer et~al\mbox{.}(2019)]%
        {feichtenhofer2019slowfast}
\bibfield{author}{\bibinfo{person}{Christoph Feichtenhofer}, \bibinfo{person}{Haoqi Fan}, \bibinfo{person}{Jitendra Malik}, {and} \bibinfo{person}{Kaiming He}.} \bibinfo{year}{2019}\natexlab{}.
\newblock \showarticletitle{Slowfast networks for video recognition}. In \bibinfo{booktitle}{\emph{Proceedings of the IEEE/CVF international conference on computer vision}}. \bibinfo{pages}{6202--6211}.
\newblock


\bibitem[Fu et~al\mbox{.}(2024a)]%
        {fu2024video}
\bibfield{author}{\bibinfo{person}{Chaoyou Fu}, \bibinfo{person}{Yuhan Dai}, \bibinfo{person}{Yongdong Luo}, \bibinfo{person}{Lei Li}, \bibinfo{person}{Shuhuai Ren}, \bibinfo{person}{Renrui Zhang}, \bibinfo{person}{Zihan Wang}, \bibinfo{person}{Chenyu Zhou}, \bibinfo{person}{Yunhang Shen}, \bibinfo{person}{Mengdan Zhang}, {et~al\mbox{.}}} \bibinfo{year}{2024}\natexlab{a}.
\newblock \showarticletitle{Video-mme: The first-ever comprehensive evaluation benchmark of multi-modal llms in video analysis}.
\newblock \bibinfo{journal}{\emph{arXiv preprint arXiv:2405.21075}} (\bibinfo{year}{2024}).
\newblock


\bibitem[Fu et~al\mbox{.}(2024b)]%
        {fu2024vita}
\bibfield{author}{\bibinfo{person}{Chaoyou Fu}, \bibinfo{person}{Haojia Lin}, \bibinfo{person}{Zuwei Long}, \bibinfo{person}{Yunhang Shen}, \bibinfo{person}{Meng Zhao}, \bibinfo{person}{Yifan Zhang}, \bibinfo{person}{Shaoqi Dong}, \bibinfo{person}{Xiong Wang}, \bibinfo{person}{Di Yin}, \bibinfo{person}{Long Ma}, {et~al\mbox{.}}} \bibinfo{year}{2024}\natexlab{b}.
\newblock \showarticletitle{Vita: Towards open-source interactive omni multimodal llm}.
\newblock \bibinfo{journal}{\emph{arXiv preprint arXiv:2408.05211}} (\bibinfo{year}{2024}).
\newblock


\bibitem[Gao et~al\mbox{.}(2024)]%
        {gao2024mini}
\bibfield{author}{\bibinfo{person}{Zhangwei Gao}, \bibinfo{person}{Zhe Chen}, \bibinfo{person}{Erfei Cui}, \bibinfo{person}{Yiming Ren}, \bibinfo{person}{Weiyun Wang}, \bibinfo{person}{Jinguo Zhu}, \bibinfo{person}{Hao Tian}, \bibinfo{person}{Shenglong Ye}, \bibinfo{person}{Junjun He}, \bibinfo{person}{Xizhou Zhu}, {et~al\mbox{.}}} \bibinfo{year}{2024}\natexlab{}.
\newblock \showarticletitle{Mini-internvl: A flexible-transfer pocket multimodal model with 5\% parameters and 90\% performance}.
\newblock \bibinfo{journal}{\emph{arXiv preprint arXiv:2410.16261}} (\bibinfo{year}{2024}).
\newblock


\bibitem[Gemmeke et~al\mbox{.}(2017)]%
        {gemmeke2017audio}
\bibfield{author}{\bibinfo{person}{Jort~F Gemmeke}, \bibinfo{person}{Daniel~PW Ellis}, \bibinfo{person}{Dylan Freedman}, \bibinfo{person}{Aren Jansen}, \bibinfo{person}{Wade Lawrence}, \bibinfo{person}{R~Channing Moore}, \bibinfo{person}{Manoj Plakal}, {and} \bibinfo{person}{Marvin Ritter}.} \bibinfo{year}{2017}\natexlab{}.
\newblock \showarticletitle{Audio set: An ontology and human-labeled dataset for audio events}. In \bibinfo{booktitle}{\emph{2017 IEEE international conference on acoustics, speech and signal processing (ICASSP)}}. IEEE, \bibinfo{pages}{776--780}.
\newblock


\bibitem[Ghosh et~al\mbox{.}(2024)]%
        {ghosh2024exploring}
\bibfield{author}{\bibinfo{person}{Akash Ghosh}, \bibinfo{person}{Arkadeep Acharya}, \bibinfo{person}{Sriparna Saha}, \bibinfo{person}{Vinija Jain}, {and} \bibinfo{person}{Aman Chadha}.} \bibinfo{year}{2024}\natexlab{}.
\newblock \showarticletitle{Exploring the frontier of vision-language models: A survey of current methodologies and future directions}.
\newblock \bibinfo{journal}{\emph{arXiv preprint arXiv:2404.07214}} (\bibinfo{year}{2024}).
\newblock


\bibitem[Girdhar et~al\mbox{.}(2023a)]%
        {girdhar2023imagebind}
\bibfield{author}{\bibinfo{person}{Rohit Girdhar}, \bibinfo{person}{Alaaeldin El-Nouby}, \bibinfo{person}{Zhuang Liu}, \bibinfo{person}{Mannat Singh}, \bibinfo{person}{Kalyan~Vasudev Alwala}, \bibinfo{person}{Armand Joulin}, {and} \bibinfo{person}{Ishan Misra}.} \bibinfo{year}{2023}\natexlab{a}.
\newblock \showarticletitle{Imagebind: One embedding space to bind them all}. In \bibinfo{booktitle}{\emph{Proceedings of the IEEE/CVF Conference on Computer Vision and Pattern Recognition}}. \bibinfo{pages}{15180--15190}.
\newblock


\bibitem[Girdhar et~al\mbox{.}(2023b)]%
        {girdhar2023omnimae}
\bibfield{author}{\bibinfo{person}{Rohit Girdhar}, \bibinfo{person}{Alaaeldin El-Nouby}, \bibinfo{person}{Mannat Singh}, \bibinfo{person}{Kalyan~Vasudev Alwala}, \bibinfo{person}{Armand Joulin}, {and} \bibinfo{person}{Ishan Misra}.} \bibinfo{year}{2023}\natexlab{b}.
\newblock \showarticletitle{Omnimae: Single model masked pretraining on images and videos}. In \bibinfo{booktitle}{\emph{Proceedings of the IEEE/CVF conference on computer vision and pattern recognition}}. \bibinfo{pages}{10406--10417}.
\newblock


\bibitem[Girdhar and Ramanan(2019)]%
        {girdhar2019cater}
\bibfield{author}{\bibinfo{person}{Rohit Girdhar} {and} \bibinfo{person}{Deva Ramanan}.} \bibinfo{year}{2019}\natexlab{}.
\newblock \showarticletitle{CATER: A diagnostic dataset for Compositional Actions and TEmporal Reasoning}.
\newblock \bibinfo{journal}{\emph{arXiv preprint arXiv:1910.04744}} (\bibinfo{year}{2019}).
\newblock


\bibitem[Girdhar et~al\mbox{.}(2022)]%
        {girdhar2022omnivore}
\bibfield{author}{\bibinfo{person}{Rohit Girdhar}, \bibinfo{person}{Mannat Singh}, \bibinfo{person}{Nikhila Ravi}, \bibinfo{person}{Laurens Van Der~Maaten}, \bibinfo{person}{Armand Joulin}, {and} \bibinfo{person}{Ishan Misra}.} \bibinfo{year}{2022}\natexlab{}.
\newblock \showarticletitle{Omnivore: A single model for many visual modalities}. In \bibinfo{booktitle}{\emph{Proceedings of the IEEE/CVF conference on computer vision and pattern recognition}}. \bibinfo{pages}{16102--16112}.
\newblock


\bibitem[Goyal et~al\mbox{.}(2017)]%
        {goyal2017something}
\bibfield{author}{\bibinfo{person}{Raghav Goyal}, \bibinfo{person}{Samira Ebrahimi~Kahou}, \bibinfo{person}{Vincent Michalski}, \bibinfo{person}{Joanna Materzynska}, \bibinfo{person}{Susanne Westphal}, \bibinfo{person}{Heuna Kim}, \bibinfo{person}{Valentin Haenel}, \bibinfo{person}{Ingo Fruend}, \bibinfo{person}{Peter Yianilos}, \bibinfo{person}{Moritz Mueller-Freitag}, {et~al\mbox{.}}} \bibinfo{year}{2017}\natexlab{}.
\newblock \showarticletitle{The" something something" video database for learning and evaluating visual common sense}. In \bibinfo{booktitle}{\emph{Proceedings of the IEEE international conference on computer vision}}. \bibinfo{pages}{5842--5850}.
\newblock


\bibitem[Grauman et~al\mbox{.}(2022)]%
        {grauman2022ego4d}
\bibfield{author}{\bibinfo{person}{Kristen Grauman}, \bibinfo{person}{Andrew Westbury}, \bibinfo{person}{Eugene Byrne}, \bibinfo{person}{Zachary Chavis}, \bibinfo{person}{Antonino Furnari}, \bibinfo{person}{Rohit Girdhar}, \bibinfo{person}{Jackson Hamburger}, \bibinfo{person}{Hao Jiang}, \bibinfo{person}{Miao Liu}, \bibinfo{person}{Xingyu Liu}, {et~al\mbox{.}}} \bibinfo{year}{2022}\natexlab{}.
\newblock \showarticletitle{Ego4d: Around the world in 3,000 hours of egocentric video}. In \bibinfo{booktitle}{\emph{Proceedings of the IEEE/CVF Conference on Computer Vision and Pattern Recognition}}. \bibinfo{pages}{18995--19012}.
\newblock


\bibitem[Gu et~al\mbox{.}(2018)]%
        {gu2018ava}
\bibfield{author}{\bibinfo{person}{Chunhui Gu}, \bibinfo{person}{Chen Sun}, \bibinfo{person}{David~A Ross}, \bibinfo{person}{Carl Vondrick}, \bibinfo{person}{Caroline Pantofaru}, \bibinfo{person}{Yeqing Li}, \bibinfo{person}{Sudheendra Vijayanarasimhan}, \bibinfo{person}{George Toderici}, \bibinfo{person}{Susanna Ricco}, \bibinfo{person}{Rahul Sukthankar}, {et~al\mbox{.}}} \bibinfo{year}{2018}\natexlab{}.
\newblock \showarticletitle{Ava: A video dataset of spatio-temporally localized atomic visual actions}. In \bibinfo{booktitle}{\emph{Proceedings of the IEEE conference on computer vision and pattern recognition}}. \bibinfo{pages}{6047--6056}.
\newblock


\bibitem[Gu et~al\mbox{.}(2022)]%
        {gu2022wukong}
\bibfield{author}{\bibinfo{person}{Jiaxi Gu}, \bibinfo{person}{Xiaojun Meng}, \bibinfo{person}{Guansong Lu}, \bibinfo{person}{Lu Hou}, \bibinfo{person}{Niu Minzhe}, \bibinfo{person}{Xiaodan Liang}, \bibinfo{person}{Lewei Yao}, \bibinfo{person}{Runhui Huang}, \bibinfo{person}{Wei Zhang}, \bibinfo{person}{Xin Jiang}, {et~al\mbox{.}}} \bibinfo{year}{2022}\natexlab{}.
\newblock \showarticletitle{Wukong: A 100 million large-scale chinese cross-modal pre-training benchmark}.
\newblock \bibinfo{journal}{\emph{Advances in Neural Information Processing Systems}}  \bibinfo{volume}{35} (\bibinfo{year}{2022}), \bibinfo{pages}{26418--26431}.
\newblock


\bibitem[Guo et~al\mbox{.}(2024a)]%
        {guo2024vtg}
\bibfield{author}{\bibinfo{person}{Yongxin Guo}, \bibinfo{person}{Jingyu Liu}, \bibinfo{person}{Mingda Li}, \bibinfo{person}{Xiaoying Tang}, \bibinfo{person}{Xi Chen}, {and} \bibinfo{person}{Bo Zhao}.} \bibinfo{year}{2024}\natexlab{a}.
\newblock \showarticletitle{VTG-LLM: Integrating Timestamp Knowledge into Video LLMs for Enhanced Video Temporal Grounding}.
\newblock \bibinfo{journal}{\emph{arXiv preprint arXiv:2405.13382}} (\bibinfo{year}{2024}).
\newblock


\bibitem[Guo et~al\mbox{.}(2024b)]%
        {guo2024trace}
\bibfield{author}{\bibinfo{person}{Yongxin Guo}, \bibinfo{person}{Jingyu Liu}, \bibinfo{person}{Mingda Li}, \bibinfo{person}{Xiaoying Tang}, \bibinfo{person}{Qingbin Liu}, {and} \bibinfo{person}{Xi Chen}.} \bibinfo{year}{2024}\natexlab{b}.
\newblock \showarticletitle{Trace: Temporal grounding video llm via causal event modeling}.
\newblock \bibinfo{journal}{\emph{arXiv preprint arXiv:2410.05643}} (\bibinfo{year}{2024}).
\newblock


\bibitem[Gurnee and Tegmark(2024)]%
        {gurnee2024language}
\bibfield{author}{\bibinfo{person}{Wes Gurnee} {and} \bibinfo{person}{Max Tegmark}.} \bibinfo{year}{2024}\natexlab{}.
\newblock \showarticletitle{Language Models Represent Space and Time}. In \bibinfo{booktitle}{\emph{The Twelfth International Conference on Learning Representations}}.
\newblock
\urldef\tempurl%
\url{https://openreview.net/forum?id=jE8xbmvFin}
\showURL{%
\tempurl}


\bibitem[Han et~al\mbox{.}(2023)]%
        {han2023autoad}
\bibfield{author}{\bibinfo{person}{Tengda Han}, \bibinfo{person}{Max Bain}, \bibinfo{person}{Arsha Nagrani}, \bibinfo{person}{Gul Varol}, \bibinfo{person}{Weidi Xie}, {and} \bibinfo{person}{Andrew Zisserman}.} \bibinfo{year}{2023}\natexlab{}.
\newblock \showarticletitle{Autoad ii: The sequel-who, when, and what in movie audio description}. In \bibinfo{booktitle}{\emph{Proceedings of the IEEE/CVF International Conference on Computer Vision}}. \bibinfo{pages}{13645--13655}.
\newblock


\bibitem[Han et~al\mbox{.}(2024)]%
        {han2024autoad}
\bibfield{author}{\bibinfo{person}{Tengda Han}, \bibinfo{person}{Max Bain}, \bibinfo{person}{Arsha Nagrani}, \bibinfo{person}{G{\"u}l Varol}, \bibinfo{person}{Weidi Xie}, {and} \bibinfo{person}{Andrew Zisserman}.} \bibinfo{year}{2024}\natexlab{}.
\newblock \showarticletitle{AutoAD III: The Prequel-Back to the Pixels}. In \bibinfo{booktitle}{\emph{Proceedings of the IEEE/CVF Conference on Computer Vision and Pattern Recognition}}. \bibinfo{pages}{18164--18174}.
\newblock


\bibitem[He et~al\mbox{.}(2024)]%
        {he2024ma}
\bibfield{author}{\bibinfo{person}{Bo He}, \bibinfo{person}{Hengduo Li}, \bibinfo{person}{Young~Kyun Jang}, \bibinfo{person}{Menglin Jia}, \bibinfo{person}{Xuefei Cao}, \bibinfo{person}{Ashish Shah}, \bibinfo{person}{Abhinav Shrivastava}, {and} \bibinfo{person}{Ser-Nam Lim}.} \bibinfo{year}{2024}\natexlab{}.
\newblock \showarticletitle{Ma-lmm: Memory-augmented large multimodal model for long-term video understanding}. In \bibinfo{booktitle}{\emph{Proceedings of the IEEE/CVF Conference on Computer Vision and Pattern Recognition}}. \bibinfo{pages}{13504--13514}.
\newblock


\bibitem[He et~al\mbox{.}(2016)]%
        {he2016deep}
\bibfield{author}{\bibinfo{person}{Kaiming He}, \bibinfo{person}{Xiangyu Zhang}, \bibinfo{person}{Shaoqing Ren}, {and} \bibinfo{person}{Jian Sun}.} \bibinfo{year}{2016}\natexlab{}.
\newblock \showarticletitle{Deep residual learning for image recognition}. In \bibinfo{booktitle}{\emph{Proceedings of the IEEE conference on computer vision and pattern recognition}}. \bibinfo{pages}{770--778}.
\newblock


\bibitem[Hendricks et~al\mbox{.}(2017)]%
        {hendricks2017localizingmomentsvideonatural}
\bibfield{author}{\bibinfo{person}{Lisa~Anne Hendricks}, \bibinfo{person}{Oliver Wang}, \bibinfo{person}{Eli Shechtman}, \bibinfo{person}{Josef Sivic}, \bibinfo{person}{Trevor Darrell}, {and} \bibinfo{person}{Bryan Russell}.} \bibinfo{year}{2017}\natexlab{}.
\newblock \bibinfo{title}{Localizing Moments in Video with Natural Language}.
\newblock
\newblock
\showeprint[arxiv]{1708.01641}~[cs.CV]
\urldef\tempurl%
\url{https://arxiv.org/abs/1708.01641}
\showURL{%
\tempurl}


\bibitem[Hoffmann et~al\mbox{.}(2022)]%
        {hoffmann2022training}
\bibfield{author}{\bibinfo{person}{Jordan Hoffmann}, \bibinfo{person}{Sebastian Borgeaud}, \bibinfo{person}{Arthur Mensch}, \bibinfo{person}{Elena Buchatskaya}, \bibinfo{person}{Trevor Cai}, \bibinfo{person}{Eliza Rutherford}, \bibinfo{person}{Diego de~Las Casas}, \bibinfo{person}{Lisa~Anne Hendricks}, \bibinfo{person}{Johannes Welbl}, \bibinfo{person}{Aidan Clark}, {et~al\mbox{.}}} \bibinfo{year}{2022}\natexlab{}.
\newblock \showarticletitle{Training compute-optimal large language models}.
\newblock \bibinfo{journal}{\emph{arXiv preprint arXiv:2203.15556}} (\bibinfo{year}{2022}).
\newblock


\bibitem[Hua et~al\mbox{.}(2024)]%
        {hua2024v2xum}
\bibfield{author}{\bibinfo{person}{Hang Hua}, \bibinfo{person}{Yunlong Tang}, \bibinfo{person}{Chenliang Xu}, {and} \bibinfo{person}{Jiebo Luo}.} \bibinfo{year}{2024}\natexlab{}.
\newblock \showarticletitle{V2xum-llm: Cross-modal video summarization with temporal prompt instruction tuning}.
\newblock \bibinfo{journal}{\emph{arXiv preprint arXiv:2404.12353}} (\bibinfo{year}{2024}).
\newblock


\bibitem[Huang et~al\mbox{.}(2024)]%
        {huang2024vtimellm}
\bibfield{author}{\bibinfo{person}{Bin Huang}, \bibinfo{person}{Xin Wang}, \bibinfo{person}{Hong Chen}, \bibinfo{person}{Zihan Song}, {and} \bibinfo{person}{Wenwu Zhu}.} \bibinfo{year}{2024}\natexlab{}.
\newblock \showarticletitle{Vtimellm: Empower llm to grasp video moments}. In \bibinfo{booktitle}{\emph{Proceedings of the IEEE/CVF Conference on Computer Vision and Pattern Recognition}}. \bibinfo{pages}{14271--14280}.
\newblock


\bibitem[Hummel et~al\mbox{.}(2024)]%
        {hummel2024egocvr}
\bibfield{author}{\bibinfo{person}{Thomas Hummel}, \bibinfo{person}{Shyamgopal Karthik}, \bibinfo{person}{Mariana-Iuliana Georgescu}, {and} \bibinfo{person}{Zeynep Akata}.} \bibinfo{year}{2024}\natexlab{}.
\newblock \showarticletitle{EgoCVR: An Egocentric Benchmark for Fine-Grained Composed Video Retrieval}.
\newblock \bibinfo{journal}{\emph{arXiv preprint arXiv:2407.16658}} (\bibinfo{year}{2024}).
\newblock


\bibitem[Islam et~al\mbox{.}(2024)]%
        {islam2024video}
\bibfield{author}{\bibinfo{person}{Md~Mohaiminul Islam}, \bibinfo{person}{Ngan Ho}, \bibinfo{person}{Xitong Yang}, \bibinfo{person}{Tushar Nagarajan}, \bibinfo{person}{Lorenzo Torresani}, {and} \bibinfo{person}{Gedas Bertasius}.} \bibinfo{year}{2024}\natexlab{}.
\newblock \showarticletitle{Video ReCap: Recursive Captioning of Hour-Long Videos}. In \bibinfo{booktitle}{\emph{Proceedings of the IEEE/CVF Conference on Computer Vision and Pattern Recognition}}. \bibinfo{pages}{18198--18208}.
\newblock


\bibitem[Iyer et~al\mbox{.}(2022)]%
        {iyer2022opt}
\bibfield{author}{\bibinfo{person}{Srinivasan Iyer}, \bibinfo{person}{Xi~Victoria Lin}, \bibinfo{person}{Ramakanth Pasunuru}, \bibinfo{person}{Todor Mihaylov}, \bibinfo{person}{Daniel Simig}, \bibinfo{person}{Ping Yu}, \bibinfo{person}{Kurt Shuster}, \bibinfo{person}{Tianlu Wang}, \bibinfo{person}{Qing Liu}, \bibinfo{person}{Punit~Singh Koura}, {et~al\mbox{.}}} \bibinfo{year}{2022}\natexlab{}.
\newblock \showarticletitle{Opt-iml: Scaling language model instruction meta learning through the lens of generalization}.
\newblock \bibinfo{journal}{\emph{arXiv preprint arXiv:2212.12017}} (\bibinfo{year}{2022}).
\newblock


\bibitem[Jain et~al\mbox{.}(2023)]%
        {jain2023do}
\bibfield{author}{\bibinfo{person}{Raghav Jain}, \bibinfo{person}{Daivik Sojitra}, \bibinfo{person}{Arkadeep Acharya}, \bibinfo{person}{Sriparna Saha}, \bibinfo{person}{Adam Jatowt}, {and} \bibinfo{person}{Sandipan Dandapat}.} \bibinfo{year}{2023}\natexlab{}.
\newblock \showarticletitle{Do Language Models Have a Common Sense regarding Time? Revisiting Temporal Commonsense Reasoning in the Era of Large Language Models}. In \bibinfo{booktitle}{\emph{The 2023 Conference on Empirical Methods in Natural Language Processing}}.
\newblock
\urldef\tempurl%
\url{https://openreview.net/forum?id=akJUrevmwI}
\showURL{%
\tempurl}


\bibitem[Jang et~al\mbox{.}(2019)]%
        {jang2019video}
\bibfield{author}{\bibinfo{person}{Yunseok Jang}, \bibinfo{person}{Yale Song}, \bibinfo{person}{Chris~Dongjoo Kim}, \bibinfo{person}{Youngjae Yu}, \bibinfo{person}{Youngjin Kim}, {and} \bibinfo{person}{Gunhee Kim}.} \bibinfo{year}{2019}\natexlab{}.
\newblock \showarticletitle{Video question answering with spatio-temporal reasoning}.
\newblock \bibinfo{journal}{\emph{International Journal of Computer Vision}}  \bibinfo{volume}{127} (\bibinfo{year}{2019}), \bibinfo{pages}{1385--1412}.
\newblock


\bibitem[Jang et~al\mbox{.}(2017)]%
        {jang2017tgif}
\bibfield{author}{\bibinfo{person}{Yunseok Jang}, \bibinfo{person}{Yale Song}, \bibinfo{person}{Youngjae Yu}, \bibinfo{person}{Youngjin Kim}, {and} \bibinfo{person}{Gunhee Kim}.} \bibinfo{year}{2017}\natexlab{}.
\newblock \showarticletitle{Tgif-qa: Toward spatio-temporal reasoning in visual question answering}. In \bibinfo{booktitle}{\emph{Proceedings of the IEEE conference on computer vision and pattern recognition}}. \bibinfo{pages}{2758--2766}.
\newblock


\bibitem[Jeong(2024)]%
        {jeong2024fine}
\bibfield{author}{\bibinfo{person}{Cheonsu Jeong}.} \bibinfo{year}{2024}\natexlab{}.
\newblock \showarticletitle{Fine-tuning and utilization methods of domain-specific llms}.
\newblock \bibinfo{journal}{\emph{arXiv preprint arXiv:2401.02981}} (\bibinfo{year}{2024}).
\newblock


\bibitem[Jiang et~al\mbox{.}(2023)]%
        {jiang2023mistral}
\bibfield{author}{\bibinfo{person}{Albert~Q Jiang}, \bibinfo{person}{Alexandre Sablayrolles}, \bibinfo{person}{Arthur Mensch}, \bibinfo{person}{Chris Bamford}, \bibinfo{person}{Devendra~Singh Chaplot}, \bibinfo{person}{Diego de~las Casas}, \bibinfo{person}{Florian Bressand}, \bibinfo{person}{Gianna Lengyel}, \bibinfo{person}{Guillaume Lample}, \bibinfo{person}{Lucile Saulnier}, {et~al\mbox{.}}} \bibinfo{year}{2023}\natexlab{}.
\newblock \showarticletitle{Mistral 7B}.
\newblock \bibinfo{journal}{\emph{arXiv preprint arXiv:2310.06825}} (\bibinfo{year}{2023}).
\newblock


\bibitem[Jiang et~al\mbox{.}(2024)]%
        {jiang2024mixtral}
\bibfield{author}{\bibinfo{person}{Albert~Q Jiang}, \bibinfo{person}{Alexandre Sablayrolles}, \bibinfo{person}{Antoine Roux}, \bibinfo{person}{Arthur Mensch}, \bibinfo{person}{Blanche Savary}, \bibinfo{person}{Chris Bamford}, \bibinfo{person}{Devendra~Singh Chaplot}, \bibinfo{person}{Diego de~las Casas}, \bibinfo{person}{Emma~Bou Hanna}, \bibinfo{person}{Florian Bressand}, {et~al\mbox{.}}} \bibinfo{year}{2024}\natexlab{}.
\newblock \showarticletitle{Mixtral of experts}.
\newblock \bibinfo{journal}{\emph{arXiv preprint arXiv:2401.04088}} (\bibinfo{year}{2024}).
\newblock


\bibitem[Kay et~al\mbox{.}(2017)]%
        {kay2017kinetics}
\bibfield{author}{\bibinfo{person}{Will Kay}, \bibinfo{person}{Joao Carreira}, \bibinfo{person}{Karen Simonyan}, \bibinfo{person}{Brian Zhang}, \bibinfo{person}{Chloe Hillier}, \bibinfo{person}{Sudheendra Vijayanarasimhan}, \bibinfo{person}{Fabio Viola}, \bibinfo{person}{Tim Green}, \bibinfo{person}{Trevor Back}, \bibinfo{person}{Paul Natsev}, {et~al\mbox{.}}} \bibinfo{year}{2017}\natexlab{}.
\newblock \showarticletitle{The kinetics human action video dataset}.
\newblock \bibinfo{journal}{\emph{arXiv preprint arXiv:1705.06950}} (\bibinfo{year}{2017}).
\newblock


\bibitem[Kim et~al\mbox{.}(2024)]%
        {kim2024show}
\bibfield{author}{\bibinfo{person}{Byoungjip Kim}, \bibinfo{person}{Dasol Hwang}, \bibinfo{person}{Sungjun Cho}, \bibinfo{person}{Youngsoo Jang}, \bibinfo{person}{Honglak Lee}, {and} \bibinfo{person}{Moontae Lee}.} \bibinfo{year}{2024}\natexlab{}.
\newblock \showarticletitle{Show Think and Tell: Thought-Augmented Fine-Tuning of Large Language Models for Video Captioning}. In \bibinfo{booktitle}{\emph{Proceedings of the IEEE/CVF Conference on Computer Vision and Pattern Recognition}}. \bibinfo{pages}{1808--1817}.
\newblock


\bibitem[Koniusz et~al\mbox{.}(2021)]%
        {koniusz2021tensor}
\bibfield{author}{\bibinfo{person}{Piotr Koniusz}, \bibinfo{person}{Lei Wang}, {and} \bibinfo{person}{Anoop Cherian}.} \bibinfo{year}{2021}\natexlab{}.
\newblock \showarticletitle{Tensor representations for action recognition}.
\newblock \bibinfo{journal}{\emph{IEEE Transactions on Pattern Analysis and Machine Intelligence}} \bibinfo{volume}{44}, \bibinfo{number}{2} (\bibinfo{year}{2021}), \bibinfo{pages}{648--665}.
\newblock


\bibitem[Kordopatis-Zilos et~al\mbox{.}(2019)]%
        {kordopatis2019fivr}
\bibfield{author}{\bibinfo{person}{Giorgos Kordopatis-Zilos}, \bibinfo{person}{Symeon Papadopoulos}, \bibinfo{person}{Ioannis Patras}, {and} \bibinfo{person}{Ioannis Kompatsiaris}.} \bibinfo{year}{2019}\natexlab{}.
\newblock \showarticletitle{FIVR: Fine-grained incident video retrieval}.
\newblock \bibinfo{journal}{\emph{IEEE Transactions on Multimedia}} \bibinfo{volume}{21}, \bibinfo{number}{10} (\bibinfo{year}{2019}), \bibinfo{pages}{2638--2652}.
\newblock


\bibitem[Krishna et~al\mbox{.}(2017)]%
        {krishna2017dense}
\bibfield{author}{\bibinfo{person}{Ranjay Krishna}, \bibinfo{person}{Kenji Hata}, \bibinfo{person}{Frederic Ren}, \bibinfo{person}{Li Fei-Fei}, {and} \bibinfo{person}{Juan Carlos~Niebles}.} \bibinfo{year}{2017}\natexlab{}.
\newblock \showarticletitle{Dense-captioning events in videos}. In \bibinfo{booktitle}{\emph{Proceedings of the IEEE international conference on computer vision}}. \bibinfo{pages}{706--715}.
\newblock


\bibitem[Kuehne et~al\mbox{.}(2011a)]%
        {Kuehne11}
\bibfield{author}{\bibinfo{person}{H. Kuehne}, \bibinfo{person}{H. Jhuang}, \bibinfo{person}{E. Garrote}, \bibinfo{person}{T. Poggio}, {and} \bibinfo{person}{T. Serre}.} \bibinfo{year}{2011}\natexlab{a}.
\newblock \showarticletitle{{HMDB}: a large video database for human motion recognition}. In \bibinfo{booktitle}{\emph{Proceedings of the International Conference on Computer Vision (ICCV)}}.
\newblock


\bibitem[Kuehne et~al\mbox{.}(2011b)]%
        {kuehne2011hmdb}
\bibfield{author}{\bibinfo{person}{Hildegard Kuehne}, \bibinfo{person}{Hueihan Jhuang}, \bibinfo{person}{Est{\'\i}baliz Garrote}, \bibinfo{person}{Tomaso Poggio}, {and} \bibinfo{person}{Thomas Serre}.} \bibinfo{year}{2011}\natexlab{b}.
\newblock \showarticletitle{HMDB: a large video database for human motion recognition}. In \bibinfo{booktitle}{\emph{2011 International conference on computer vision}}. IEEE, \bibinfo{pages}{2556--2563}.
\newblock


\bibitem[Lei et~al\mbox{.}(2018)]%
        {lei2018tvqa}
\bibfield{author}{\bibinfo{person}{Jie Lei}, \bibinfo{person}{Licheng Yu}, \bibinfo{person}{Mohit Bansal}, {and} \bibinfo{person}{Tamara~L Berg}.} \bibinfo{year}{2018}\natexlab{}.
\newblock \showarticletitle{Tvqa: Localized, compositional video question answering}.
\newblock \bibinfo{journal}{\emph{arXiv preprint arXiv:1809.01696}} (\bibinfo{year}{2018}).
\newblock


\bibitem[Lei et~al\mbox{.}(2020)]%
        {lei2020tvr}
\bibfield{author}{\bibinfo{person}{Jie Lei}, \bibinfo{person}{Licheng Yu}, \bibinfo{person}{Tamara~L Berg}, {and} \bibinfo{person}{Mohit Bansal}.} \bibinfo{year}{2020}\natexlab{}.
\newblock \showarticletitle{Tvr: A large-scale dataset for video-subtitle moment retrieval}. In \bibinfo{booktitle}{\emph{Computer Vision--ECCV 2020: 16th European Conference, Glasgow, UK, August 23--28, 2020, Proceedings, Part XXI 16}}. Springer, \bibinfo{pages}{447--463}.
\newblock


\bibitem[Lewis et~al\mbox{.}(2019)]%
        {lewis2020bart}
\bibfield{author}{\bibinfo{person}{Mike Lewis}, \bibinfo{person}{Yinhan Liu}, \bibinfo{person}{Naman Goyal}, \bibinfo{person}{Marjan Ghazvininejad}, \bibinfo{person}{Abdel rahman Mohamed}, \bibinfo{person}{Omer Levy}, \bibinfo{person}{Veselin Stoyanov}, {and} \bibinfo{person}{Luke Zettlemoyer}.} \bibinfo{year}{2019}\natexlab{}.
\newblock \showarticletitle{BART: Denoising Sequence-to-Sequence Pre-training for Natural Language Generation, Translation, and Comprehension}. In \bibinfo{booktitle}{\emph{Annual Meeting of the Association for Computational Linguistics}}.
\newblock
\urldef\tempurl%
\url{https://api.semanticscholar.org/CorpusID:204960716}
\showURL{%
\tempurl}


\bibitem[Li et~al\mbox{.}(2023b)]%
        {li2023blip}
\bibfield{author}{\bibinfo{person}{Junnan Li}, \bibinfo{person}{Dongxu Li}, \bibinfo{person}{Silvio Savarese}, {and} \bibinfo{person}{Steven Hoi}.} \bibinfo{year}{2023}\natexlab{b}.
\newblock \showarticletitle{Blip-2: Bootstrapping language-image pre-training with frozen image encoders and large language models}. In \bibinfo{booktitle}{\emph{International conference on machine learning}}. PMLR, \bibinfo{pages}{19730--19742}.
\newblock


\bibitem[Li et~al\mbox{.}(2023a)]%
        {li2023videochat}
\bibfield{author}{\bibinfo{person}{KunChang Li}, \bibinfo{person}{Yinan He}, \bibinfo{person}{Yi Wang}, \bibinfo{person}{Yizhuo Li}, \bibinfo{person}{Wenhai Wang}, \bibinfo{person}{Ping Luo}, \bibinfo{person}{Yali Wang}, \bibinfo{person}{Limin Wang}, {and} \bibinfo{person}{Yu Qiao}.} \bibinfo{year}{2023}\natexlab{a}.
\newblock \showarticletitle{Videochat: Chat-centric video understanding}.
\newblock \bibinfo{journal}{\emph{arXiv preprint arXiv:2305.06355}} (\bibinfo{year}{2023}).
\newblock


\bibitem[Li et~al\mbox{.}(2024)]%
        {li2024mvbench}
\bibfield{author}{\bibinfo{person}{Kunchang Li}, \bibinfo{person}{Yali Wang}, \bibinfo{person}{Yinan He}, \bibinfo{person}{Yizhuo Li}, \bibinfo{person}{Yi Wang}, \bibinfo{person}{Yi Liu}, \bibinfo{person}{Zun Wang}, \bibinfo{person}{Jilan Xu}, \bibinfo{person}{Guo Chen}, \bibinfo{person}{Ping Luo}, {et~al\mbox{.}}} \bibinfo{year}{2024}\natexlab{}.
\newblock \showarticletitle{Mvbench: A comprehensive multi-modal video understanding benchmark}. In \bibinfo{booktitle}{\emph{Proceedings of the IEEE/CVF Conference on Computer Vision and Pattern Recognition}}. \bibinfo{pages}{22195--22206}.
\newblock


\bibitem[Li et~al\mbox{.}(2020)]%
        {li2020hero}
\bibfield{author}{\bibinfo{person}{Linjie Li}, \bibinfo{person}{Yen-Chun Chen}, \bibinfo{person}{Yu Cheng}, \bibinfo{person}{Zhe Gan}, \bibinfo{person}{Licheng Yu}, {and} \bibinfo{person}{Jingjing Liu}.} \bibinfo{year}{2020}\natexlab{}.
\newblock \showarticletitle{Hero: Hierarchical encoder for video+ language omni-representation pre-training}.
\newblock \bibinfo{journal}{\emph{arXiv preprint arXiv:2005.00200}} (\bibinfo{year}{2020}).
\newblock


\bibitem[Li et~al\mbox{.}(2025)]%
        {li2025llama}
\bibfield{author}{\bibinfo{person}{Yanwei Li}, \bibinfo{person}{Chengyao Wang}, {and} \bibinfo{person}{Jiaya Jia}.} \bibinfo{year}{2025}\natexlab{}.
\newblock \showarticletitle{Llama-vid: An image is worth 2 tokens in large language models}. In \bibinfo{booktitle}{\emph{European Conference on Computer Vision}}. Springer, \bibinfo{pages}{323--340}.
\newblock


\bibitem[Liao et~al\mbox{.}(2024)]%
        {liao2024videoinsta}
\bibfield{author}{\bibinfo{person}{Ruotong Liao}, \bibinfo{person}{Max Erler}, \bibinfo{person}{Huiyu Wang}, \bibinfo{person}{Guangyao Zhai}, \bibinfo{person}{Gengyuan Zhang}, \bibinfo{person}{Yunpu Ma}, {and} \bibinfo{person}{Volker Tresp}.} \bibinfo{year}{2024}\natexlab{}.
\newblock \showarticletitle{VideoINSTA: Zero-shot Long Video Understanding via Informative Spatial-Temporal Reasoning with LLMs}.
\newblock \bibinfo{journal}{\emph{arXiv preprint arXiv:2409.20365}} (\bibinfo{year}{2024}).
\newblock


\bibitem[Lin et~al\mbox{.}(2023)]%
        {lin2023video}
\bibfield{author}{\bibinfo{person}{Bin Lin}, \bibinfo{person}{Yang Ye}, \bibinfo{person}{Bin Zhu}, \bibinfo{person}{Jiaxi Cui}, \bibinfo{person}{Munan Ning}, \bibinfo{person}{Peng Jin}, {and} \bibinfo{person}{Li Yuan}.} \bibinfo{year}{2023}\natexlab{}.
\newblock \showarticletitle{Video-llava: Learning united visual representation by alignment before projection}.
\newblock \bibinfo{journal}{\emph{arXiv preprint arXiv:2311.10122}} (\bibinfo{year}{2023}).
\newblock


\bibitem[Lin et~al\mbox{.}(2024)]%
        {lin2024vila}
\bibfield{author}{\bibinfo{person}{Ji Lin}, \bibinfo{person}{Hongxu Yin}, \bibinfo{person}{Wei Ping}, \bibinfo{person}{Pavlo Molchanov}, \bibinfo{person}{Mohammad Shoeybi}, {and} \bibinfo{person}{Song Han}.} \bibinfo{year}{2024}\natexlab{}.
\newblock \showarticletitle{Vila: On pre-training for visual language models}. In \bibinfo{booktitle}{\emph{Proceedings of the IEEE/CVF Conference on Computer Vision and Pattern Recognition}}. \bibinfo{pages}{26689--26699}.
\newblock


\bibitem[Lin et~al\mbox{.}(2022)]%
        {lin2022causal}
\bibfield{author}{\bibinfo{person}{Xiangru Lin}, \bibinfo{person}{Yuyang Chen}, \bibinfo{person}{Guanbin Li}, {and} \bibinfo{person}{Yizhou Yu}.} \bibinfo{year}{2022}\natexlab{}.
\newblock \showarticletitle{A causal inference look at unsupervised video anomaly detection}. In \bibinfo{booktitle}{\emph{Proceedings of the AAAI Conference on Artificial Intelligence}}, Vol.~\bibinfo{volume}{36}. \bibinfo{pages}{1620--1629}.
\newblock


\bibitem[Liu et~al\mbox{.}(2018)]%
        {8328914}
\bibfield{author}{\bibinfo{person}{Ding Liu}, \bibinfo{person}{Zhaowen Wang}, \bibinfo{person}{Yuchen Fan}, \bibinfo{person}{Xianming Liu}, \bibinfo{person}{Zhangyang Wang}, \bibinfo{person}{Shiyu Chang}, \bibinfo{person}{Xinchao Wang}, {and} \bibinfo{person}{Thomas~S. Huang}.} \bibinfo{year}{2018}\natexlab{}.
\newblock \showarticletitle{Learning Temporal Dynamics for Video Super-Resolution: A Deep Learning Approach}.
\newblock \bibinfo{journal}{\emph{IEEE Transactions on Image Processing}} \bibinfo{volume}{27}, \bibinfo{number}{7} (\bibinfo{year}{2018}), \bibinfo{pages}{3432--3445}.
\newblock
\urldef\tempurl%
\url{https://doi.org/10.1109/TIP.2018.2820807}
\showDOI{\tempurl}


\bibitem[Liu et~al\mbox{.}(2024b)]%
        {liu2024improvedbaselinesvisualinstruction}
\bibfield{author}{\bibinfo{person}{Haotian Liu}, \bibinfo{person}{Chunyuan Li}, \bibinfo{person}{Yuheng Li}, {and} \bibinfo{person}{Yong~Jae Lee}.} \bibinfo{year}{2024}\natexlab{b}.
\newblock \bibinfo{title}{Improved Baselines with Visual Instruction Tuning}.
\newblock
\newblock
\showeprint[arxiv]{2310.03744}~[cs.CV]
\urldef\tempurl%
\url{https://arxiv.org/abs/2310.03744}
\showURL{%
\tempurl}


\bibitem[Liu et~al\mbox{.}(2023)]%
        {liu2023visualinstructiontuning}
\bibfield{author}{\bibinfo{person}{Haotian Liu}, \bibinfo{person}{Chunyuan Li}, \bibinfo{person}{Qingyang Wu}, {and} \bibinfo{person}{Yong~Jae Lee}.} \bibinfo{year}{2023}\natexlab{}.
\newblock \bibinfo{title}{Visual Instruction Tuning}.
\newblock
\newblock
\showeprint[arxiv]{2304.08485}~[cs.CV]
\urldef\tempurl%
\url{https://arxiv.org/abs/2304.08485}
\showURL{%
\tempurl}


\bibitem[Liu et~al\mbox{.}(2024d)]%
        {liu2024kangaroo}
\bibfield{author}{\bibinfo{person}{Jiajun Liu}, \bibinfo{person}{Yibing Wang}, \bibinfo{person}{Hanghang Ma}, \bibinfo{person}{Xiaoping Wu}, \bibinfo{person}{Xiaoqi Ma}, \bibinfo{person}{Xiaoming Wei}, \bibinfo{person}{Jianbin Jiao}, \bibinfo{person}{Enhua Wu}, {and} \bibinfo{person}{Jie Hu}.} \bibinfo{year}{2024}\natexlab{d}.
\newblock \showarticletitle{Kangaroo: A powerful video-language model supporting long-context video input}.
\newblock \bibinfo{journal}{\emph{arXiv preprint arXiv:2408.15542}} (\bibinfo{year}{2024}).
\newblock


\bibitem[Liu et~al\mbox{.}(2025)]%
        {liu2025st}
\bibfield{author}{\bibinfo{person}{Ruyang Liu}, \bibinfo{person}{Chen Li}, \bibinfo{person}{Haoran Tang}, \bibinfo{person}{Yixiao Ge}, \bibinfo{person}{Ying Shan}, {and} \bibinfo{person}{Ge Li}.} \bibinfo{year}{2025}\natexlab{}.
\newblock \showarticletitle{St-llm: Large language models are effective temporal learners}. In \bibinfo{booktitle}{\emph{European Conference on Computer Vision}}. Springer, \bibinfo{pages}{1--18}.
\newblock


\bibitem[Liu et~al\mbox{.}(2024c)]%
        {liu2024tempcompass}
\bibfield{author}{\bibinfo{person}{Yuanxin Liu}, \bibinfo{person}{Shicheng Li}, \bibinfo{person}{Yi Liu}, \bibinfo{person}{Yuxiang Wang}, \bibinfo{person}{Shuhuai Ren}, \bibinfo{person}{Lei Li}, \bibinfo{person}{Sishuo Chen}, \bibinfo{person}{Xu Sun}, {and} \bibinfo{person}{Lu Hou}.} \bibinfo{year}{2024}\natexlab{c}.
\newblock \showarticletitle{Tempcompass: Do video llms really understand videos?}
\newblock \bibinfo{journal}{\emph{arXiv preprint arXiv:2403.00476}} (\bibinfo{year}{2024}).
\newblock


\bibitem[Liu et~al\mbox{.}(2022a)]%
        {liu2022umt}
\bibfield{author}{\bibinfo{person}{Ye Liu}, \bibinfo{person}{Siyuan Li}, \bibinfo{person}{Yang Wu}, \bibinfo{person}{Chang-Wen Chen}, \bibinfo{person}{Ying Shan}, {and} \bibinfo{person}{Xiaohu Qie}.} \bibinfo{year}{2022}\natexlab{a}.
\newblock \showarticletitle{Umt: Unified multi-modal transformers for joint video moment retrieval and highlight detection}. In \bibinfo{booktitle}{\emph{Proceedings of the IEEE/CVF Conference on Computer Vision and Pattern Recognition}}. \bibinfo{pages}{3042--3051}.
\newblock


\bibitem[Liu et~al\mbox{.}(2024a)]%
        {liu2024oryx}
\bibfield{author}{\bibinfo{person}{Zuyan Liu}, \bibinfo{person}{Yuhao Dong}, \bibinfo{person}{Ziwei Liu}, \bibinfo{person}{Winston Hu}, \bibinfo{person}{Jiwen Lu}, {and} \bibinfo{person}{Yongming Rao}.} \bibinfo{year}{2024}\natexlab{a}.
\newblock \showarticletitle{Oryx mllm: On-demand spatial-temporal understanding at arbitrary resolution}.
\newblock \bibinfo{journal}{\emph{arXiv preprint arXiv:2409.12961}} (\bibinfo{year}{2024}).
\newblock


\bibitem[Liu et~al\mbox{.}(2022b)]%
        {liu2022video}
\bibfield{author}{\bibinfo{person}{Ze Liu}, \bibinfo{person}{Jia Ning}, \bibinfo{person}{Yue Cao}, \bibinfo{person}{Yixuan Wei}, \bibinfo{person}{Zheng Zhang}, \bibinfo{person}{Stephen Lin}, {and} \bibinfo{person}{Han Hu}.} \bibinfo{year}{2022}\natexlab{b}.
\newblock \showarticletitle{Video swin transformer}. In \bibinfo{booktitle}{\emph{Proceedings of the IEEE/CVF conference on computer vision and pattern recognition}}. \bibinfo{pages}{3202--3211}.
\newblock


\bibitem[Long et~al\mbox{.}(2024)]%
        {long2024videodrafter}
\bibfield{author}{\bibinfo{person}{Fuchen Long}, \bibinfo{person}{Zhaofan Qiu}, \bibinfo{person}{Ting Yao}, {and} \bibinfo{person}{Tao Mei}.} \bibinfo{year}{2024}\natexlab{}.
\newblock \showarticletitle{Videodrafter: Content-consistent multi-scene video generation with llm}.
\newblock \bibinfo{journal}{\emph{arXiv preprint arXiv:2401.01256}} (\bibinfo{year}{2024}).
\newblock


\bibitem[Lu et~al\mbox{.}(2013)]%
        {6751449}
\bibfield{author}{\bibinfo{person}{Cewu Lu}, \bibinfo{person}{Jianping Shi}, {and} \bibinfo{person}{Jiaya Jia}.} \bibinfo{year}{2013}\natexlab{}.
\newblock \showarticletitle{Abnormal Event Detection at 150 FPS in MATLAB}. In \bibinfo{booktitle}{\emph{2013 IEEE International Conference on Computer Vision}}. \bibinfo{pages}{2720--2727}.
\newblock
\urldef\tempurl%
\url{https://doi.org/10.1109/ICCV.2013.338}
\showDOI{\tempurl}


\bibitem[Luo et~al\mbox{.}(2023)]%
        {luo2023valley}
\bibfield{author}{\bibinfo{person}{Ruipu Luo}, \bibinfo{person}{Ziwang Zhao}, \bibinfo{person}{Min Yang}, \bibinfo{person}{Junwei Dong}, \bibinfo{person}{Da Li}, \bibinfo{person}{Pengcheng Lu}, \bibinfo{person}{Tao Wang}, \bibinfo{person}{Linmei Hu}, \bibinfo{person}{Minghui Qiu}, {and} \bibinfo{person}{Zhongyu Wei}.} \bibinfo{year}{2023}\natexlab{}.
\newblock \showarticletitle{Valley: Video assistant with large language model enhanced ability}.
\newblock \bibinfo{journal}{\emph{arXiv preprint arXiv:2306.07207}} (\bibinfo{year}{2023}).
\newblock


\bibitem[Lyu et~al\mbox{.}(2023)]%
        {lyu2023macaw}
\bibfield{author}{\bibinfo{person}{Chenyang Lyu}, \bibinfo{person}{Minghao Wu}, \bibinfo{person}{Longyue Wang}, \bibinfo{person}{Xinting Huang}, \bibinfo{person}{Bingshuai Liu}, \bibinfo{person}{Zefeng Du}, \bibinfo{person}{Shuming Shi}, {and} \bibinfo{person}{Zhaopeng Tu}.} \bibinfo{year}{2023}\natexlab{}.
\newblock \showarticletitle{Macaw-llm: Multi-modal language modeling with image, audio, video, and text integration}.
\newblock \bibinfo{journal}{\emph{arXiv preprint arXiv:2306.09093}} (\bibinfo{year}{2023}).
\newblock


\bibitem[Maaz et~al\mbox{.}(2024)]%
        {maaz2024videogpt+}
\bibfield{author}{\bibinfo{person}{Muhammad Maaz}, \bibinfo{person}{Hanoona Rasheed}, \bibinfo{person}{Salman Khan}, {and} \bibinfo{person}{Fahad Khan}.} \bibinfo{year}{2024}\natexlab{}.
\newblock \showarticletitle{VideoGPT+: Integrating Image and Video Encoders for Enhanced Video Understanding}.
\newblock \bibinfo{journal}{\emph{arXiv preprint arXiv:2406.09418}} (\bibinfo{year}{2024}).
\newblock


\bibitem[Maaz et~al\mbox{.}(2023)]%
        {maaz2023video}
\bibfield{author}{\bibinfo{person}{Muhammad Maaz}, \bibinfo{person}{Hanoona Rasheed}, \bibinfo{person}{Salman Khan}, {and} \bibinfo{person}{Fahad~Shahbaz Khan}.} \bibinfo{year}{2023}\natexlab{}.
\newblock \showarticletitle{Video-chatgpt: Towards detailed video understanding via large vision and language models}.
\newblock \bibinfo{journal}{\emph{arXiv preprint arXiv:2306.05424}} (\bibinfo{year}{2023}).
\newblock


\bibitem[Mangalam et~al\mbox{.}(2023)]%
        {mangalam2023egoschema}
\bibfield{author}{\bibinfo{person}{Karttikeya Mangalam}, \bibinfo{person}{Raiymbek Akshulakov}, {and} \bibinfo{person}{Jitendra Malik}.} \bibinfo{year}{2023}\natexlab{}.
\newblock \showarticletitle{Egoschema: A diagnostic benchmark for very long-form video language understanding}.
\newblock \bibinfo{journal}{\emph{Advances in Neural Information Processing Systems}}  \bibinfo{volume}{36} (\bibinfo{year}{2023}), \bibinfo{pages}{46212--46244}.
\newblock


\bibitem[Miech et~al\mbox{.}(2019)]%
        {miech2019howto100m}
\bibfield{author}{\bibinfo{person}{Antoine Miech}, \bibinfo{person}{Dimitri Zhukov}, \bibinfo{person}{Jean-Baptiste Alayrac}, \bibinfo{person}{Makarand Tapaswi}, \bibinfo{person}{Ivan Laptev}, {and} \bibinfo{person}{Josef Sivic}.} \bibinfo{year}{2019}\natexlab{}.
\newblock \showarticletitle{Howto100m: Learning a text-video embedding by watching hundred million narrated video clips}. In \bibinfo{booktitle}{\emph{Proceedings of the IEEE/CVF international conference on computer vision}}. \bibinfo{pages}{2630--2640}.
\newblock


\bibitem[Nguyen et~al\mbox{.}(2024)]%
        {nguyen2024video}
\bibfield{author}{\bibinfo{person}{Thong Nguyen}, \bibinfo{person}{Yi Bin}, \bibinfo{person}{Junbin Xiao}, \bibinfo{person}{Leigang Qu}, \bibinfo{person}{Yicong Li}, \bibinfo{person}{Jay~Zhangjie Wu}, \bibinfo{person}{Cong-Duy Nguyen}, \bibinfo{person}{See-Kiong Ng}, {and} \bibinfo{person}{Luu~Anh Tuan}.} \bibinfo{year}{2024}\natexlab{}.
\newblock \showarticletitle{Video-Language Understanding: A Survey from Model Architecture, Model Training, and Data Perspectives}.
\newblock \bibinfo{journal}{\emph{arXiv preprint arXiv:2406.05615}} (\bibinfo{year}{2024}).
\newblock


\bibitem[Nie et~al\mbox{.}(2024)]%
        {nie2024slowfocus}
\bibfield{author}{\bibinfo{person}{Ming Nie}, \bibinfo{person}{Dan Ding}, \bibinfo{person}{Chunwei Wang}, \bibinfo{person}{Yuanfan Guo}, \bibinfo{person}{Jianhua Han}, \bibinfo{person}{Hang Xu}, {and} \bibinfo{person}{Li Zhang}.} \bibinfo{year}{2024}\natexlab{}.
\newblock \showarticletitle{SlowFocus: Enhancing Fine-grained Temporal Understanding in Video {LLM}}. In \bibinfo{booktitle}{\emph{The Thirty-eighth Annual Conference on Neural Information Processing Systems}}.
\newblock
\urldef\tempurl%
\url{https://openreview.net/forum?id=FOkKndty5B}
\showURL{%
\tempurl}


\bibitem[Patrick et~al\mbox{.}(2021)]%
        {patrick2021keeping}
\bibfield{author}{\bibinfo{person}{Mandela Patrick}, \bibinfo{person}{Dylan Campbell}, \bibinfo{person}{Yuki Asano}, \bibinfo{person}{Ishan Misra}, \bibinfo{person}{Florian Metze}, \bibinfo{person}{Christoph Feichtenhofer}, \bibinfo{person}{Andrea Vedaldi}, {and} \bibinfo{person}{Joao~F Henriques}.} \bibinfo{year}{2021}\natexlab{}.
\newblock \showarticletitle{Keeping your eye on the ball: Trajectory attention in video transformers}.
\newblock \bibinfo{journal}{\emph{Advances in neural information processing systems}}  \bibinfo{volume}{34} (\bibinfo{year}{2021}), \bibinfo{pages}{12493--12506}.
\newblock


\bibitem[Qin et~al\mbox{.}(2022)]%
        {qin2022fusing}
\bibfield{author}{\bibinfo{person}{Zhenyue Qin}, \bibinfo{person}{Yang Liu}, \bibinfo{person}{Pan Ji}, \bibinfo{person}{Dongwoo Kim}, \bibinfo{person}{Lei Wang}, \bibinfo{person}{Saeed Anwar}, {and} \bibinfo{person}{Tom Gedeon}.} \bibinfo{year}{2022}\natexlab{}.
\newblock \showarticletitle{Fusing higher-order features in graph neural networks for skeleton-based action recognition}.
\newblock \bibinfo{journal}{\emph{IEEE Transactions on Neural Networks and Learning Systems}} \bibinfo{volume}{35}, \bibinfo{number}{4} (\bibinfo{year}{2022}), \bibinfo{pages}{4783--4797}.
\newblock


\bibitem[Radford et~al\mbox{.}(2021)]%
        {radford2021learning}
\bibfield{author}{\bibinfo{person}{Alec Radford}, \bibinfo{person}{Jong~Wook Kim}, \bibinfo{person}{Chris Hallacy}, \bibinfo{person}{Aditya Ramesh}, \bibinfo{person}{Gabriel Goh}, \bibinfo{person}{Sandhini Agarwal}, \bibinfo{person}{Girish Sastry}, \bibinfo{person}{Amanda Askell}, \bibinfo{person}{Pamela Mishkin}, \bibinfo{person}{Jack Clark}, {et~al\mbox{.}}} \bibinfo{year}{2021}\natexlab{}.
\newblock \showarticletitle{Learning transferable visual models from natural language supervision}. In \bibinfo{booktitle}{\emph{International conference on machine learning}}. PMLR, \bibinfo{pages}{8748--8763}.
\newblock


\bibitem[Radford et~al\mbox{.}(2019)]%
        {radford2019language}
\bibfield{author}{\bibinfo{person}{Alec Radford}, \bibinfo{person}{Jeffrey Wu}, \bibinfo{person}{Rewon Child}, \bibinfo{person}{David Luan}, \bibinfo{person}{Dario Amodei}, \bibinfo{person}{Ilya Sutskever}, {et~al\mbox{.}}} \bibinfo{year}{2019}\natexlab{}.
\newblock \showarticletitle{Language models are unsupervised multitask learners}.
\newblock \bibinfo{journal}{\emph{OpenAI blog}} \bibinfo{volume}{1}, \bibinfo{number}{8} (\bibinfo{year}{2019}), \bibinfo{pages}{9}.
\newblock


\bibitem[Raffel et~al\mbox{.}(2020)]%
        {raffel2020exploring}
\bibfield{author}{\bibinfo{person}{Colin Raffel}, \bibinfo{person}{Noam Shazeer}, \bibinfo{person}{Adam Roberts}, \bibinfo{person}{Katherine Lee}, \bibinfo{person}{Sharan Narang}, \bibinfo{person}{Michael Matena}, \bibinfo{person}{Yanqi Zhou}, \bibinfo{person}{Wei Li}, {and} \bibinfo{person}{Peter~J Liu}.} \bibinfo{year}{2020}\natexlab{}.
\newblock \showarticletitle{Exploring the limits of transfer learning with a unified text-to-text transformer}.
\newblock \bibinfo{journal}{\emph{Journal of machine learning research}} \bibinfo{volume}{21}, \bibinfo{number}{140} (\bibinfo{year}{2020}), \bibinfo{pages}{1--67}.
\newblock


\bibitem[Raj et~al\mbox{.}(2024)]%
        {raj2024tracknetv4}
\bibfield{author}{\bibinfo{person}{Arjun Raj}, \bibinfo{person}{Lei Wang}, {and} \bibinfo{person}{Tom Gedeon}.} \bibinfo{year}{2024}\natexlab{}.
\newblock \showarticletitle{TrackNetV4: Enhancing Fast Sports Object Tracking with Motion Attention Maps}.
\newblock \bibinfo{journal}{\emph{arXiv preprint arXiv:2409.14543}} (\bibinfo{year}{2024}).
\newblock


\bibitem[Ramachandra and Jones(2020)]%
        {ramachandra2020streetscenenewdataset}
\bibfield{author}{\bibinfo{person}{Bharathkumar Ramachandra} {and} \bibinfo{person}{Michael Jones}.} \bibinfo{year}{2020}\natexlab{}.
\newblock \bibinfo{title}{Street Scene: A new dataset and evaluation protocol for video anomaly detection}.
\newblock
\newblock
\showeprint[arxiv]{1902.05872}~[cs.CV]
\urldef\tempurl%
\url{https://arxiv.org/abs/1902.05872}
\showURL{%
\tempurl}


\bibitem[Ridnik et~al\mbox{.}(2021)]%
        {ridnik2021imagenet21k}
\bibfield{author}{\bibinfo{person}{Tal Ridnik}, \bibinfo{person}{Emanuel Ben-Baruch}, \bibinfo{person}{Asaf Noy}, {and} \bibinfo{person}{Lihi Zelnik-Manor}.} \bibinfo{year}{2021}\natexlab{}.
\newblock \bibinfo{title}{ImageNet-21K Pretraining for the Masses}.
\newblock
\newblock
\showeprint[arxiv]{2104.10972}~[cs.CV]


\bibitem[Rohrbach et~al\mbox{.}(2017)]%
        {rohrbach2017movie}
\bibfield{author}{\bibinfo{person}{Anna Rohrbach}, \bibinfo{person}{Atousa Torabi}, \bibinfo{person}{Marcus Rohrbach}, \bibinfo{person}{Niket Tandon}, \bibinfo{person}{Christopher Pal}, \bibinfo{person}{Hugo Larochelle}, \bibinfo{person}{Aaron Courville}, {and} \bibinfo{person}{Bernt Schiele}.} \bibinfo{year}{2017}\natexlab{}.
\newblock \showarticletitle{Movie description}.
\newblock \bibinfo{journal}{\emph{International Journal of Computer Vision}}  \bibinfo{volume}{123} (\bibinfo{year}{2017}), \bibinfo{pages}{94--120}.
\newblock


\bibitem[Sadhu et~al\mbox{.}(2021)]%
        {sadhu2021visual}
\bibfield{author}{\bibinfo{person}{Arka Sadhu}, \bibinfo{person}{Tanmay Gupta}, \bibinfo{person}{Mark Yatskar}, \bibinfo{person}{Ram Nevatia}, {and} \bibinfo{person}{Aniruddha Kembhavi}.} \bibinfo{year}{2021}\natexlab{}.
\newblock \showarticletitle{Visual semantic role labeling for video understanding}. In \bibinfo{booktitle}{\emph{Proceedings of the IEEE/CVF Conference on Computer Vision and Pattern Recognition}}. \bibinfo{pages}{5589--5600}.
\newblock


\bibitem[Schuhmann et~al\mbox{.}(2022)]%
        {schuhmann2022laion}
\bibfield{author}{\bibinfo{person}{Christoph Schuhmann}, \bibinfo{person}{Romain Beaumont}, \bibinfo{person}{Richard Vencu}, \bibinfo{person}{Cade Gordon}, \bibinfo{person}{Ross Wightman}, \bibinfo{person}{Mehdi Cherti}, \bibinfo{person}{Theo Coombes}, \bibinfo{person}{Aarush Katta}, \bibinfo{person}{Clayton Mullis}, \bibinfo{person}{Mitchell Wortsman}, {et~al\mbox{.}}} \bibinfo{year}{2022}\natexlab{}.
\newblock \showarticletitle{Laion-5b: An open large-scale dataset for training next generation image-text models}.
\newblock \bibinfo{journal}{\emph{Advances in Neural Information Processing Systems}}  \bibinfo{volume}{35} (\bibinfo{year}{2022}), \bibinfo{pages}{25278--25294}.
\newblock


\bibitem[Sharma et~al\mbox{.}(2018)]%
        {sharma2018conceptual}
\bibfield{author}{\bibinfo{person}{Piyush Sharma}, \bibinfo{person}{Nan Ding}, \bibinfo{person}{Sebastian Goodman}, {and} \bibinfo{person}{Radu Soricut}.} \bibinfo{year}{2018}\natexlab{}.
\newblock \showarticletitle{Conceptual captions: A cleaned, hypernymed, image alt-text dataset for automatic image captioning}. In \bibinfo{booktitle}{\emph{Proceedings of the 56th Annual Meeting of the Association for Computational Linguistics (Volume 1: Long Papers)}}. \bibinfo{pages}{2556--2565}.
\newblock


\bibitem[Sherstinsky(2020)]%
        {sherstinsky2020fundamentals}
\bibfield{author}{\bibinfo{person}{Alex Sherstinsky}.} \bibinfo{year}{2020}\natexlab{}.
\newblock \showarticletitle{Fundamentals of recurrent neural network (RNN) and long short-term memory (LSTM) network}.
\newblock \bibinfo{journal}{\emph{Physica D: Nonlinear Phenomena}}  \bibinfo{volume}{404} (\bibinfo{year}{2020}), \bibinfo{pages}{132306}.
\newblock


\bibitem[Shi et~al\mbox{.}(2024)]%
        {shi2024eagle}
\bibfield{author}{\bibinfo{person}{Min Shi}, \bibinfo{person}{Fuxiao Liu}, \bibinfo{person}{Shihao Wang}, \bibinfo{person}{Shijia Liao}, \bibinfo{person}{Subhashree Radhakrishnan}, \bibinfo{person}{De-An Huang}, \bibinfo{person}{Hongxu Yin}, \bibinfo{person}{Karan Sapra}, \bibinfo{person}{Yaser Yacoob}, \bibinfo{person}{Humphrey Shi}, {et~al\mbox{.}}} \bibinfo{year}{2024}\natexlab{}.
\newblock \showarticletitle{Eagle: Exploring the design space for multimodal llms with mixture of encoders}.
\newblock \bibinfo{journal}{\emph{arXiv preprint arXiv:2408.15998}} (\bibinfo{year}{2024}).
\newblock


\bibitem[Shu et~al\mbox{.}(2023)]%
        {shu2023audio}
\bibfield{author}{\bibinfo{person}{Fangxun Shu}, \bibinfo{person}{Lei Zhang}, \bibinfo{person}{Hao Jiang}, {and} \bibinfo{person}{Cihang Xie}.} \bibinfo{year}{2023}\natexlab{}.
\newblock \showarticletitle{Audio-visual llm for video understanding}.
\newblock \bibinfo{journal}{\emph{arXiv preprint arXiv:2312.06720}} (\bibinfo{year}{2023}).
\newblock


\bibitem[Shu et~al\mbox{.}(2024)]%
        {shu2024video}
\bibfield{author}{\bibinfo{person}{Yan Shu}, \bibinfo{person}{Peitian Zhang}, \bibinfo{person}{Zheng Liu}, \bibinfo{person}{Minghao Qin}, \bibinfo{person}{Junjie Zhou}, \bibinfo{person}{Tiejun Huang}, {and} \bibinfo{person}{Bo Zhao}.} \bibinfo{year}{2024}\natexlab{}.
\newblock \showarticletitle{Video-xl: Extra-long vision language model for hour-scale video understanding}.
\newblock \bibinfo{journal}{\emph{arXiv preprint arXiv:2409.14485}} (\bibinfo{year}{2024}).
\newblock


\bibitem[Sigurdsson et~al\mbox{.}(2016)]%
        {sigurdsson2016hollywood}
\bibfield{author}{\bibinfo{person}{Gunnar~A Sigurdsson}, \bibinfo{person}{G{\"u}l Varol}, \bibinfo{person}{Xiaolong Wang}, \bibinfo{person}{Ali Farhadi}, \bibinfo{person}{Ivan Laptev}, {and} \bibinfo{person}{Abhinav Gupta}.} \bibinfo{year}{2016}\natexlab{}.
\newblock \showarticletitle{Hollywood in homes: Crowdsourcing data collection for activity understanding}. In \bibinfo{booktitle}{\emph{Computer Vision--ECCV 2016: 14th European Conference, Amsterdam, The Netherlands, October 11--14, 2016, Proceedings, Part I 14}}. Springer, \bibinfo{pages}{510--526}.
\newblock


\bibitem[Soomro(2012)]%
        {soomro2012ucf101}
\bibfield{author}{\bibinfo{person}{K Soomro}.} \bibinfo{year}{2012}\natexlab{}.
\newblock \showarticletitle{UCF101: A dataset of 101 human actions classes from videos in the wild}.
\newblock \bibinfo{journal}{\emph{arXiv preprint arXiv:1212.0402}} (\bibinfo{year}{2012}).
\newblock


\bibitem[Srinivasan et~al\mbox{.}(2021)]%
        {10.1145/3404835.3463257}
\bibfield{author}{\bibinfo{person}{Krishna Srinivasan}, \bibinfo{person}{Karthik Raman}, \bibinfo{person}{Jiecao Chen}, \bibinfo{person}{Michael Bendersky}, {and} \bibinfo{person}{Marc Najork}.} \bibinfo{year}{2021}\natexlab{}.
\newblock \showarticletitle{WIT: Wikipedia-Based Image Text Dataset for Multimodal Multilingual Machine Learning}. In \bibinfo{booktitle}{\emph{Proceedings of the 44th International ACM SIGIR Conference on Research and Development in Information Retrieval}} (Virtual Event, Canada) \emph{(\bibinfo{series}{SIGIR '21})}. \bibinfo{publisher}{Association for Computing Machinery}, \bibinfo{address}{New York, NY, USA}, \bibinfo{pages}{2443–2449}.
\newblock
\showISBNx{9781450380379}
\urldef\tempurl%
\url{https://doi.org/10.1145/3404835.3463257}
\showDOI{\tempurl}


\bibitem[Sun et~al\mbox{.}(2019)]%
        {sun2019videobert}
\bibfield{author}{\bibinfo{person}{Chen Sun}, \bibinfo{person}{Austin Myers}, \bibinfo{person}{Carl Vondrick}, \bibinfo{person}{Kevin Murphy}, {and} \bibinfo{person}{Cordelia Schmid}.} \bibinfo{year}{2019}\natexlab{}.
\newblock \showarticletitle{Videobert: A joint model for video and language representation learning}. In \bibinfo{booktitle}{\emph{Proceedings of the IEEE/CVF international conference on computer vision}}. \bibinfo{pages}{7464--7473}.
\newblock


\bibitem[Sun et~al\mbox{.}(2023)]%
        {sun2023eva}
\bibfield{author}{\bibinfo{person}{Quan Sun}, \bibinfo{person}{Yuxin Fang}, \bibinfo{person}{Ledell Wu}, \bibinfo{person}{Xinlong Wang}, {and} \bibinfo{person}{Yue Cao}.} \bibinfo{year}{2023}\natexlab{}.
\newblock \showarticletitle{Eva-clip: Improved training techniques for clip at scale}.
\newblock \bibinfo{journal}{\emph{arXiv preprint arXiv:2303.15389}} (\bibinfo{year}{2023}).
\newblock


\bibitem[Tan et~al\mbox{.}(2024)]%
        {tan2024are}
\bibfield{author}{\bibinfo{person}{Mingtian Tan}, \bibinfo{person}{Mike~A Merrill}, \bibinfo{person}{Vinayak Gupta}, \bibinfo{person}{Tim Althoff}, {and} \bibinfo{person}{Thomas Hartvigsen}.} \bibinfo{year}{2024}\natexlab{}.
\newblock \showarticletitle{Are Language Models Actually Useful for Time Series Forecasting?}. In \bibinfo{booktitle}{\emph{The Thirty-eighth Annual Conference on Neural Information Processing Systems}}.
\newblock
\urldef\tempurl%
\url{https://openreview.net/forum?id=DV15UbHCY1}
\showURL{%
\tempurl}


\bibitem[Tang et~al\mbox{.}(2023)]%
        {tang2023video}
\bibfield{author}{\bibinfo{person}{Yunlong Tang}, \bibinfo{person}{Jing Bi}, \bibinfo{person}{Siting Xu}, \bibinfo{person}{Luchuan Song}, \bibinfo{person}{Susan Liang}, \bibinfo{person}{Teng Wang}, \bibinfo{person}{Daoan Zhang}, \bibinfo{person}{Jie An}, \bibinfo{person}{Jingyang Lin}, \bibinfo{person}{Rongyi Zhu}, {et~al\mbox{.}}} \bibinfo{year}{2023}\natexlab{}.
\newblock \showarticletitle{Video understanding with large language models: A survey}.
\newblock \bibinfo{journal}{\emph{arXiv preprint arXiv:2312.17432}} (\bibinfo{year}{2023}).
\newblock


\bibitem[Tang et~al\mbox{.}(2019)]%
        {tang2019coin}
\bibfield{author}{\bibinfo{person}{Yansong Tang}, \bibinfo{person}{Dajun Ding}, \bibinfo{person}{Yongming Rao}, \bibinfo{person}{Yu Zheng}, \bibinfo{person}{Danyang Zhang}, \bibinfo{person}{Lili Zhao}, \bibinfo{person}{Jiwen Lu}, {and} \bibinfo{person}{Jie Zhou}.} \bibinfo{year}{2019}\natexlab{}.
\newblock \showarticletitle{Coin: A large-scale dataset for comprehensive instructional video analysis}. In \bibinfo{booktitle}{\emph{Proceedings of the IEEE/CVF Conference on Computer Vision and Pattern Recognition}}. \bibinfo{pages}{1207--1216}.
\newblock


\bibitem[Tapaswi et~al\mbox{.}(2016)]%
        {tapaswi2016movieqa}
\bibfield{author}{\bibinfo{person}{Makarand Tapaswi}, \bibinfo{person}{Yukun Zhu}, \bibinfo{person}{Rainer Stiefelhagen}, \bibinfo{person}{Antonio Torralba}, \bibinfo{person}{Raquel Urtasun}, {and} \bibinfo{person}{Sanja Fidler}.} \bibinfo{year}{2016}\natexlab{}.
\newblock \showarticletitle{Movieqa: Understanding stories in movies through question-answering}. In \bibinfo{booktitle}{\emph{Proceedings of the IEEE conference on computer vision and pattern recognition}}. \bibinfo{pages}{4631--4640}.
\newblock


\bibitem[Together.xyz(2023)]%
        {together2023redpajama}
\bibfield{author}{\bibinfo{person}{Together.xyz}.} \bibinfo{year}{2023}\natexlab{}.
\newblock \bibinfo{title}{Releasing 3b and 7b redpajama incite family of models including base, instruction-tuned and chat models}.
\newblock
\newblock
\urldef\tempurl%
\url{https://www.together.xyz/blog/redpajama-models-v1}
\showURL{%
\tempurl}


\bibitem[Touvron et~al\mbox{.}(2023)]%
        {touvron2023llama}
\bibfield{author}{\bibinfo{person}{Hugo Touvron}, \bibinfo{person}{Thibaut Lavril}, \bibinfo{person}{Gautier Izacard}, \bibinfo{person}{Xavier Martinet}, \bibinfo{person}{Marie-Anne Lachaux}, \bibinfo{person}{Timoth{\'e}e Lacroix}, \bibinfo{person}{Baptiste Rozi{\`e}re}, \bibinfo{person}{Naman Goyal}, \bibinfo{person}{Eric Hambro}, \bibinfo{person}{Faisal Azhar}, {et~al\mbox{.}}} \bibinfo{year}{2023}\natexlab{}.
\newblock \showarticletitle{Llama: Open and efficient foundation language models}.
\newblock \bibinfo{journal}{\emph{arXiv preprint arXiv:2302.13971}} (\bibinfo{year}{2023}).
\newblock


\bibitem[Tran et~al\mbox{.}(2015)]%
        {tran2015learning}
\bibfield{author}{\bibinfo{person}{Du Tran}, \bibinfo{person}{Lubomir Bourdev}, \bibinfo{person}{Rob Fergus}, \bibinfo{person}{Lorenzo Torresani}, {and} \bibinfo{person}{Manohar Paluri}.} \bibinfo{year}{2015}\natexlab{}.
\newblock \showarticletitle{Learning spatiotemporal features with 3d convolutional networks}. In \bibinfo{booktitle}{\emph{Proceedings of the IEEE international conference on computer vision}}. \bibinfo{pages}{4489--4497}.
\newblock


\bibitem[Vaswani et~al\mbox{.}(2017)]%
        {NIPS2017_3f5ee243}
\bibfield{author}{\bibinfo{person}{Ashish Vaswani}, \bibinfo{person}{Noam Shazeer}, \bibinfo{person}{Niki Parmar}, \bibinfo{person}{Jakob Uszkoreit}, \bibinfo{person}{Llion Jones}, \bibinfo{person}{Aidan~N Gomez}, \bibinfo{person}{\L~ukasz Kaiser}, {and} \bibinfo{person}{Illia Polosukhin}.} \bibinfo{year}{2017}\natexlab{}.
\newblock \showarticletitle{Attention is All you Need}. In \bibinfo{booktitle}{\emph{Advances in Neural Information Processing Systems}}, \bibfield{editor}{\bibinfo{person}{I.~Guyon}, \bibinfo{person}{U.~Von Luxburg}, \bibinfo{person}{S.~Bengio}, \bibinfo{person}{H.~Wallach}, \bibinfo{person}{R.~Fergus}, \bibinfo{person}{S.~Vishwanathan}, {and} \bibinfo{person}{R.~Garnett}} (Eds.), Vol.~\bibinfo{volume}{30}. \bibinfo{publisher}{Curran Associates, Inc.}
\newblock
\urldef\tempurl%
\url{https://proceedings.neurips.cc/paper/2017/file/3f5ee243547dee91fbd053c1c4a845aa-Paper.pdf}
\showURL{%
\tempurl}


\bibitem[Vondrick et~al\mbox{.}(2016)]%
        {vondrick2016generating}
\bibfield{author}{\bibinfo{person}{Carl Vondrick}, \bibinfo{person}{Hamed Pirsiavash}, {and} \bibinfo{person}{Antonio Torralba}.} \bibinfo{year}{2016}\natexlab{}.
\newblock \showarticletitle{Generating videos with scene dynamics}.
\newblock \bibinfo{journal}{\emph{Advances in neural information processing systems}}  \bibinfo{volume}{29} (\bibinfo{year}{2016}).
\newblock


\bibitem[Wang et~al\mbox{.}(2024i)]%
        {wang2024cosmo}
\bibfield{author}{\bibinfo{person}{Alex~Jinpeng Wang}, \bibinfo{person}{Linjie Li}, \bibinfo{person}{Kevin~Qinghong Lin}, \bibinfo{person}{Jianfeng Wang}, \bibinfo{person}{Kevin Lin}, \bibinfo{person}{Zhengyuan Yang}, \bibinfo{person}{Lijuan Wang}, {and} \bibinfo{person}{Mike~Zheng Shou}.} \bibinfo{year}{2024}\natexlab{i}.
\newblock \showarticletitle{COSMO: COntrastive Streamlined MultimOdal Model with Interleaved Pre-Training}.
\newblock \bibinfo{journal}{\emph{arXiv preprint arXiv:2401.00849}} (\bibinfo{year}{2024}).
\newblock


\bibitem[Wang et~al\mbox{.}(2023a)]%
        {wang2023chatvideo}
\bibfield{author}{\bibinfo{person}{Junke Wang}, \bibinfo{person}{Dongdong Chen}, \bibinfo{person}{Chong Luo}, \bibinfo{person}{Xiyang Dai}, \bibinfo{person}{Lu Yuan}, \bibinfo{person}{Zuxuan Wu}, {and} \bibinfo{person}{Yu-Gang Jiang}.} \bibinfo{year}{2023}\natexlab{a}.
\newblock \showarticletitle{Chatvideo: A tracklet-centric multimodal and versatile video understanding system}.
\newblock \bibinfo{journal}{\emph{arXiv preprint arXiv:2304.14407}} (\bibinfo{year}{2023}).
\newblock


\bibitem[Wang et~al\mbox{.}(2024c)]%
        {wang2024omnivid}
\bibfield{author}{\bibinfo{person}{Junke Wang}, \bibinfo{person}{Dongdong Chen}, \bibinfo{person}{Chong Luo}, \bibinfo{person}{Bo He}, \bibinfo{person}{Lu Yuan}, \bibinfo{person}{Zuxuan Wu}, {and} \bibinfo{person}{Yu-Gang Jiang}.} \bibinfo{year}{2024}\natexlab{c}.
\newblock \showarticletitle{Omnivid: A generative framework for universal video understanding}. In \bibinfo{booktitle}{\emph{Proceedings of the IEEE/CVF Conference on Computer Vision and Pattern Recognition}}. \bibinfo{pages}{18209--18220}.
\newblock


\bibitem[Wang et~al\mbox{.}(2022a)]%
        {wang2022omnivl}
\bibfield{author}{\bibinfo{person}{Junke Wang}, \bibinfo{person}{Dongdong Chen}, \bibinfo{person}{Zuxuan Wu}, \bibinfo{person}{Chong Luo}, \bibinfo{person}{Luowei Zhou}, \bibinfo{person}{Yucheng Zhao}, \bibinfo{person}{Yujia Xie}, \bibinfo{person}{Ce Liu}, \bibinfo{person}{Yu-Gang Jiang}, {and} \bibinfo{person}{Lu Yuan}.} \bibinfo{year}{2022}\natexlab{a}.
\newblock \showarticletitle{Omnivl: One foundation model for image-language and video-language tasks}.
\newblock \bibinfo{journal}{\emph{Advances in neural information processing systems}}  \bibinfo{volume}{35} (\bibinfo{year}{2022}), \bibinfo{pages}{5696--5710}.
\newblock


\bibitem[Wang et~al\mbox{.}(2024f)]%
        {wang2024towards}
\bibfield{author}{\bibinfo{person}{Jiexin Wang}, \bibinfo{person}{Adam Jatowt}, {and} \bibinfo{person}{Yi Cai}.} \bibinfo{year}{2024}\natexlab{f}.
\newblock \showarticletitle{Towards Effective Time-Aware Language Representation: Exploring Enhanced Temporal Understanding in Language Models}.
\newblock \bibinfo{journal}{\emph{arXiv preprint arXiv:2406.01863}} (\bibinfo{year}{2024}).
\newblock


\bibitem[Wang et~al\mbox{.}(2024g)]%
        {wang2024comprehensive}
\bibfield{author}{\bibinfo{person}{Jiaqi Wang}, \bibinfo{person}{Hanqi Jiang}, \bibinfo{person}{Yiheng Liu}, \bibinfo{person}{Chong Ma}, \bibinfo{person}{Xu Zhang}, \bibinfo{person}{Yi Pan}, \bibinfo{person}{Mengyuan Liu}, \bibinfo{person}{Peiran Gu}, \bibinfo{person}{Sichen Xia}, \bibinfo{person}{Wenjun Li}, {et~al\mbox{.}}} \bibinfo{year}{2024}\natexlab{g}.
\newblock \showarticletitle{A comprehensive review of multimodal large language models: Performance and challenges across different tasks}.
\newblock \bibinfo{journal}{\emph{arXiv preprint arXiv:2408.01319}} (\bibinfo{year}{2024}).
\newblock


\bibitem[Wang(2021)]%
        {wang2021analysis}
\bibfield{author}{\bibinfo{person}{Lei Wang}.} \bibinfo{year}{2021}\natexlab{}.
\newblock \showarticletitle{Analysis and evaluation of Kinect-based action recognition algorithms}.
\newblock \bibinfo{journal}{\emph{arXiv preprint arXiv:2112.08626}} (\bibinfo{year}{2021}).
\newblock


\bibitem[Wang(2023)]%
        {wang2023robust}
\bibfield{author}{\bibinfo{person}{Lei Wang}.} \bibinfo{year}{2023}\natexlab{}.
\newblock \emph{\bibinfo{title}{Robust human action modelling}}.
\newblock \bibinfo{thesistype}{Ph.\,D. Dissertation}. \bibinfo{school}{The Australian National University (Australia)}.
\newblock


\bibitem[Wang et~al\mbox{.}(2023b)]%
        {wang2023videomae}
\bibfield{author}{\bibinfo{person}{Limin Wang}, \bibinfo{person}{Bingkun Huang}, \bibinfo{person}{Zhiyu Zhao}, \bibinfo{person}{Zhan Tong}, \bibinfo{person}{Yinan He}, \bibinfo{person}{Yi Wang}, \bibinfo{person}{Yali Wang}, {and} \bibinfo{person}{Yu Qiao}.} \bibinfo{year}{2023}\natexlab{b}.
\newblock \showarticletitle{Videomae v2: Scaling video masked autoencoders with dual masking}. In \bibinfo{booktitle}{\emph{Proceedings of the IEEE/CVF Conference on Computer Vision and Pattern Recognition}}. \bibinfo{pages}{14549--14560}.
\newblock


\bibitem[Wang et~al\mbox{.}(2019a)]%
        {wang2019comparative}
\bibfield{author}{\bibinfo{person}{Lei Wang}, \bibinfo{person}{Du~Q Huynh}, {and} \bibinfo{person}{Piotr Koniusz}.} \bibinfo{year}{2019}\natexlab{a}.
\newblock \showarticletitle{A comparative review of recent kinect-based action recognition algorithms}.
\newblock \bibinfo{journal}{\emph{IEEE Transactions on Image Processing}}  \bibinfo{volume}{29} (\bibinfo{year}{2019}), \bibinfo{pages}{15--28}.
\newblock


\bibitem[Wang et~al\mbox{.}(2019b)]%
        {wang2019loss}
\bibfield{author}{\bibinfo{person}{Lei Wang}, \bibinfo{person}{Du~Q Huynh}, {and} \bibinfo{person}{Moussa~Reda Mansour}.} \bibinfo{year}{2019}\natexlab{b}.
\newblock \showarticletitle{Loss switching fusion with similarity search for video classification}. In \bibinfo{booktitle}{\emph{2019 IEEE international conference on image processing (ICIP)}}. IEEE, \bibinfo{pages}{974--978}.
\newblock


\bibitem[Wang and Koniusz(2021)]%
        {wang2021self}
\bibfield{author}{\bibinfo{person}{Lei Wang} {and} \bibinfo{person}{Piotr Koniusz}.} \bibinfo{year}{2021}\natexlab{}.
\newblock \showarticletitle{Self-supervising action recognition by statistical moment and subspace descriptors}. In \bibinfo{booktitle}{\emph{Proceedings of the 29th ACM international conference on multimedia}}. \bibinfo{pages}{4324--4333}.
\newblock


\bibitem[Wang and Koniusz(2022a)]%
        {wang2022temporal}
\bibfield{author}{\bibinfo{person}{Lei Wang} {and} \bibinfo{person}{Piotr Koniusz}.} \bibinfo{year}{2022}\natexlab{a}.
\newblock \showarticletitle{Temporal-viewpoint transportation plan for skeletal few-shot action recognition}. In \bibinfo{booktitle}{\emph{Proceedings of the Asian Conference on Computer Vision}}. \bibinfo{pages}{4176--4193}.
\newblock


\bibitem[Wang and Koniusz(2022b)]%
        {wang2022uncertainty}
\bibfield{author}{\bibinfo{person}{Lei Wang} {and} \bibinfo{person}{Piotr Koniusz}.} \bibinfo{year}{2022}\natexlab{b}.
\newblock \showarticletitle{Uncertainty-dtw for time series and sequences}. In \bibinfo{booktitle}{\emph{European Conference on Computer Vision}}. Springer, \bibinfo{pages}{176--195}.
\newblock


\bibitem[Wang and Koniusz(2023)]%
        {wang20233mformer}
\bibfield{author}{\bibinfo{person}{Lei Wang} {and} \bibinfo{person}{Piotr Koniusz}.} \bibinfo{year}{2023}\natexlab{}.
\newblock \showarticletitle{3mformer: Multi-order multi-mode transformer for skeletal action recognition}. In \bibinfo{booktitle}{\emph{Proceedings of the IEEE/CVF Conference on Computer Vision and Pattern Recognition}}. \bibinfo{pages}{5620--5631}.
\newblock


\bibitem[Wang and Koniusz(2024)]%
        {wang2024flow}
\bibfield{author}{\bibinfo{person}{Lei Wang} {and} \bibinfo{person}{Piotr Koniusz}.} \bibinfo{year}{2024}\natexlab{}.
\newblock \showarticletitle{Flow dynamics correction for action recognition}. In \bibinfo{booktitle}{\emph{ICASSP 2024-2024 IEEE International Conference on Acoustics, Speech and Signal Processing (ICASSP)}}. IEEE, \bibinfo{pages}{3795--3799}.
\newblock


\bibitem[Wang et~al\mbox{.}(2019c)]%
        {wang2019hallucinating}
\bibfield{author}{\bibinfo{person}{Lei Wang}, \bibinfo{person}{Piotr Koniusz}, {and} \bibinfo{person}{Du~Q Huynh}.} \bibinfo{year}{2019}\natexlab{c}.
\newblock \showarticletitle{Hallucinating idt descriptors and i3d optical flow features for action recognition with cnns}. In \bibinfo{booktitle}{\emph{Proceedings of the IEEE/CVF international conference on computer vision}}. \bibinfo{pages}{8698--8708}.
\newblock


\bibitem[Wang et~al\mbox{.}(2021a)]%
        {wang20213d}
\bibfield{author}{\bibinfo{person}{Lei Wang}, \bibinfo{person}{Jun Liu}, {and} \bibinfo{person}{Piotr Koniusz}.} \bibinfo{year}{2021}\natexlab{a}.
\newblock \showarticletitle{3D Skeleton-based Few-shot Action Recognition with JEANIE is not so Na\"ive}.
\newblock \bibinfo{journal}{\emph{arXiv preprint arXiv:2112.12668}} (\bibinfo{year}{2021}).
\newblock


\bibitem[Wang et~al\mbox{.}(2024j)]%
        {wang2024meet}
\bibfield{author}{\bibinfo{person}{Lei Wang}, \bibinfo{person}{Jun Liu}, \bibinfo{person}{Liang Zheng}, \bibinfo{person}{Tom Gedeon}, {and} \bibinfo{person}{Piotr Koniusz}.} \bibinfo{year}{2024}\natexlab{j}.
\newblock \showarticletitle{Meet JEANIE: a Similarity Measure for 3D Skeleton Sequences via Temporal-Viewpoint Alignment}.
\newblock \bibinfo{journal}{\emph{International Journal of Computer Vision}} (\bibinfo{year}{2024}), \bibinfo{pages}{1--32}.
\newblock


\bibitem[Wang et~al\mbox{.}(2024k)]%
        {wang2024high}
\bibfield{author}{\bibinfo{person}{Lei Wang}, \bibinfo{person}{Ke Sun}, {and} \bibinfo{person}{Piotr Koniusz}.} \bibinfo{year}{2024}\natexlab{k}.
\newblock \showarticletitle{High-order tensor pooling with attention for action recognition}. In \bibinfo{booktitle}{\emph{ICASSP 2024-2024 IEEE International Conference on Acoustics, Speech and Signal Processing (ICASSP)}}. IEEE, \bibinfo{pages}{3885--3889}.
\newblock


\bibitem[Wang et~al\mbox{.}({[n.\,d.]})]%
        {wangtaylor}
\bibfield{author}{\bibinfo{person}{Lei Wang}, \bibinfo{person}{Xiuyuan Yuan}, \bibinfo{person}{Tom Gedeon}, {and} \bibinfo{person}{Liang Zheng}.} \bibinfo{year}{[n.\,d.]}\natexlab{}.
\newblock \showarticletitle{Taylor Videos for Action Recognition}. In \bibinfo{booktitle}{\emph{Forty-first International Conference on Machine Learning}}.
\newblock


\bibitem[Wang et~al\mbox{.}(2024a)]%
        {wang2024qwen2}
\bibfield{author}{\bibinfo{person}{Peng Wang}, \bibinfo{person}{Shuai Bai}, \bibinfo{person}{Sinan Tan}, \bibinfo{person}{Shijie Wang}, \bibinfo{person}{Zhihao Fan}, \bibinfo{person}{Jinze Bai}, \bibinfo{person}{Keqin Chen}, \bibinfo{person}{Xuejing Liu}, \bibinfo{person}{Jialin Wang}, \bibinfo{person}{Wenbin Ge}, {et~al\mbox{.}}} \bibinfo{year}{2024}\natexlab{a}.
\newblock \showarticletitle{Qwen2-vl: Enhancing vision-language model's perception of the world at any resolution}.
\newblock \bibinfo{journal}{\emph{arXiv preprint arXiv:2409.12191}} (\bibinfo{year}{2024}).
\newblock


\bibitem[Wang et~al\mbox{.}(2021b)]%
        {wang2021make}
\bibfield{author}{\bibinfo{person}{Shaojie Wang}, \bibinfo{person}{Wentian Zhao}, \bibinfo{person}{Ziyi Kou}, \bibinfo{person}{Jing Shi}, {and} \bibinfo{person}{Chenliang Xu}.} \bibinfo{year}{2021}\natexlab{b}.
\newblock \showarticletitle{How to make a blt sandwich? learning vqa towards understanding web instructional videos}. In \bibinfo{booktitle}{\emph{Proceedings of the IEEE/CVF Winter Conference on Applications of Computer Vision}}. \bibinfo{pages}{1130--1139}.
\newblock


\bibitem[Wang et~al\mbox{.}(2024b)]%
        {wang2024visionllm}
\bibfield{author}{\bibinfo{person}{Wenhai Wang}, \bibinfo{person}{Zhe Chen}, \bibinfo{person}{Xiaokang Chen}, \bibinfo{person}{Jiannan Wu}, \bibinfo{person}{Xizhou Zhu}, \bibinfo{person}{Gang Zeng}, \bibinfo{person}{Ping Luo}, \bibinfo{person}{Tong Lu}, \bibinfo{person}{Jie Zhou}, \bibinfo{person}{Yu Qiao}, {et~al\mbox{.}}} \bibinfo{year}{2024}\natexlab{b}.
\newblock \showarticletitle{Visionllm: Large language model is also an open-ended decoder for vision-centric tasks}.
\newblock \bibinfo{journal}{\emph{Advances in Neural Information Processing Systems}}  \bibinfo{volume}{36} (\bibinfo{year}{2024}).
\newblock


\bibitem[Wang et~al\mbox{.}(2024d)]%
        {wang2024mpo}
\bibfield{author}{\bibinfo{person}{Weiyun Wang}, \bibinfo{person}{Zhe Chen}, \bibinfo{person}{Wenhai Wang}, \bibinfo{person}{Yue Cao}, \bibinfo{person}{Yangzhou Liu}, \bibinfo{person}{Zhangwei Gao}, \bibinfo{person}{Jinguo Zhu}, \bibinfo{person}{Xizhou Zhu}, \bibinfo{person}{Lewei Lu}, \bibinfo{person}{Yu Qiao}, {and} \bibinfo{person}{Jifeng Dai}.} \bibinfo{year}{2024}\natexlab{d}.
\newblock \showarticletitle{Enhancing the Reasoning Ability of Multimodal Large Language Models via Mixed Preference Optimization}.
\newblock \bibinfo{journal}{\emph{arXiv preprint arXiv:2411.10442}} (\bibinfo{year}{2024}).
\newblock


\bibitem[Wang et~al\mbox{.}(2019d)]%
        {wang2019vatex}
\bibfield{author}{\bibinfo{person}{Xin Wang}, \bibinfo{person}{Jiawei Wu}, \bibinfo{person}{Junkun Chen}, \bibinfo{person}{Lei Li}, \bibinfo{person}{Yuan-Fang Wang}, {and} \bibinfo{person}{William~Yang Wang}.} \bibinfo{year}{2019}\natexlab{d}.
\newblock \showarticletitle{Vatex: A large-scale, high-quality multilingual dataset for video-and-language research}. In \bibinfo{booktitle}{\emph{Proceedings of the IEEE/CVF international conference on computer vision}}. \bibinfo{pages}{4581--4591}.
\newblock


\bibitem[Wang et~al\mbox{.}(2024e)]%
        {wang2024internvidlargescalevideotextdataset}
\bibfield{author}{\bibinfo{person}{Yi Wang}, \bibinfo{person}{Yinan He}, \bibinfo{person}{Yizhuo Li}, \bibinfo{person}{Kunchang Li}, \bibinfo{person}{Jiashuo Yu}, \bibinfo{person}{Xin Ma}, \bibinfo{person}{Xinhao Li}, \bibinfo{person}{Guo Chen}, \bibinfo{person}{Xinyuan Chen}, \bibinfo{person}{Yaohui Wang}, \bibinfo{person}{Conghui He}, \bibinfo{person}{Ping Luo}, \bibinfo{person}{Ziwei Liu}, \bibinfo{person}{Yali Wang}, \bibinfo{person}{Limin Wang}, {and} \bibinfo{person}{Yu Qiao}.} \bibinfo{year}{2024}\natexlab{e}.
\newblock \bibinfo{title}{InternVid: A Large-scale Video-Text Dataset for Multimodal Understanding and Generation}.
\newblock
\newblock
\showeprint[arxiv]{2307.06942}~[cs.CV]
\urldef\tempurl%
\url{https://arxiv.org/abs/2307.06942}
\showURL{%
\tempurl}


\bibitem[Wang et~al\mbox{.}(2024h)]%
        {wang2024internvideo2}
\bibfield{author}{\bibinfo{person}{Yi Wang}, \bibinfo{person}{Kunchang Li}, \bibinfo{person}{Xinhao Li}, \bibinfo{person}{Jiashuo Yu}, \bibinfo{person}{Yinan He}, \bibinfo{person}{Guo Chen}, \bibinfo{person}{Baoqi Pei}, \bibinfo{person}{Rongkun Zheng}, \bibinfo{person}{Jilan Xu}, \bibinfo{person}{Zun Wang}, {et~al\mbox{.}}} \bibinfo{year}{2024}\natexlab{h}.
\newblock \showarticletitle{Internvideo2: Scaling video foundation models for multimodal video understanding}.
\newblock \bibinfo{journal}{\emph{arXiv e-prints}} (\bibinfo{year}{2024}), \bibinfo{pages}{arXiv--2403}.
\newblock


\bibitem[Wang et~al\mbox{.}(2022b)]%
        {wang2022internvideo}
\bibfield{author}{\bibinfo{person}{Yi Wang}, \bibinfo{person}{Kunchang Li}, \bibinfo{person}{Yizhuo Li}, \bibinfo{person}{Yinan He}, \bibinfo{person}{Bingkun Huang}, \bibinfo{person}{Zhiyu Zhao}, \bibinfo{person}{Hongjie Zhang}, \bibinfo{person}{Jilan Xu}, \bibinfo{person}{Yi Liu}, \bibinfo{person}{Zun Wang}, {et~al\mbox{.}}} \bibinfo{year}{2022}\natexlab{b}.
\newblock \showarticletitle{Internvideo: General video foundation models via generative and discriminative learning}.
\newblock \bibinfo{journal}{\emph{arXiv preprint arXiv:2212.03191}} (\bibinfo{year}{2022}).
\newblock


\bibitem[Wang et~al\mbox{.}(2024m)]%
        {wang2024loong}
\bibfield{author}{\bibinfo{person}{Yuqing Wang}, \bibinfo{person}{Tianwei Xiong}, \bibinfo{person}{Daquan Zhou}, \bibinfo{person}{Zhijie Lin}, \bibinfo{person}{Yang Zhao}, \bibinfo{person}{Bingyi Kang}, \bibinfo{person}{Jiashi Feng}, {and} \bibinfo{person}{Xihui Liu}.} \bibinfo{year}{2024}\natexlab{m}.
\newblock \showarticletitle{Loong: Generating Minute-level Long Videos with Autoregressive Language Models}.
\newblock \bibinfo{journal}{\emph{arXiv preprint arXiv:2410.02757}} (\bibinfo{year}{2024}).
\newblock


\bibitem[Wang et~al\mbox{.}(2024l)]%
        {wang2024gpt4video}
\bibfield{author}{\bibinfo{person}{Zhanyu Wang}, \bibinfo{person}{Longyue Wang}, \bibinfo{person}{Zhen Zhao}, \bibinfo{person}{Minghao Wu}, \bibinfo{person}{Chenyang Lyu}, \bibinfo{person}{Huayang Li}, \bibinfo{person}{Deng Cai}, \bibinfo{person}{Luping Zhou}, \bibinfo{person}{Shuming Shi}, {and} \bibinfo{person}{Zhaopeng Tu}.} \bibinfo{year}{2024}\natexlab{l}.
\newblock \showarticletitle{Gpt4video: A unified multimodal large language model for lnstruction-followed understanding and safety-aware generation}. In \bibinfo{booktitle}{\emph{Proceedings of the 32nd ACM International Conference on Multimedia}}. \bibinfo{pages}{3907--3916}.
\newblock


\bibitem[Wu et~al\mbox{.}(2024)]%
        {wu2024star}
\bibfield{author}{\bibinfo{person}{Bo Wu}, \bibinfo{person}{Shoubin Yu}, \bibinfo{person}{Zhenfang Chen}, \bibinfo{person}{Joshua~B Tenenbaum}, {and} \bibinfo{person}{Chuang Gan}.} \bibinfo{year}{2024}\natexlab{}.
\newblock \showarticletitle{Star: A benchmark for situated reasoning in real-world videos}.
\newblock \bibinfo{journal}{\emph{arXiv preprint arXiv:2405.09711}} (\bibinfo{year}{2024}).
\newblock


\bibitem[Wu et~al\mbox{.}(2023b)]%
        {wu2023visualchatgpttalkingdrawing}
\bibfield{author}{\bibinfo{person}{Chenfei Wu}, \bibinfo{person}{Shengming Yin}, \bibinfo{person}{Weizhen Qi}, \bibinfo{person}{Xiaodong Wang}, \bibinfo{person}{Zecheng Tang}, {and} \bibinfo{person}{Nan Duan}.} \bibinfo{year}{2023}\natexlab{b}.
\newblock \bibinfo{title}{Visual ChatGPT: Talking, Drawing and Editing with Visual Foundation Models}.
\newblock
\newblock
\showeprint[arxiv]{2303.04671}~[cs.CV]
\urldef\tempurl%
\url{https://arxiv.org/abs/2303.04671}
\showURL{%
\tempurl}


\bibitem[Wu et~al\mbox{.}(2020)]%
        {wu2020looklistenlearningmultimodal}
\bibfield{author}{\bibinfo{person}{Peng Wu}, \bibinfo{person}{Jing Liu}, \bibinfo{person}{Yujia Shi}, \bibinfo{person}{Yujia Sun}, \bibinfo{person}{Fangtao Shao}, \bibinfo{person}{Zhaoyang Wu}, {and} \bibinfo{person}{Zhiwei Yang}.} \bibinfo{year}{2020}\natexlab{}.
\newblock \bibinfo{title}{Not only Look, but also Listen: Learning Multimodal Violence Detection under Weak Supervision}.
\newblock
\newblock
\showeprint[arxiv]{2007.04687}~[cs.CV]
\urldef\tempurl%
\url{https://arxiv.org/abs/2007.04687}
\showURL{%
\tempurl}


\bibitem[Wu et~al\mbox{.}(2023a)]%
        {wu2023next}
\bibfield{author}{\bibinfo{person}{Shengqiong Wu}, \bibinfo{person}{Hao Fei}, \bibinfo{person}{Leigang Qu}, \bibinfo{person}{Wei Ji}, {and} \bibinfo{person}{Tat-Seng Chua}.} \bibinfo{year}{2023}\natexlab{a}.
\newblock \showarticletitle{Next-gpt: Any-to-any multimodal llm}.
\newblock \bibinfo{journal}{\emph{arXiv preprint arXiv:2309.05519}} (\bibinfo{year}{2023}).
\newblock


\bibitem[Wu et~al\mbox{.}(2023c)]%
        {wu2023largecrossmodalvideoretrieval}
\bibfield{author}{\bibinfo{person}{Weijia Wu}, \bibinfo{person}{Yuzhong Zhao}, \bibinfo{person}{Zhuang Li}, \bibinfo{person}{Jiahong Li}, \bibinfo{person}{Hong Zhou}, \bibinfo{person}{Mike~Zheng Shou}, {and} \bibinfo{person}{Xiang Bai}.} \bibinfo{year}{2023}\natexlab{c}.
\newblock \bibinfo{title}{A Large Cross-Modal Video Retrieval Dataset with Reading Comprehension}.
\newblock
\newblock
\showeprint[arxiv]{2305.03347}~[cs.CV]
\urldef\tempurl%
\url{https://arxiv.org/abs/2305.03347}
\showURL{%
\tempurl}


\bibitem[Xu and Poo(2023)]%
        {xu2023large}
\bibfield{author}{\bibinfo{person}{Bo Xu} {and} \bibinfo{person}{Mu-ming Poo}.} \bibinfo{year}{2023}\natexlab{}.
\newblock \showarticletitle{Large language models and brain-inspired general intelligence}.
\newblock \bibinfo{journal}{\emph{National Science Review}} \bibinfo{volume}{10}, \bibinfo{number}{10} (\bibinfo{year}{2023}), \bibinfo{pages}{nwad267}.
\newblock


\bibitem[Xu et~al\mbox{.}({[n.\,d.]})]%
        {xu2017video}
\bibfield{author}{\bibinfo{person}{Dejing Xu}, \bibinfo{person}{Zhou Zhao}, \bibinfo{person}{Jun Xiao}, \bibinfo{person}{Fei Wu}, \bibinfo{person}{Hanwang Zhang}, \bibinfo{person}{Xiangnan He}, {and} \bibinfo{person}{Yueting Zhuang}.} \bibinfo{year}{[n.\,d.]}\natexlab{}.
\newblock \showarticletitle{Video Question Answering via Gradually Refined Attention over Appearance and Motion}. In \bibinfo{booktitle}{\emph{ACM Multimedia}}.
\newblock


\bibitem[Xu et~al\mbox{.}(2023)]%
        {xu2023mplug}
\bibfield{author}{\bibinfo{person}{Haiyang Xu}, \bibinfo{person}{Qinghao Ye}, \bibinfo{person}{Ming Yan}, \bibinfo{person}{Yaya Shi}, \bibinfo{person}{Jiabo Ye}, \bibinfo{person}{Yuanhong Xu}, \bibinfo{person}{Chenliang Li}, \bibinfo{person}{Bin Bi}, \bibinfo{person}{Qi Qian}, \bibinfo{person}{Wei Wang}, {et~al\mbox{.}}} \bibinfo{year}{2023}\natexlab{}.
\newblock \showarticletitle{mplug-2: A modularized multi-modal foundation model across text, image and video}. In \bibinfo{booktitle}{\emph{International Conference on Machine Learning}}. PMLR, \bibinfo{pages}{38728--38748}.
\newblock


\bibitem[Xu et~al\mbox{.}(2016)]%
        {xu2016msr}
\bibfield{author}{\bibinfo{person}{Jun Xu}, \bibinfo{person}{Tao Mei}, \bibinfo{person}{Ting Yao}, {and} \bibinfo{person}{Yong Rui}.} \bibinfo{year}{2016}\natexlab{}.
\newblock \showarticletitle{Msr-vtt: A large video description dataset for bridging video and language}. In \bibinfo{booktitle}{\emph{Proceedings of the IEEE conference on computer vision and pattern recognition}}. \bibinfo{pages}{5288--5296}.
\newblock


\bibitem[Xu et~al\mbox{.}(2024)]%
        {xu2024pllava}
\bibfield{author}{\bibinfo{person}{Lin Xu}, \bibinfo{person}{Yilin Zhao}, \bibinfo{person}{Daquan Zhou}, \bibinfo{person}{Zhijie Lin}, \bibinfo{person}{See~Kiong Ng}, {and} \bibinfo{person}{Jiashi Feng}.} \bibinfo{year}{2024}\natexlab{}.
\newblock \showarticletitle{Pllava: Parameter-free llava extension from images to videos for video dense captioning}.
\newblock \bibinfo{journal}{\emph{arXiv preprint arXiv:2404.16994}} (\bibinfo{year}{2024}).
\newblock


\bibitem[Yang et~al\mbox{.}(2023)]%
        {yang2023vid2seq}
\bibfield{author}{\bibinfo{person}{Antoine Yang}, \bibinfo{person}{Arsha Nagrani}, \bibinfo{person}{Paul~Hongsuck Seo}, \bibinfo{person}{Antoine Miech}, \bibinfo{person}{Jordi Pont-Tuset}, \bibinfo{person}{Ivan Laptev}, \bibinfo{person}{Josef Sivic}, {and} \bibinfo{person}{Cordelia Schmid}.} \bibinfo{year}{2023}\natexlab{}.
\newblock \showarticletitle{Vid2seq: Large-scale pretraining of a visual language model for dense video captioning}. In \bibinfo{booktitle}{\emph{Proceedings of the IEEE/CVF Conference on Computer Vision and Pattern Recognition}}. \bibinfo{pages}{10714--10726}.
\newblock


\bibitem[Yang et~al\mbox{.}(2024b)]%
        {yang2024qwen2}
\bibfield{author}{\bibinfo{person}{An Yang}, \bibinfo{person}{Baosong Yang}, \bibinfo{person}{Binyuan Hui}, \bibinfo{person}{Bo Zheng}, \bibinfo{person}{Bowen Yu}, \bibinfo{person}{Chang Zhou}, \bibinfo{person}{Chengpeng Li}, \bibinfo{person}{Chengyuan Li}, \bibinfo{person}{Dayiheng Liu}, \bibinfo{person}{Fei Huang}, {et~al\mbox{.}}} \bibinfo{year}{2024}\natexlab{b}.
\newblock \showarticletitle{Qwen2 technical report}.
\newblock \bibinfo{journal}{\emph{arXiv preprint arXiv:2407.10671}} (\bibinfo{year}{2024}).
\newblock


\bibitem[Yang et~al\mbox{.}(2024a)]%
        {yang2024vript}
\bibfield{author}{\bibinfo{person}{Dongjie Yang}, \bibinfo{person}{Suyuan Huang}, \bibinfo{person}{Chengqiang Lu}, \bibinfo{person}{Xiaodong Han}, \bibinfo{person}{Haoxin Zhang}, \bibinfo{person}{Yan Gao}, \bibinfo{person}{Yao Hu}, {and} \bibinfo{person}{Hai Zhao}.} \bibinfo{year}{2024}\natexlab{a}.
\newblock \showarticletitle{Vript: A Video Is Worth Thousands of Words}.
\newblock \bibinfo{journal}{\emph{arXiv preprint arXiv:2406.06040}} (\bibinfo{year}{2024}).
\newblock


\bibitem[Yang et~al\mbox{.}(2024c)]%
        {yang2024emollm}
\bibfield{author}{\bibinfo{person}{Qu Yang}, \bibinfo{person}{Mang Ye}, {and} \bibinfo{person}{Bo Du}.} \bibinfo{year}{2024}\natexlab{c}.
\newblock \showarticletitle{Emollm: Multimodal emotional understanding meets large language models}.
\newblock \bibinfo{journal}{\emph{arXiv preprint arXiv:2406.16442}} (\bibinfo{year}{2024}).
\newblock


\bibitem[Yu et~al\mbox{.}(2024)]%
        {yu2024languagemodelbeatsdiffusion}
\bibfield{author}{\bibinfo{person}{Lijun Yu}, \bibinfo{person}{José Lezama}, \bibinfo{person}{Nitesh~B. Gundavarapu}, \bibinfo{person}{Luca Versari}, \bibinfo{person}{Kihyuk Sohn}, \bibinfo{person}{David Minnen}, \bibinfo{person}{Yong Cheng}, \bibinfo{person}{Vighnesh Birodkar}, \bibinfo{person}{Agrim Gupta}, \bibinfo{person}{Xiuye Gu}, \bibinfo{person}{Alexander~G. Hauptmann}, \bibinfo{person}{Boqing Gong}, \bibinfo{person}{Ming-Hsuan Yang}, \bibinfo{person}{Irfan Essa}, \bibinfo{person}{David~A. Ross}, {and} \bibinfo{person}{Lu Jiang}.} \bibinfo{year}{2024}\natexlab{}.
\newblock \bibinfo{title}{Language Model Beats Diffusion -- Tokenizer is Key to Visual Generation}.
\newblock
\newblock
\showeprint[arxiv]{2310.05737}~[cs.CV]
\urldef\tempurl%
\url{https://arxiv.org/abs/2310.05737}
\showURL{%
\tempurl}


\bibitem[Yu et~al\mbox{.}(2019)]%
        {yu2019activitynet}
\bibfield{author}{\bibinfo{person}{Zhou Yu}, \bibinfo{person}{Dejing Xu}, \bibinfo{person}{Jun Yu}, \bibinfo{person}{Ting Yu}, \bibinfo{person}{Zhou Zhao}, \bibinfo{person}{Yueting Zhuang}, {and} \bibinfo{person}{Dacheng Tao}.} \bibinfo{year}{2019}\natexlab{}.
\newblock \showarticletitle{Activitynet-qa: A dataset for understanding complex web videos via question answering}. In \bibinfo{booktitle}{\emph{Proceedings of the AAAI Conference on Artificial Intelligence}}, Vol.~\bibinfo{volume}{33}. \bibinfo{pages}{9127--9134}.
\newblock


\bibitem[Yuan et~al\mbox{.}(2021)]%
        {yuan2021florence}
\bibfield{author}{\bibinfo{person}{Lu Yuan}, \bibinfo{person}{Dongdong Chen}, \bibinfo{person}{Yi-Ling Chen}, \bibinfo{person}{Noel Codella}, \bibinfo{person}{Xiyang Dai}, \bibinfo{person}{Jianfeng Gao}, \bibinfo{person}{Houdong Hu}, \bibinfo{person}{Xuedong Huang}, \bibinfo{person}{Boxin Li}, \bibinfo{person}{Chunyuan Li}, {et~al\mbox{.}}} \bibinfo{year}{2021}\natexlab{}.
\newblock \showarticletitle{Florence: A new foundation model for computer vision}.
\newblock \bibinfo{journal}{\emph{arXiv preprint arXiv:2111.11432}} (\bibinfo{year}{2021}).
\newblock


\bibitem[Zanella et~al\mbox{.}(2024)]%
        {zanella2024harnessing}
\bibfield{author}{\bibinfo{person}{Luca Zanella}, \bibinfo{person}{Willi Menapace}, \bibinfo{person}{Massimiliano Mancini}, \bibinfo{person}{Yiming Wang}, {and} \bibinfo{person}{Elisa Ricci}.} \bibinfo{year}{2024}\natexlab{}.
\newblock \showarticletitle{Harnessing Large Language Models for Training-free Video Anomaly Detection}. In \bibinfo{booktitle}{\emph{Proceedings of the IEEE/CVF Conference on Computer Vision and Pattern Recognition}}. \bibinfo{pages}{18527--18536}.
\newblock


\bibitem[Zhai et~al\mbox{.}(2023)]%
        {zhai2023sigmoid}
\bibfield{author}{\bibinfo{person}{Xiaohua Zhai}, \bibinfo{person}{Basil Mustafa}, \bibinfo{person}{Alexander Kolesnikov}, {and} \bibinfo{person}{Lucas Beyer}.} \bibinfo{year}{2023}\natexlab{}.
\newblock \showarticletitle{Sigmoid loss for language image pre-training}. In \bibinfo{booktitle}{\emph{Proceedings of the IEEE/CVF International Conference on Computer Vision}}. \bibinfo{pages}{11975--11986}.
\newblock


\bibitem[Zhang et~al\mbox{.}(2022)]%
        {zhang2022actionformer}
\bibfield{author}{\bibinfo{person}{Chen-Lin Zhang}, \bibinfo{person}{Jianxin Wu}, {and} \bibinfo{person}{Yin Li}.} \bibinfo{year}{2022}\natexlab{}.
\newblock \showarticletitle{Actionformer: Localizing moments of actions with transformers}. In \bibinfo{booktitle}{\emph{European Conference on Computer Vision}}. Springer, \bibinfo{pages}{492--510}.
\newblock


\bibitem[Zhang et~al\mbox{.}(2023a)]%
        {Zhang2023VideoLLaMAAI}
\bibfield{author}{\bibinfo{person}{Hang Zhang}, \bibinfo{person}{Xin Li}, {and} \bibinfo{person}{Lidong Bing}.} \bibinfo{year}{2023}\natexlab{a}.
\newblock \showarticletitle{Video-LLaMA: An Instruction-tuned Audio-Visual Language Model for Video Understanding}. In \bibinfo{booktitle}{\emph{Conference on Empirical Methods in Natural Language Processing}}.
\newblock
\urldef\tempurl%
\url{https://api.semanticscholar.org/CorpusID:259075356}
\showURL{%
\tempurl}


\bibitem[Zhang et~al\mbox{.}(2024c)]%
        {zhang2024holmes}
\bibfield{author}{\bibinfo{person}{Huaxin Zhang}, \bibinfo{person}{Xiaohao Xu}, \bibinfo{person}{Xiang Wang}, \bibinfo{person}{Jialong Zuo}, \bibinfo{person}{Chuchu Han}, \bibinfo{person}{Xiaonan Huang}, \bibinfo{person}{Changxin Gao}, \bibinfo{person}{Yuehuan Wang}, {and} \bibinfo{person}{Nong Sang}.} \bibinfo{year}{2024}\natexlab{c}.
\newblock \showarticletitle{Holmes-VAD: Towards Unbiased and Explainable Video Anomaly Detection via Multi-modal LLM}.
\newblock \bibinfo{journal}{\emph{arXiv preprint arXiv:2406.12235}} (\bibinfo{year}{2024}).
\newblock


\bibitem[Zhang et~al\mbox{.}(2024b)]%
        {zhang2024task}
\bibfield{author}{\bibinfo{person}{Jieyu Zhang}, \bibinfo{person}{Weikai Huang}, \bibinfo{person}{Zixian Ma}, \bibinfo{person}{Oscar Michel}, \bibinfo{person}{Dong He}, \bibinfo{person}{Tanmay Gupta}, \bibinfo{person}{Wei-Chiu Ma}, \bibinfo{person}{Ali Farhadi}, \bibinfo{person}{Aniruddha Kembhavi}, {and} \bibinfo{person}{Ranjay Krishna}.} \bibinfo{year}{2024}\natexlab{b}.
\newblock \showarticletitle{Task Me Anything}.
\newblock \bibinfo{journal}{\emph{arXiv preprint arXiv:2406.11775}} (\bibinfo{year}{2024}).
\newblock


\bibitem[Zhang et~al\mbox{.}(2023b)]%
        {zhang2023t2mgptgeneratinghumanmotion}
\bibfield{author}{\bibinfo{person}{Jianrong Zhang}, \bibinfo{person}{Yangsong Zhang}, \bibinfo{person}{Xiaodong Cun}, \bibinfo{person}{Shaoli Huang}, \bibinfo{person}{Yong Zhang}, \bibinfo{person}{Hongwei Zhao}, \bibinfo{person}{Hongtao Lu}, {and} \bibinfo{person}{Xi Shen}.} \bibinfo{year}{2023}\natexlab{b}.
\newblock \bibinfo{title}{T2M-GPT: Generating Human Motion from Textual Descriptions with Discrete Representations}.
\newblock
\newblock
\showeprint[arxiv]{2301.06052}~[cs.CV]
\urldef\tempurl%
\url{https://arxiv.org/abs/2301.06052}
\showURL{%
\tempurl}


\bibitem[Zhang et~al\mbox{.}(2024a)]%
        {zhang2024analyzing}
\bibfield{author}{\bibinfo{person}{Zhihan Zhang}, \bibinfo{person}{Yixin Cao}, \bibinfo{person}{Chenchen Ye}, \bibinfo{person}{Yunshan Ma}, \bibinfo{person}{Lizi Liao}, {and} \bibinfo{person}{Tat-Seng Chua}.} \bibinfo{year}{2024}\natexlab{a}.
\newblock \showarticletitle{Analyzing Temporal Complex Events with Large Language Models? A Benchmark towards Temporal, Long Context Understanding}.
\newblock \bibinfo{journal}{\emph{arXiv preprint arXiv:2406.02472}} (\bibinfo{year}{2024}).
\newblock


\bibitem[Zhao et~al\mbox{.}(2024)]%
        {zhao2024expel}
\bibfield{author}{\bibinfo{person}{Andrew Zhao}, \bibinfo{person}{Daniel Huang}, \bibinfo{person}{Quentin Xu}, \bibinfo{person}{Matthieu Lin}, \bibinfo{person}{Yong-Jin Liu}, {and} \bibinfo{person}{Gao Huang}.} \bibinfo{year}{2024}\natexlab{}.
\newblock \showarticletitle{Expel: Llm agents are experiential learners}. In \bibinfo{booktitle}{\emph{Proceedings of the AAAI Conference on Artificial Intelligence}}, Vol.~\bibinfo{volume}{38}. \bibinfo{pages}{19632--19642}.
\newblock


\bibitem[Zhao et~al\mbox{.}(2023)]%
        {zhao2023learning}
\bibfield{author}{\bibinfo{person}{Yue Zhao}, \bibinfo{person}{Ishan Misra}, \bibinfo{person}{Philipp Kr{\"a}henb{\"u}hl}, {and} \bibinfo{person}{Rohit Girdhar}.} \bibinfo{year}{2023}\natexlab{}.
\newblock \showarticletitle{Learning video representations from large language models}. In \bibinfo{booktitle}{\emph{Proceedings of the IEEE/CVF Conference on Computer Vision and Pattern Recognition}}. \bibinfo{pages}{6586--6597}.
\newblock


\bibitem[Zheng et~al\mbox{.}(2023)]%
        {zheng2023judgingllmasajudgemtbenchchatbot}
\bibfield{author}{\bibinfo{person}{Lianmin Zheng}, \bibinfo{person}{Wei-Lin Chiang}, \bibinfo{person}{Ying Sheng}, \bibinfo{person}{Siyuan Zhuang}, \bibinfo{person}{Zhanghao Wu}, \bibinfo{person}{Yonghao Zhuang}, \bibinfo{person}{Zi Lin}, \bibinfo{person}{Zhuohan Li}, \bibinfo{person}{Dacheng Li}, \bibinfo{person}{Eric~P. Xing}, \bibinfo{person}{Hao Zhang}, \bibinfo{person}{Joseph~E. Gonzalez}, {and} \bibinfo{person}{Ion Stoica}.} \bibinfo{year}{2023}\natexlab{}.
\newblock \bibinfo{title}{Judging LLM-as-a-Judge with MT-Bench and Chatbot Arena}.
\newblock
\newblock
\showeprint[arxiv]{2306.05685}~[cs.CL]
\urldef\tempurl%
\url{https://arxiv.org/abs/2306.05685}
\showURL{%
\tempurl}


\bibitem[Zhou et~al\mbox{.}(2018)]%
        {zhou2018temporal}
\bibfield{author}{\bibinfo{person}{Bolei Zhou}, \bibinfo{person}{Alex Andonian}, \bibinfo{person}{Aude Oliva}, {and} \bibinfo{person}{Antonio Torralba}.} \bibinfo{year}{2018}\natexlab{}.
\newblock \showarticletitle{Temporal relational reasoning in videos}. In \bibinfo{booktitle}{\emph{Proceedings of the European conference on computer vision (ECCV)}}. \bibinfo{pages}{803--818}.
\newblock


\bibitem[Zhou et~al\mbox{.}(2024)]%
        {zhou2024survey}
\bibfield{author}{\bibinfo{person}{Pengyuan Zhou}, \bibinfo{person}{Lin Wang}, \bibinfo{person}{Zhi Liu}, \bibinfo{person}{Yanbin Hao}, \bibinfo{person}{Pan Hui}, \bibinfo{person}{Sasu Tarkoma}, {and} \bibinfo{person}{Jussi Kangasharju}.} \bibinfo{year}{2024}\natexlab{}.
\newblock \showarticletitle{A survey on generative ai and llm for video generation, understanding, and streaming}.
\newblock \bibinfo{journal}{\emph{arXiv preprint arXiv:2404.16038}} (\bibinfo{year}{2024}).
\newblock


\bibitem[Zhu et~al\mbox{.}(2023)]%
        {zhu2023languagebind}
\bibfield{author}{\bibinfo{person}{Bin Zhu}, \bibinfo{person}{Bin Lin}, \bibinfo{person}{Munan Ning}, \bibinfo{person}{Yang Yan}, \bibinfo{person}{Jiaxi Cui}, \bibinfo{person}{HongFa Wang}, \bibinfo{person}{Yatian Pang}, \bibinfo{person}{Wenhao Jiang}, \bibinfo{person}{Junwu Zhang}, \bibinfo{person}{Zongwei Li}, {et~al\mbox{.}}} \bibinfo{year}{2023}\natexlab{}.
\newblock \showarticletitle{Languagebind: Extending video-language pretraining to n-modality by language-based semantic alignment}.
\newblock \bibinfo{journal}{\emph{arXiv preprint arXiv:2310.01852}} (\bibinfo{year}{2023}).
\newblock


\bibitem[Zhu et~al\mbox{.}(2024)]%
        {msad2024}
\bibfield{author}{\bibinfo{person}{Liyun Zhu}, \bibinfo{person}{Lei Wang}, \bibinfo{person}{Arjun Raj}, \bibinfo{person}{Tom Gedeon}, {and} \bibinfo{person}{Chen Chen}.} \bibinfo{year}{2024}\natexlab{}.
\newblock \showarticletitle{Advancing Video Anomaly Detection: A Concise Review and a New Dataset}. In \bibinfo{booktitle}{\emph{The Thirty-eighth Conference on Neural Information Processing Systems Datasets and Benchmarks Track}}.
\newblock


\bibitem[Zohar et~al\mbox{.}(2024)]%
        {zohar2024apolloexplorationvideounderstanding}
\bibfield{author}{\bibinfo{person}{Orr Zohar}, \bibinfo{person}{Xiaohan Wang}, \bibinfo{person}{Yann Dubois}, \bibinfo{person}{Nikhil Mehta}, \bibinfo{person}{Tong Xiao}, \bibinfo{person}{Philippe Hansen-Estruch}, \bibinfo{person}{Licheng Yu}, \bibinfo{person}{Xiaofang Wang}, \bibinfo{person}{Felix Juefei-Xu}, \bibinfo{person}{Ning Zhang}, \bibinfo{person}{Serena Yeung-Levy}, {and} \bibinfo{person}{Xide Xia}.} \bibinfo{year}{2024}\natexlab{}.
\newblock \bibinfo{title}{Apollo: An Exploration of Video Understanding in Large Multimodal Models}.
\newblock
\newblock
\showeprint[arxiv]{2412.10360}~[cs.CV]
\urldef\tempurl%
\url{https://arxiv.org/abs/2412.10360}
\showURL{%
\tempurl}


\end{thebibliography}






\end{document}